\pdfoutput=1

\documentclass[11pt]{article}

\usepackage[]{acl}

\usepackage{times}
\usepackage{latexsym}

\usepackage[T1]{fontenc}

\usepackage[utf8]{inputenc}

\usepackage{microtype}
\usepackage{amsmath}
\usepackage{ulem}
\usepackage{array}
\renewcommand{\arraystretch}{1.5} 
\usepackage{amsmath, amssymb, bm}

\usepackage{booktabs} 
\usepackage{multirow}

\usepackage{graphicx}
\usepackage{subfigure}
\usepackage{subcaption}

\usepackage{lipsum} 
\title{WilKE: Wise-Layer Knowledge Editor for Lifelong Knowledge Editing}

\author{
\vspace{-0.15cm} Chenhui Hu$^{1,2}$, Pengfei Cao$^{1,2}$, Yubo Chen$^{1,2*}$, Kang Liu$^{1,2}$, Jun Zhao$^{1,2}$\thanks{~~Corresponding author.}
 \\
	\vspace{-0.25cm}$^1$The Laboratory of Cognition and Decision Intelligence for Complex Systems, \\
 \vspace{-0.25cm} Institute of Automation, Chinese Academy of Sciences, Beijing, China \\
 \vspace{-0.15cm}
 $^2$School of Artificial Intelligence, University of Chinese Academy of Sciences, Beijing, China \\
	\vspace{-0.25cm}{\tt huchenhui2024@ia.ac.cn} \\
	\vspace{-0.25cm}{\tt \{pengfei.cao,yubo.chen,kliu,jzhao\}@nlpr.ia.ac.cn}
	} 

\begin{document}
\maketitle
\begin{abstract}
Knowledge editing aims to rectify inaccuracies in large language models (LLMs) without costly retraining for outdated or erroneous knowledge. However, current knowledge editing methods primarily focus on single editing, failing to meet the requirements for lifelong editing\footnote{In this paper, lifelong editing is synonymous with lifelong knowledge editing.}. This study reveals a performance degradation encountered by knowledge editing in lifelong editing, characterized by toxicity buildup and toxicity flash, with the primary cause identified as pattern unmatch. We introduce a knowledge editing approach named Wise-Layer Knowledge Editor (WilKE), which selects editing layer based on the pattern matching degree of editing knowledge across different layers in language models. Experimental results demonstrate that, in lifelong editing, WilKE exhibits an average improvement of 46.2\% and 67.8\% on editing GPT2-XL and GPT-J relative to state-of-the-art knowledge editing methods.
 
\end{abstract}

\section{Introduction} \label{introduction}

Large language models (LLMs) encode a wealth of world knowledge through pretraining on massive corpus (\citealp{radford2019language}; \citealp{brown2020language}; \citealp{achiam2023gpt}; \citealp{li2023evaluation}; \citealp{kale2023provenance}). However, outdated or erroneous knowledge may persist, and retraining these models with updated corpus incurs prohibitively high costs. To address this challenge, numerous studies have introduced knowledge editing (\citealp{de2021editing}; \citealp{mitchell2021fast}; \citealp{meng2022locating}; \citealp{meng2022mass}) as a solution, which involves updating the internal parameters of language models to edit specific knowledge.

\begin{figure}
    \centering
    \includegraphics[width=0.48\textwidth]{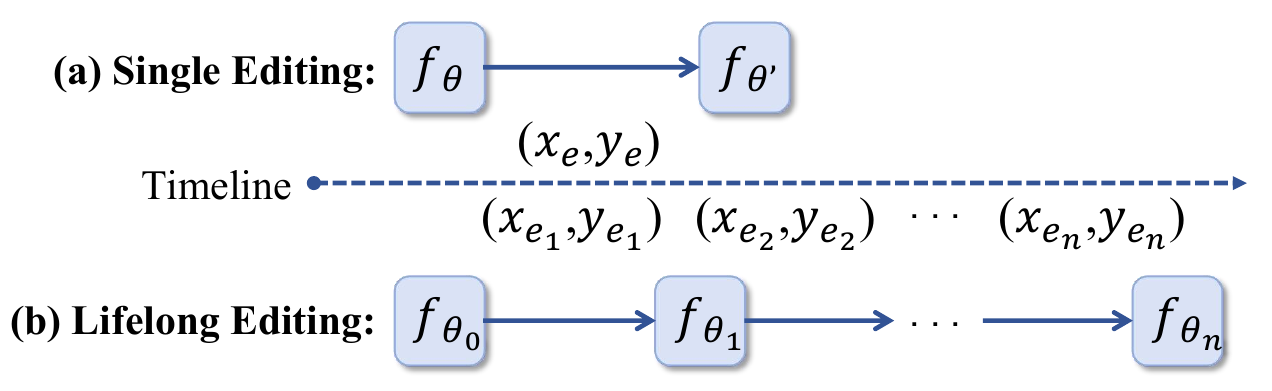}
    \caption{Single editing versus lifelong edit. (a) Single editing only involves making an edit. (b) Life-long editing involves continuous edits and monitoring performance.}
  \label{fig:single&lifelong}
\end{figure}

Current knowledge editing methods are evaluated in single editing by default, which updates a single knowledge \((x_e,y_o)\) to \((x_e,y_e)\) on initial model \(f_\theta\) for each test point, as shown in Figure~\ref{fig:single&lifelong}(a). However, knowledge should be updated continuously in fact, making single editing insufficient to meet the demands. Therefore, we focus on lifelong editing, which updates a knowledge \((x_{e_i},y_{o_i})\) to \((x_{e_i},y_{e_i})\) on model \(f_{e_{i-1}}\) that doesn't have to be initial model \(f_{\theta_0}\), as shown in Figure~\ref{fig:single&lifelong}(b).
 
\begin{figure*}
    \centering
    \includegraphics[width=0.8\textwidth]{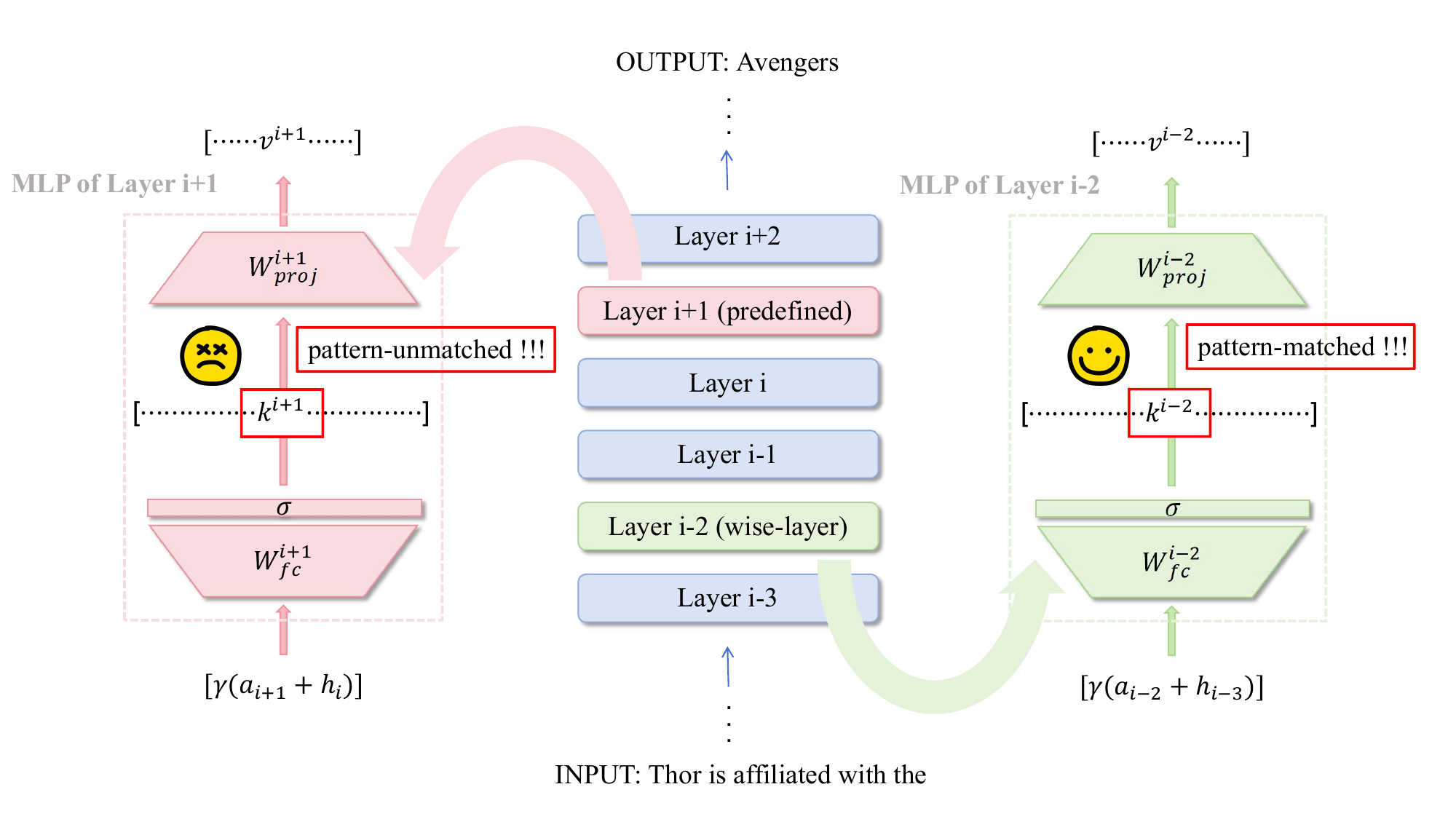}
    \caption{Illustration of our work. Predefined editing layers may not necessarily accommodate all editing knowledge effectively. Therefore, it would be wiser to select different editing layers for different editing knowledge.}
  \label{fig:frame}
\end{figure*}

In this paper, we conduct an analysis of state-of-the-art knowledge editing methods such as ROME \cite{meng2022locating} and MEMIT \cite{meng2022mass}, revealing a severe performance degradation when applied in lifelong editing. Investigating this issue further, our experiments indicate that these methods suffer from \textbf{toxicity buildup} and \textbf{toxicity flash} during ongoing editing. As shown in Figure~\ref{fig:toxicity_gpt2-xl_a},~\ref{fig:toxicity_gpt-j_a}, the combined effects of both phenomena result in a "step-like" shape. On the one hand, the toxicity buildup signifies that one edit induces minor changes in irrelevant parameters, gradually leading to model's failure. On the other hand, the toxicity flash suggests that one edit modifies model's parameters abnormally, resulting in severe overfitting to specific edit, which is not reported in previous research. It's worth noting that due to overfitting, such failures are undetectable in single editing, and achieve respectable scores.

We analyze the primary reasons for these two phenomena, attributing them to pattern unmatch, as illustrated in Figure~\ref{fig:frame}. Specifically, different layers of language model may detect different patterns, which is called key in key-value memories (\citealp{sukhbaatar2015end}; \citealp{sukhbaatar2019augmenting}; \citealp{geva2020transformer}), thus extracting relevant information according to patterns and updating the hidden states. In other words, different knowledge may be stored in different layers, as illustrated in Section~\ref{toxicity_flash}. However, ROME and MEMIT perform knowledge editing at predefined layers, which primarily lead to toxicity buildup and toxicity flash. 

To address this issue, we propose \textbf{Wi}se-\textbf{L}ayer \textbf{K}nowledge \textbf{E}ditor (\textbf{WilKE}), which eliminates the need for predefined editing layer. Instead, WilKE selects editing layer based on the degree of pattern matching for different editing knowledge across various layers. Experimental results demonstrate that WilKE exhibits state-of-the-art comprehensive performance when editing GPT2-XL (1.5B) \cite{radford2019language} and GPT-J (6B) \cite{wang2021gpt}. Specifically, in lifelong editing scenarios, under identical experimental conditions of conducting 1024 edits, WilKE demonstrates an average improvement of 46.2\% and 67.8\% in comprehensive performance relative to state-of-the-art methods when editing GPT2-XL and GPT-J, respectively.

In summary, our primary contributions are as follows:

\begin{itemize}
    \item We investigate the failure of ROME and MEMIT in lifelong editing, revealing toxicicty buildup and toxicity flash during ongoing editing. The underlying primary cause of these phenomena is found to be pattern unmatch.
    \item To address this issue, we introduce WilKE. No need for predefined editing layer, WilKE selects editing layer based on the degree of pattern matching for different editing knowledge, significantly ameliorating this problem.
    \item We conduct experiments in lifelong editing using popular knowledge editing methods on GPT-XL (1.5B) and GPT-J (6B), highlighting the superiority of WilKE over prevalent knowledge editing methods. The source code is available at \href{https://github.com/ChenhuiHu/WilKE}{https://github.com/ChenhuiHu/WilKE}.
\end{itemize}

\section{Related Work} \label{relatedwork}

Generally, knowledge editing aims to edit the knowledge of a language model so that its outputs reflect the revised state when presented with relevant inputs \cite{de2021editing}. \citet{yao2023editing} categorized knowledge editing methods into two major classes: preserving model's parameters and modifying model's parameters. 

Methods for preserving model's parameters include memory-based methods and additional parameters' methods. Memory-based methods utilize external storage to store editing facts, for example, SERAC \cite{mitchell2022memory} employs an additional network to store editing knowledge, whereas GRACE \cite{hartvigsen2022aging} utilizes a codebook to store editing knowledge. Additional parameters' methods employ extra neurons to store editing facts, for instance, \citet{huang2023transformer} and \citet{dong2022calibrating} adding extra neurons in MLP to memorize additional facts. 

Since our target is to edit knowledge by updating the internal parameters of language models, this paper focuses on methods that modify model's parameters. Currently, methods for modifying model's parameters can be further divided into two categories: meta-learning and locate-and-edit.

\textbf{Meta-learning} methods use a hyper-network, and subsequently apply this hyper-network to edit language models. For instance, \citet{de2021editing} employed a bidirectional LSTM to predict weight updates for editing, \citet{mitchell2021fast} utilized low-rank decomposition of gradients to learn fine-tuning for language models, and \citet{tan2023massive} extended single editing to batch editing using a least-squares approach built upon MEND \cite{mitchell2021fast}.

\textbf{Locate-and-edit} methods first identify parameters corresponding to specific knowledge and achieve knowledge editing by updating these parameters. For example, \citet{dai2021knowledge} used knowledge attribution to determine the location of neurons, followed by parameter updates on these neurons for knowledge editing. \citet{meng2022locating} employed causal mediation analysis to identify the center of causal effects and performed updates on that position. \citet{meng2022mass} extended upon ROME \cite{meng2022locating} by distributing residuals across multiple layers and achieved batch editing, and \citet{li2023pmet} achieved more precise residual allocation.

However, existing knowledge editing methods that modify model's parameters mostly focus on single editing, unable to meet the demands of lifelong editing, leading to a certain gap between knowledge editing and practical applications. Although some current research focused on lifelong editing, such as \citet{yin2024history} focusing on temporal editing abilities in lifelong editing, \citet{hartvigsen2022aging} and \citet{huang2023transformer} developing knowledge editing methods that preserve model's parameters for lifelong editing (as mentioned earlier), the reasons for the failure of knowledge editing methods that modify model's parameters in lifelong editing lack exploration, resulting in a lack of core insights for further developing effective knowledge editing methods. Consequently, research into lifelong editing is imperative.

\section{Preliminary}\label{preliminary}

The language model $f_{\theta} \in \mathcal{F}$ can be defined as a function $f_{\theta}: \mathcal{X} \mapsto \mathcal{Y}$, mapping input $\pmb x \in \mathcal{X}$ to its prediction $\pmb y \in \mathcal{Y}$. For an editing example $(\pmb x_e, \pmb y_e)$, where $f_{\theta}(\pmb x_e) \neq \pmb y_e$, the goal of the knowledge editing (KE) is to edit the parameters $\theta \in \Theta$ of the model $f_{\theta}$ to obtain an edited model $f_{\theta'}$, such that $f_{\theta'}(\pmb x_e) = \pmb y_e$.

\begin{equation}
KE:\mathcal F\times \mathcal X\times\mathcal Y \mapsto\mathcal F
\end{equation}

In lifelong editing, such a process continues iteratively. In other words, for an initial language model $f_{\theta_0}$, there exists a potential sequence to be edited ${ (\pmb x_{e_i}, \pmb y_{e_i}) }_{i=1}^n$, and the model undergoes continuous editing:

\begin{equation}
f_{\theta_ i} = KE(f_{\theta_{i-1}},\pmb x_{e_i},\pmb y_{e_i})
\end{equation}

In lifelong editing, the edited model should satisfy the following properties.

\noindent\textbf{Effectiveness}: The edited model should produce the expected predictions.
    \begin{equation}
        f_{\theta_ i}(\pmb x_{e_i})=\pmb y_{e_i}
    \end{equation}
\textbf{Generality}: The edited model should remain consistent on its edited data equivalent input set $\mathcal{E}(\pmb x_{e_i})$.
    \begin{equation}
        f_{\theta_ i}(\pmb x_j)=\pmb y_{e_i},\forall \pmb x_j\in\mathcal E(\pmb x_{e_i})
    \end{equation}
\textbf{Locality}: The edited model should maintain the original output on data unrelated to the editing, denoted as $\mathcal{I}(\pmb x_{e_i})$.
    \begin{equation}
        f_{\theta_ i}(\pmb x_j)=\pmb y_{\pmb x_{j}},\forall\pmb x_j\in \mathcal I(\pmb x_{e_i})
    \end{equation}
\textbf{Retention}: The edited model should preserve the editing results based on the previously completed edits.
    \begin{equation} \label{equ:retention}
        f_{\theta _i}(\pmb x_{e_j})=\pmb y_{e_j}',\forall 1\le j<i
    \end{equation}
Here is \(\pmb y_{e_j}'\) rather than \(\pmb y_{e_j}\) because we consistently adhere to a principle: the later the edit, the higher the priority. Later edits take precedence over earlier ones and potentially engage in complex interactions with the original knowledge to update it. For further explanations and details, please refer to Appendix~\ref{sec:appendix_experimental_details}.

\begin{figure*}
  \centering
  \subfigure[L2 norm over steps on predefined layer.]{\includegraphics[width=0.35\textwidth]{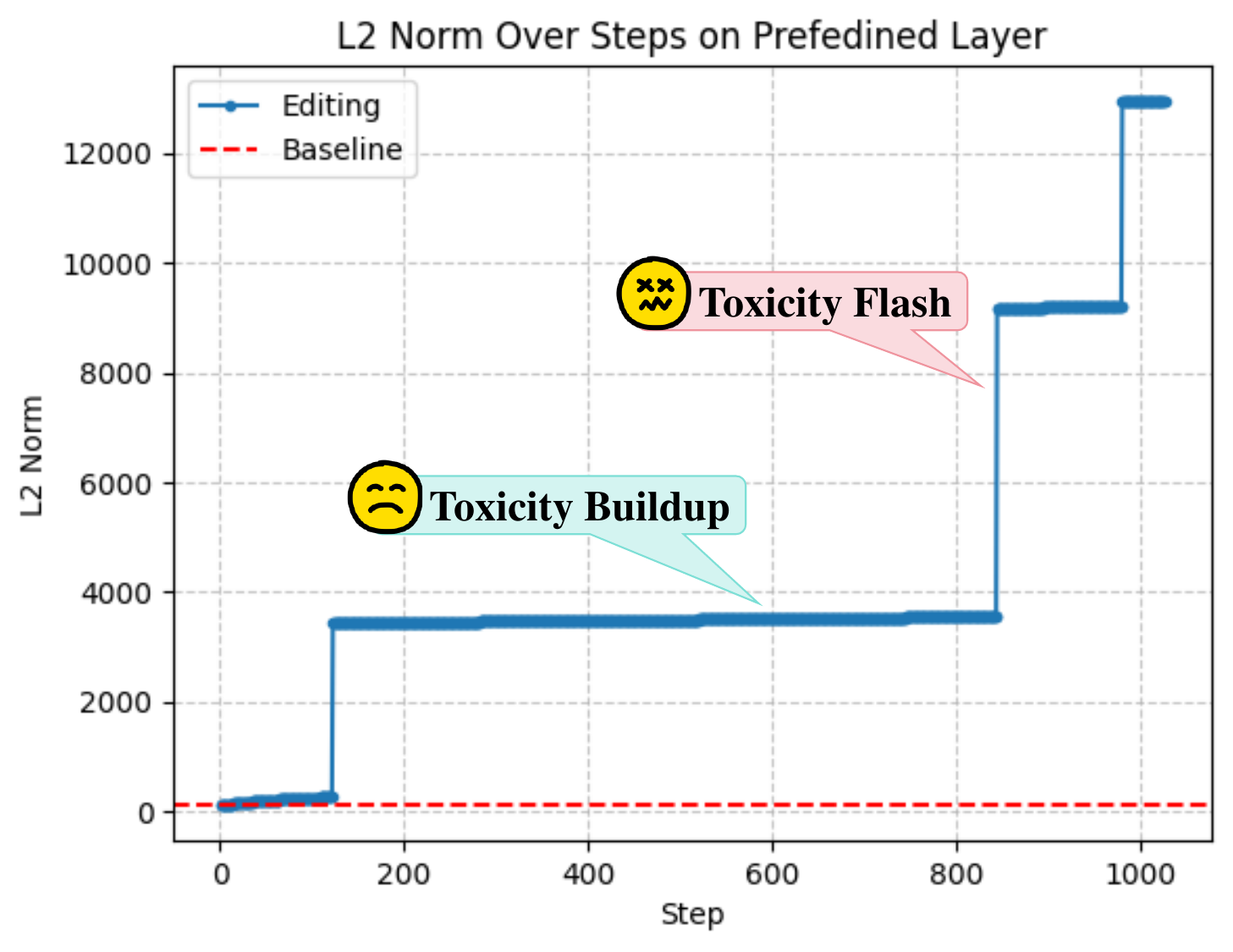}\label{fig:toxicity_gpt2-xl_a}
  }
   \hfill
  \subfigure[
Visualization of toxicity at specific steps. Darker color, larger changes.]{\includegraphics[width=0.59\textwidth]{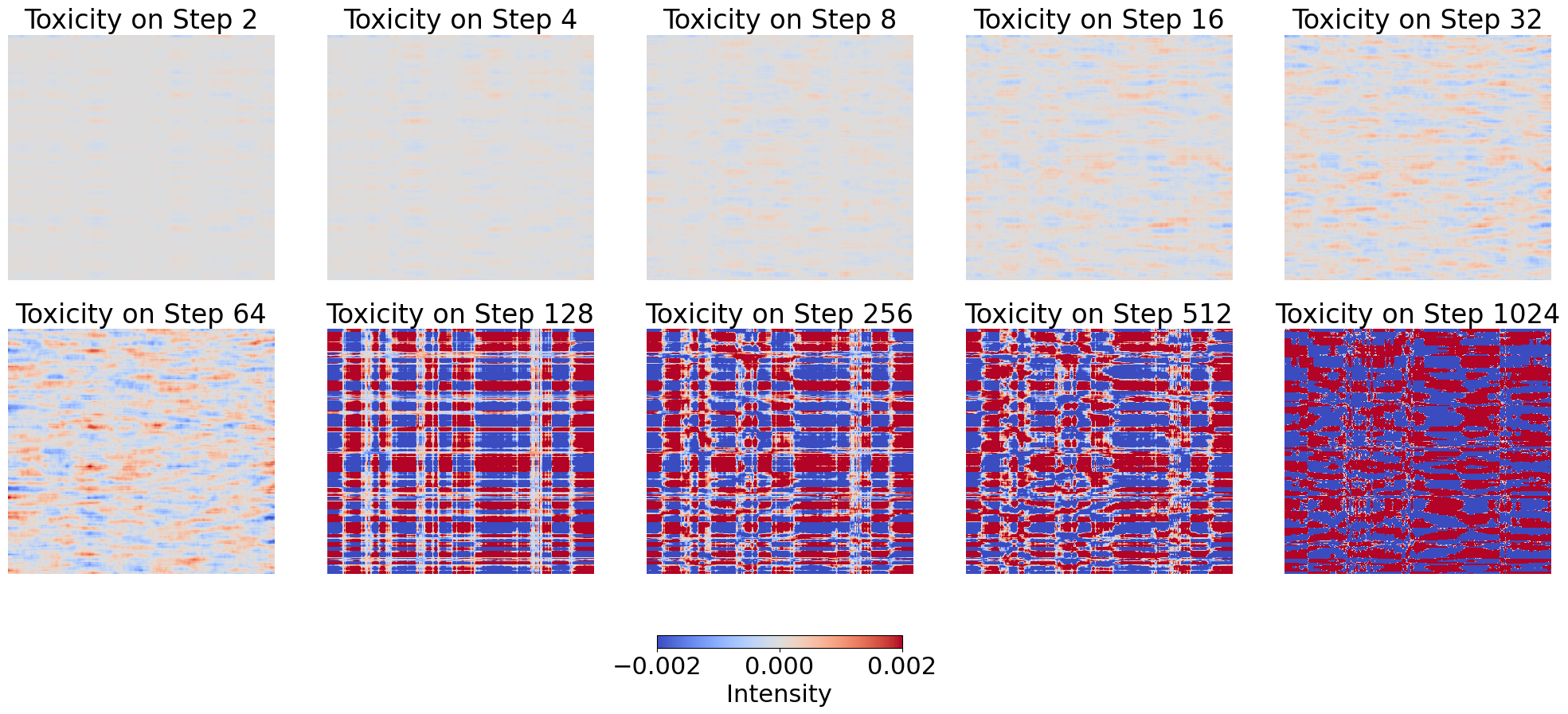}\label{fig:toxicity_gpt2-xl_b}
}

  \caption{The toxicity on GPT2-XL with editing steps.}
  \label{fig:toxicity_gpt2-xl}
\end{figure*}

\begin{figure*}
  \centering
  \subfigure[L2 norm over steps on predefined layer.]{\includegraphics[width=0.35\textwidth]{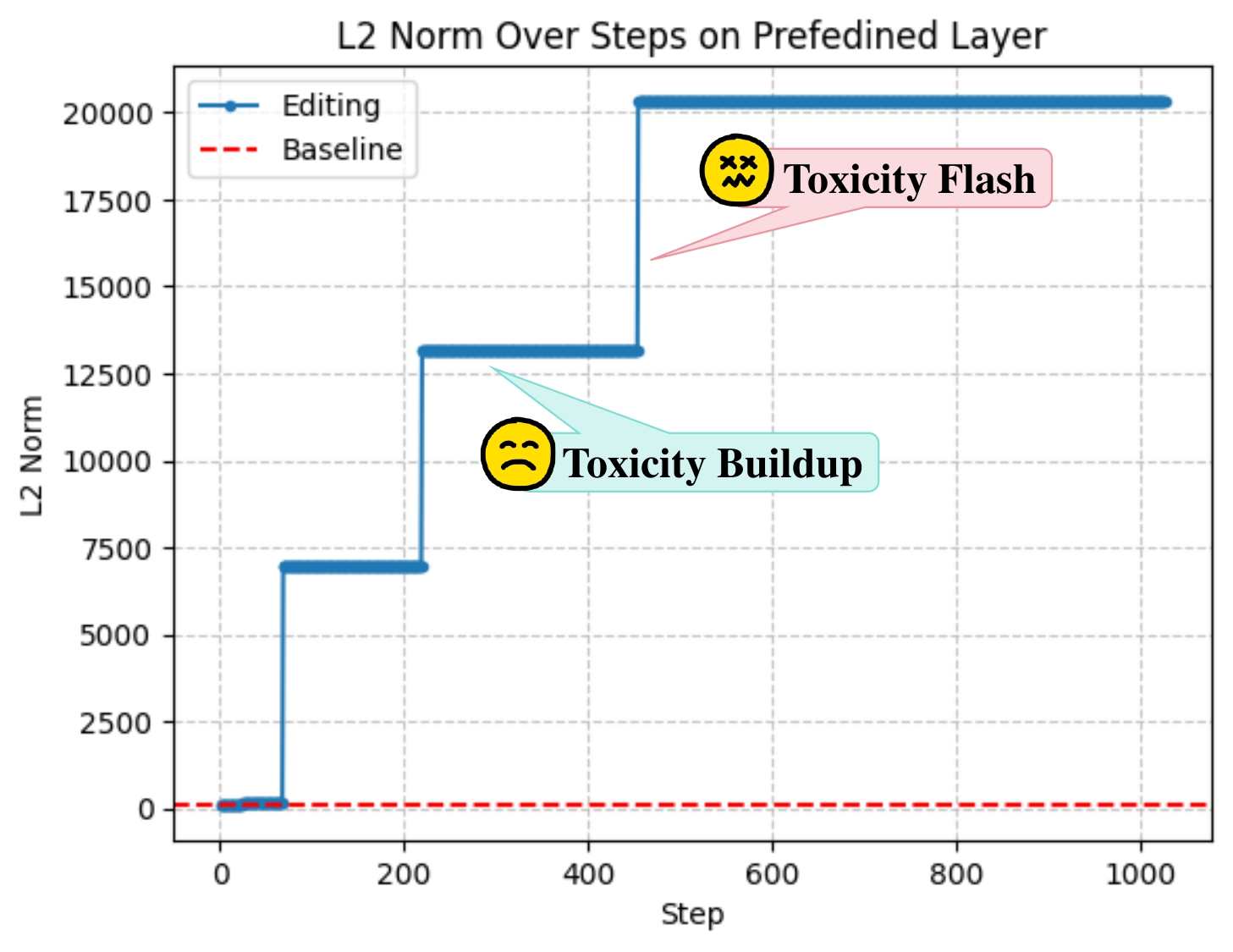}\label{fig:toxicity_gpt-j_a}
  }
   \hfill
  \subfigure[
Visualization of toxicity at specific steps. Darker color, larger changes.]{\includegraphics[width=0.59\textwidth]{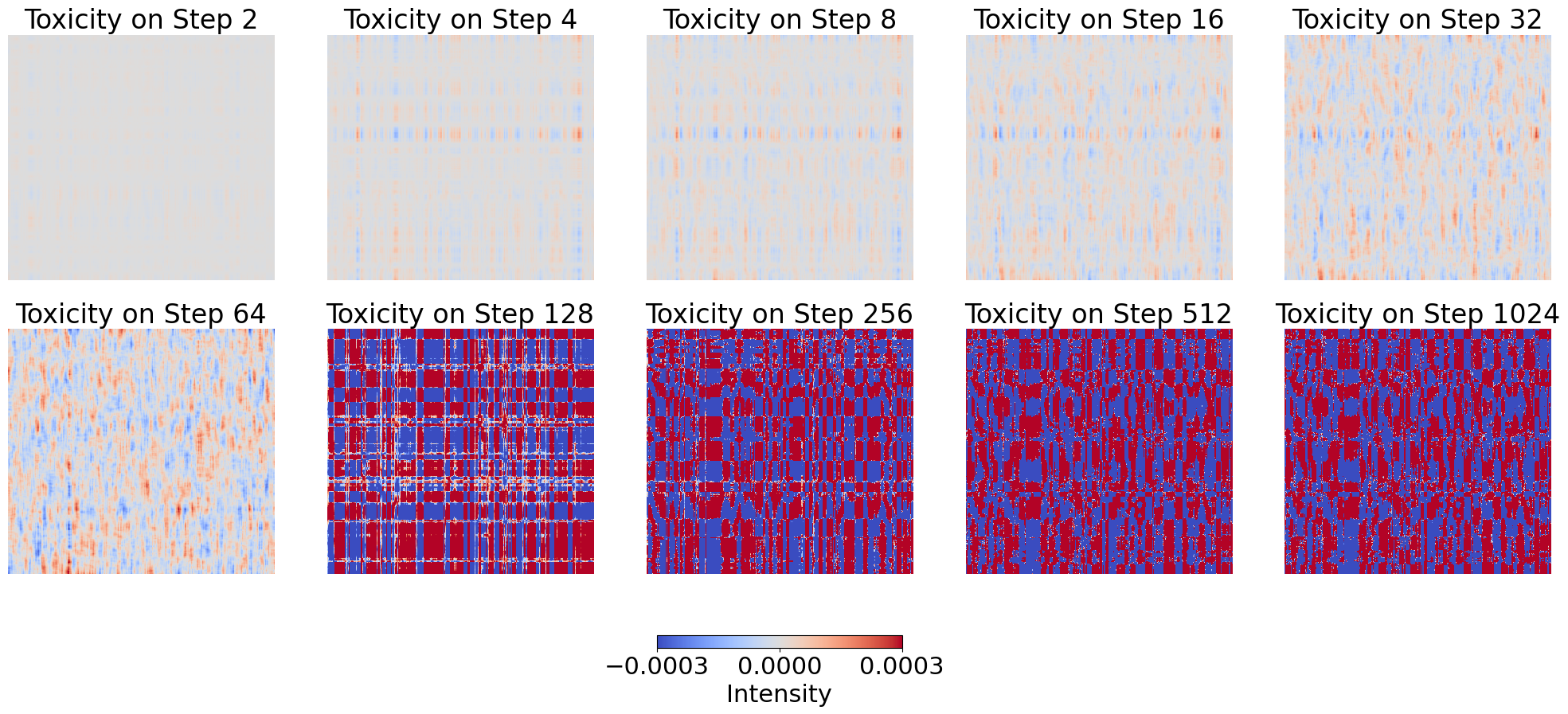}\label{fig:toxicity_gpt-j_b}
}

  \caption{The toxicity on GPT-J with editing steps.}
  \label{fig:toxicity_gpt-j}
\end{figure*}

\section{Toxicity in Lifelong Editing} \label{toxicity_in_lifelong_editing}

ROME \cite{meng2022locating} and MEMIT \cite{meng2022mass} are currently the state-of-the-art knowledge editing methods. As MEMIT is based on ROME, implementing residual distribution across multiple layers, our analysis in the main text focuses primarily on ROME. The analysis of MEMIT is provided in Appendix~\ref{sec:appendix_toxicity_analysis_on_memit}. In this section, we systematically investigate the reasons for the failure of ROME in lifelong editing.

\subsection{Toxicity} \label{toxicty}

As editing progresses, the performance of the language model continuously deteriorates \cite{yao2023editing}, indicating that ongoing editing seems to introduce certain side effects. In this section, we refer to these side effects as "\textbf{toxicity}" and utilize rollback editing \cite{li2023unveiling} to define toxicity:

\begin{equation}
\begin{gathered}
    Toxicity =\theta^*-\theta\\
    \text{s.t.}\ f_{\theta^*}=KE(KE(f_{\theta},\pmb x_e,\pmb y_e),\pmb x_e,\pmb y_o),
\end{gathered}
\end{equation}
where \(f_{\theta}(\pmb x_e)=\pmb y_o\). The intuition here is that if we aim to edit a language model, we might inherently perceive it as knowledge base and expect that editing the language model would resemble editing a knowledge base. Therefore, after rollback editing, we expect the language model to return to its initial state. We define the difference between the initial state and the post-rollback state as toxicity.

\begin{figure*}
  \centering
  \subfigure[L2 norm over steps on predefined layer.]{\includegraphics[width=0.35\textwidth]{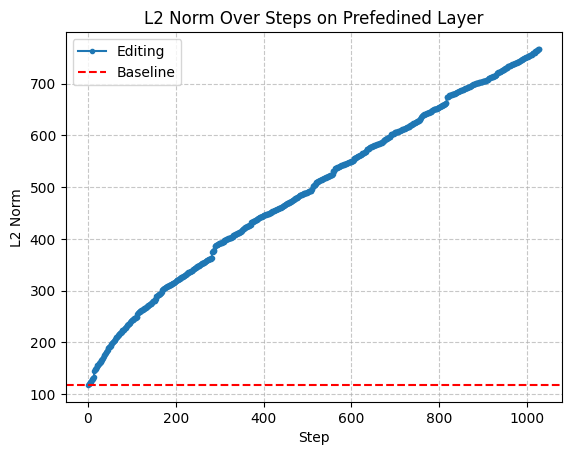}\label{fig:toxicity_buildup_gpt2-xl_a}
  }
   \hfill
  \subfigure[
Visualization of toxicity at specific steps. Darker color, larger changes.]{\includegraphics[width=0.59\textwidth]{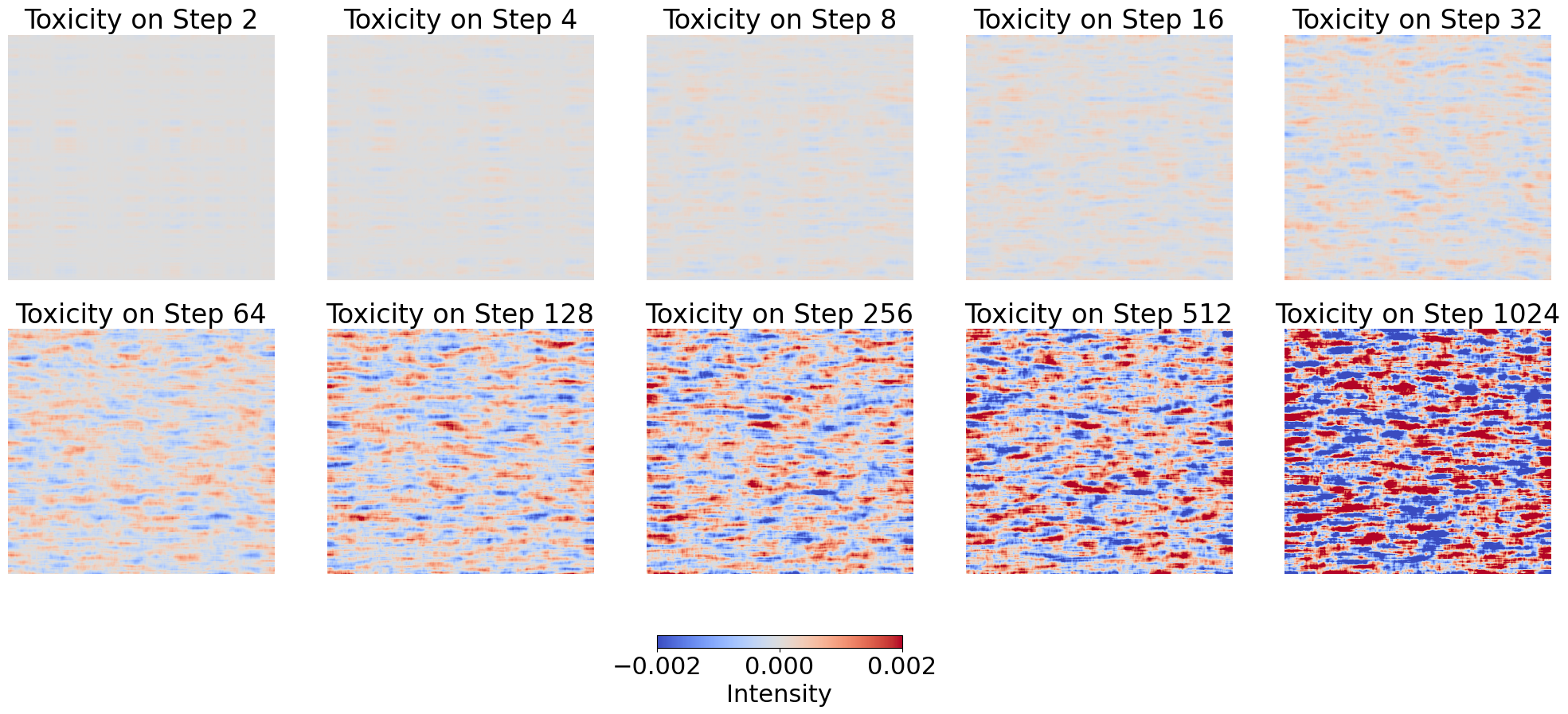}\label{fig:toxicity_buildup_gpt2-xl_b}
}

  \caption{Toxicity buildup on GPT2-XL with editing steps.}
  \label{fig:toxicity_buildup_gpt2-xl}
\end{figure*}

\begin{figure*}
  \centering
  \subfigure[L2 norm over steps on predefined layer.]{\includegraphics[width=0.35\textwidth]{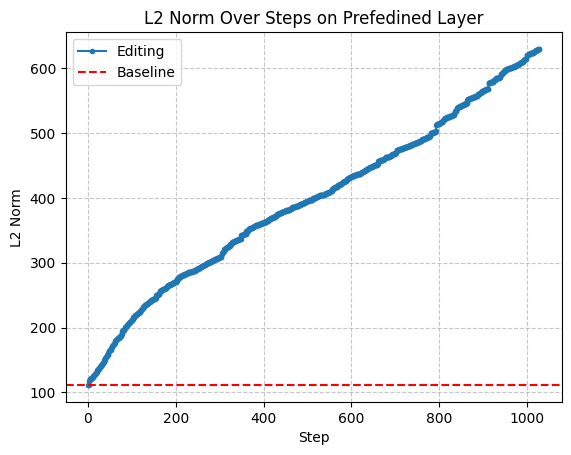}\label{fig:toxicity_builup_gpt-j_a}
  }
   \hfill
  \subfigure[
Visualization of toxicity at specific steps. Darker color, larger changes.]{\includegraphics[width=0.59\textwidth]{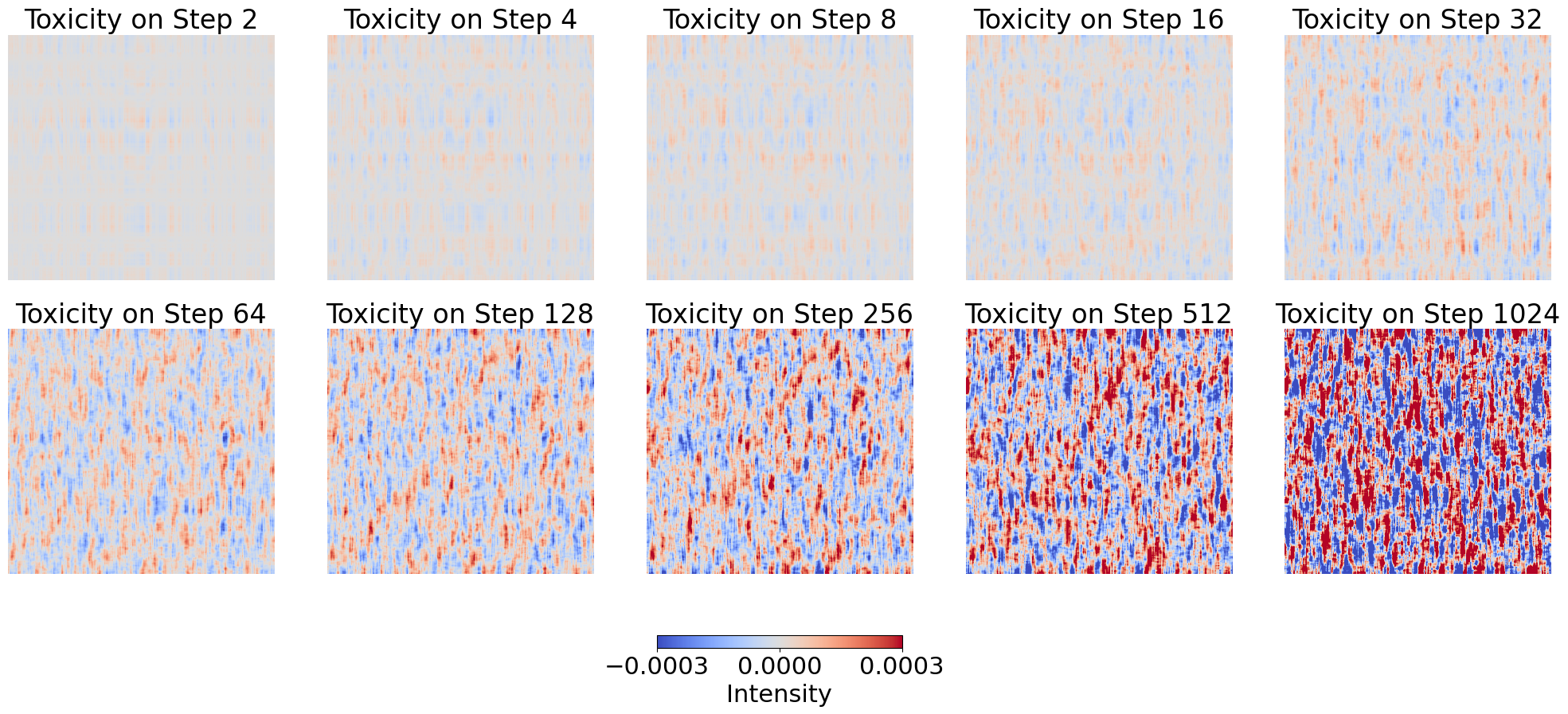}\label{fig:toxicity_builup_gpt-j_b}
}

  \caption{Toxicity buildup on GPT-J with editing steps.}
  \label{fig:toxicity_builup_gpt-j}
\end{figure*}

To better simulate real-world knowledge editing scenarios, we first filter data points corresponding to known knowledge in CounterFact dataset \cite{meng2022locating} for both GPT2-XL \cite{radford2019language} and GPT-J \cite{wang2021gpt}. Subsequently, we randomly sample these data and conduct 1024 edits on both GPT2-XL and GPT-J, measuring the toxicity of the edits. As depicted in Figure~\ref{fig:toxicity_gpt2-xl_a},~\ref{fig:toxicity_gpt-j_a}, the red dashed line represents the L2 norm of the original parameters on the predefined editing layer, while the blue solid line represents the L2 norm of the actual parameters on the predefined editing layer as editing progresses. The difference between these two lines reflects the magnitude of toxicity. Figure~\ref{fig:toxicity_gpt2-xl_b},~\ref{fig:toxicity_gpt-j_b} visualizes the accumulated toxicity at specific steps, such as steps 2, 4, 8, 16, 32, 64, 128, 256, 512, and 1024, corresponding to specific positions in Figure~\ref{fig:toxicity_gpt2-xl_a},~\ref{fig:toxicity_gpt-j_a}, to illustrate the toxicity status at these steps.

The experimental results indicate that toxicity accumulates throughout the editing process, a phenomenon we term "\textbf{toxicity buildup}." Additionally, "spikes" in toxicity are observed at certain data points, which we term "\textbf{toxicity flash}." Consequently, the overall measurement exhibits a staircase shape. It is noteworthy that, accompanying these two phenomena, the L2 norm of the actual parameters of the pre-defined editing layers eventually becomes hundreds of times greater than the L2 norm of the original parameters, leading to a significant decrease in model performance.

Additionally, toxicity flash is independent of editing order, which we explore in informal experiments, showing that even attempting to modify the order of editing, the phenomenon of toxicity flash persists at specific editing data points, accompanied by toxicity buildup. Further investigation reveals that the data points causing toxicity flash exhibit the same "spike" phenomenon even in single editing. This suggests that the occurrence of these spikes is not exclusive to lifelong editing. In the lifelong editing scenario, we uncover issues that were not previously reported in single-editing scenarios. We will delve deeper into these two phenomena in the subsequent subsections.

\begin{figure*}
    \centering
    \subfigure[L2 norm over layers on case 3561.]{\includegraphics[width=0.23\textwidth]{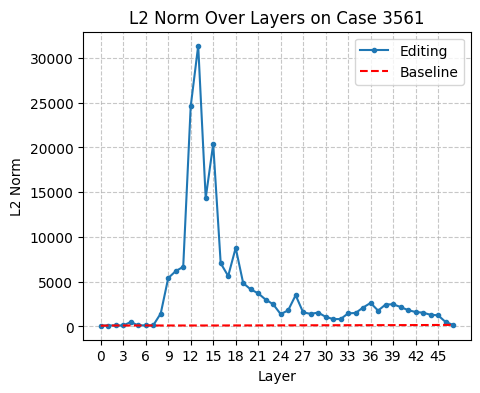}\label{fig:toxicity_flash_gpt2-xl_a}
  }
    \hfill
    \subfigure[L2 norm over layers on case 8793.]{\includegraphics[width=0.23\textwidth]{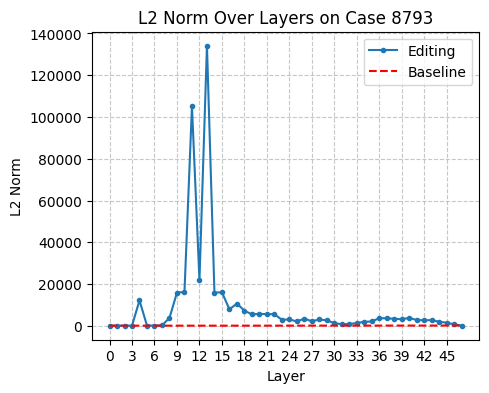}\label{fig:toxicity_flash_gpt2-xl_b}
}
    \hfill
    \subfigure[L2 norm over layers on case 16575.]{\includegraphics[width=0.23\textwidth]{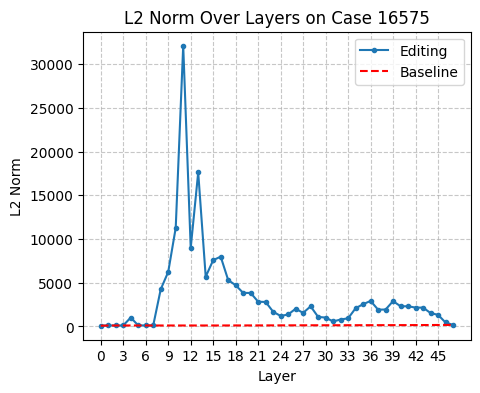}\label{fig:toxicity_flash_gpt2-xl_c}
}
    \hfill
    \subfigure[L2 norm over layers on case 16781.]{\includegraphics[width=0.23\textwidth]{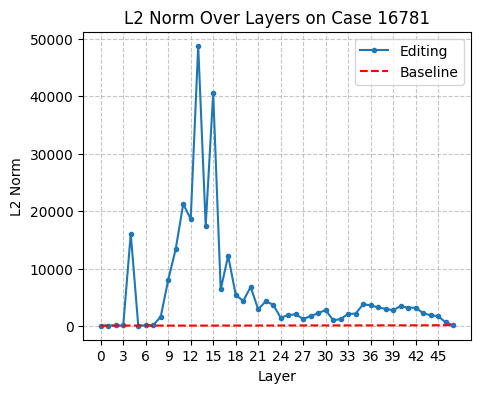}\label{fig:toxicity_flash_gpt2-xl_d}
}

  \caption{Toxicity flash on GPT2-XL among editing layers.}
  \label{fig:toxicity_flash_gpt2-xl}
\end{figure*}

\begin{figure*}
    \centering
    \subfigure[L2 norm over layers on case 3561.]{\includegraphics[width=0.23\textwidth]{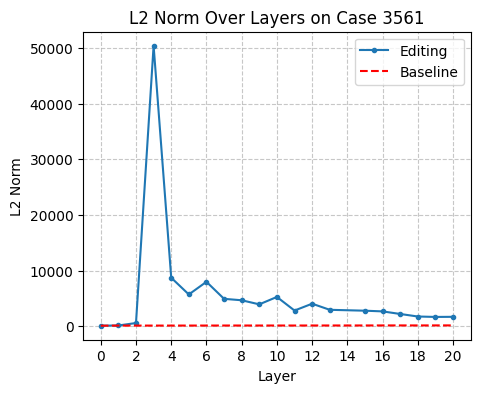}\label{fig:toxicity_flash_gpt-j_a}
  }
    \hfill
    \subfigure[L2 norm over layers on case 8793.]{\includegraphics[width=0.23\textwidth]{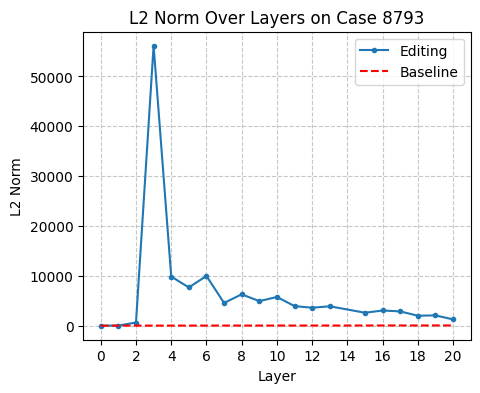}\label{fig:toxicity_flash_gpt-j_b}
}
    \hfill
    \subfigure[L2 norm over layers on case 16575.]{\includegraphics[width=0.23\textwidth]{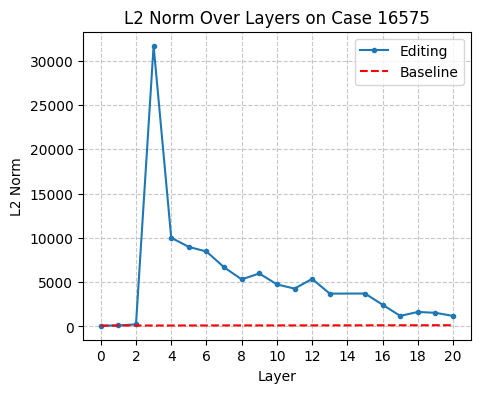}\label{fig:toxicity_flash_gpt-j_c}
}
    \hfill
    \subfigure[L2 norm over layers on case 16781.]{\includegraphics[width=0.23\textwidth]{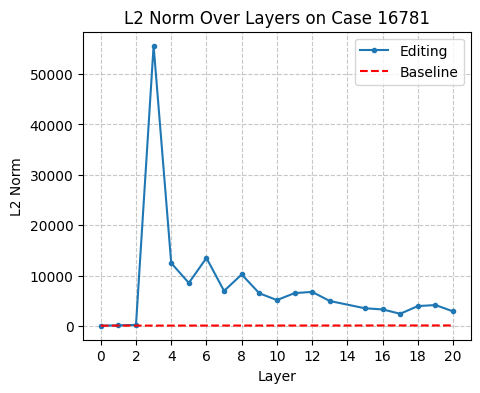}\label{fig:toxicity_flash_gpt-j_d}
}

  \caption{Toxicity flash on GPT-J among editing layers.}
  \label{fig:toxicity_flash_gpt-j}
\end{figure*}

\subsection{Toxicity Buildup} \label{toxicity_buildup}

Based on the data corresponding to known knowledge, we filter out the data causing toxicity flash during editing on GPT2-XL and GPT-J. Details of the filtering process and the results are described in Appendix~\ref{sec:appendix_toxicity_buildup_and_toxicty_flash_data_spliter}. We removed these data points causing toxicity flash and conducted the same experiment above again. As shown in Figure~\ref{fig:toxicity_buildup_gpt2-xl_a},~\ref{fig:toxicity_builup_gpt-j_a}, the red dashed line represents the L2 norm of the original parameters of the predefined editing layer, and the blue solid line represents the L2 norm of the actual parameters of the predefined editing layer as editing progresses. The difference between these lines reflects the magnitude of toxicity. Figure~\ref{fig:toxicity_buildup_gpt2-xl_b},\ref{fig:toxicity_builup_gpt-j_b} visualizes the accumulated toxicity at specific steps.

From the experimental results, it is evident that after filtering out data causing toxicity flash, the magnitude of toxicity significantly decreases. Moreover, as shown in Figure~\ref{fig:toxicity_buildup_gpt2-xl_b},~\ref{fig:toxicity_builup_gpt-j_b}, the process of toxicity buildup becomes more uniform and gradual than Figure~\ref{fig:toxicity_gpt2-xl_b},~\ref{fig:toxicity_gpt-j_b}. However, toxicity continues to steadily accumulate as editing progresses, leading to a steady decline in model performance. Additionally, this observation may suggest that editing results in the disruption of superposition (\citealp{elhage2022toy}; \citealp{henighan2023superposition}) and polysemantic neurons \cite{elhage2022softmax} in the original model, which could be important factors contributing to the continuous decline in models' performance during the editing process.

\subsection{Toxicity Flash} \label{toxicity_flash}

Subsequently, we focus on the data causing toxicity flash during editing. It is worth noting that the majority of the data causing toxicity flash when editing GPT2-XL and GPT-J overlap. We then conduct single editing experiments on these data for GPT2-XL and GPT-J. Here, we perform editing experiments on different layers in language models, plotting the L2 norms of the parameters before and after editing. The experimental results are illustrated in Figures~\ref{fig:toxicity_flash_gpt2-xl},~\ref{fig:toxicity_flash_gpt-j}, where the red dashed line represents the L2 norm of the original parameters of different layers in language modelS, and the blue solid line represents the L2 norm of the actual parameters after single editing on different layers. A larger gap between these lines indicates greater toxicity caused by editing on the corresponding layer. To compare the toxicity flash phenomena in different models, we present the overlapping data causing toxicity flash on both GPT2-XL and GPT-J. Further experiments on toxicity flash data for GPT2-XL and GPT-J, as well as comparisons with experiments on other regular data, can be found in Appendix~\ref{sec:appendix_more_edit_analysis_on_toxicity_flash}.

ROME's predefined editing layers on GPT2-XL and GPT-J are 17 and 5, respectively, where \citet{meng2022locating} described these as the center of causal effects, which has been further utilized in MEMIT \cite{meng2022mass}. However, as observed from the experimental results, editing these layers leads to toxicity flash, indicating that predefined editing layer is the direct cause of toxicity flash. From Figures~\ref{fig:toxicity_flash_gpt2-xl},~\ref{fig:toxicity_flash_gpt-j}, it can be inferred that for these data points, we should edit the layer that does not align with the predefined editing layer. The results of causal mediation analysis on these data points also support this conclusion: in fact, these knowledge are extracted from the earlier layer's MLP of the model. For detailed experimental results, please refer to Appendix~\ref{sec:appendix_causal_mediation_analysis_on_gpt2-xl}.

\subsection{Pattern Unmatch}

After further investigation, the fundamental reason lies in pattern unmatch. Actually, pattern match and unmatch are relative concepts. As depicted in Figure~\ref{fig:frame}, for the input "Thor is affiliated with the," there might not be any information extracted related to "Avengers" at layer \(i+1\), but rather, the primary information about "Avengers" might be extracted at layer \(i-2\). Therefore, layer \(i-2\) effectively detects the pattern leading to the target output from the input, which is \(\pmb k^{i-2}\), representing as the key in key-value memory. Extracting the main information about "Avengers" , which is the target output, from \(W_{proj}^{i-2}\) based on \(\pmb k^{i-2}\) and placing it in the residual flow constitutes what we refer to as pattern match. Otherwise, it is pattern unmatch. Thus, the ability to retrieve target information from $W_{proj}$ based on $\pmb k$ serves as the criterion for determining whether the pattern matches.

In summary, the patterns of editing knowledge may not be detected in the predefined editing layer, which we call pattern unmatch. Continuing to edit on such knowledge will lead to language model overfitting, resulting in toxicity flash. The investigation details of pattern unmatch and experimental evidence are provided in Appendix~\ref{sec:appendix_pattern_unmatch}.

\begin{table*}[ht]
\begin{footnotesize}
\setlength{\abovecaptionskip}{0cm} 
\centering
\renewcommand\arraystretch{0.3}
\setlength{\tabcolsep}{3.05mm}{
\begin{tabular}{llcccccc}
\toprule
\multirow{2}{*}{\textbf{ Model }} 
& \multirow{2}{*}{\textbf{ Editor }} 
& \multicolumn{1}{c}{\textbf { Score }} 
& \multicolumn{1}{c}{\textbf { Effectiveness }} 
& \multicolumn{1}{c}{\textbf { Generality }} 
& \multicolumn{1}{c}{\textbf { Locality }}
& \multicolumn{2}{c}{\textbf { Retention }} 
 \\
\cmidrule(r){3-3}
\cmidrule(r){4-4}
\cmidrule(r){5-5}
\cmidrule(r){6-6}
\cmidrule(r){7-8}

 & & $\mathrm{S} \uparrow$ & $\mathrm{ES} \uparrow$ & $\mathrm{PS} \uparrow$ & $\mathrm{NS} \uparrow$ & $\mathrm{ERS} \uparrow$ & $\mathrm{ORS} \uparrow$ \\
 \midrule[0.5pt]
 \text {GPT-2 XL} & \text {KE} &0.0(0.0) &0.1(0.0) &0.1(0.0) &0.0(0.0) &0.0(0.0) &0.0(0.0)
 \\
& \text {KN} &0.0(0.0) &0.1(0.0) &1.0(0.0) &0.0(0.0) &0.0(0.0) &0.0(0.0)
\\
& \text {MEND} &0.0(0.0) &0.5(0.0) &0.1(0.0) &0.4(0.0) &0.0(0.0) &0.0(0.0)
\\
& \text{ROME} &9.3(2.4) &15.8(4.4) &8.8(2.7) &6.8(1.4) &12.2(2.6) &7.9(2.6)
\\
& \text{MEMIT} &13.2(7.5) &\textbf{92.5(0.5)} &\textbf{55.1(0.5)} &\textbf{35.0(0.7)} &6.6(4.4) &6.6(4.5)
\\
\cmidrule(lr){2-8}
&  \textbf{WilKE(Ours)} &\textbf{19.3(5.6)} &70.7(9.2) &51.0(7.0) &12.7(0.9) &\textbf{16.1(6.4)} &\textbf{13.1(5.7)}
\\
  \midrule
 \text {GPT-J}  & \text {MEND} &3.7(3.0) &3.7(2.6) &1.7(1.0) &4.7(3.3) &9.8(8.0) &9.2(6.5)	
\\
& \text{ROME} &8.7(0.4) &28.7(1.0) &20.7(0.7) &4.6(0.3) &10.9(2.7) &5.8(0.6)
 \\
& \text{MEMIT} &0.0(0.0) &32.1(3.6) &23.8(2.2) &\textbf{9.3(1.1)}&0.0(0.0) &0.0(0.0)
 \\
 \cmidrule(lr){2-8}
&  \textbf{WilKE(Ours)} &\textbf{14.6(2.6)} &\textbf{71.3(6.5)} &\textbf{50.6(6.4)} &7.4(0.9) &\textbf{19.1(8.0)} &\textbf{8.5(1.9)}
\\
\bottomrule
\end{tabular}}
\vspace{2mm}
\caption{Evaluation results (\%) with 95\% confidence intervals in parentheses.}
\label{tab:main_results}
\end{footnotesize}
\end{table*}

\section{Wise-Layer Knowledge Editor} \label{wiselayer_konwledge_editor}

In Section~\ref{toxicity_in_lifelong_editing}, we delved into the primary reason for failure in lifelong editing - pattern unmatch, which directly leads to toxicity flash and potentially more toxicity buildup. In light of this, we propose an editing method called WilKE. Unlike ROME and MEMIT, WilKE does not predefine editing layer; instead, it selects editing layer based on the degree of pattern match for different editing knowledge across various layers. We first describe where to edit in Section~\ref{where_to_edit}, followed by an explanation of how to edit in Section~\ref{how_to_edit}.

\subsection{Where to Edit?} \label{where_to_edit}

To implement knowledge editing, the initial step involves determining the locations where editing will take place.

\citet{meng2022locating} utilizes causal mediation analysis to identify the center of causal effects, MLP at specific layer, for storing factual knowledge. The MLP of the FFN is divided into two matrices, represented as follows:

\begin{equation}
    FFN^l(\pmb x)=\sigma(\pmb x\cdot W_{fc}^l)\cdot W_{proj}^l,
\end{equation}
where \(W_{fc}^l\in\mathbb R^{d\times d_m}\) and \(W_{proj}^l\in\mathbb R^{d_m\times d}\) are the parameter matrices of the \(l\)-th layer's FFN, \(FFN^l\), and \(d_m\) is the dimension of the intermediate hidden layer in the FFN. The symbol \(\sigma\) represents the activation function, and \(\pmb x\in\mathbb R^d\) is the input to the FFN. As described in the key-value memories (\citealp{sukhbaatar2015end}; \citealp{sukhbaatar2019augmenting}; \citealp{geva2020transformer}), \(W_{fc}^l\) identifies patterns of the input \(\pmb{x}\) to obtain the key vector \(\pmb k^l\), and then the value vector \(\pmb v^l\) is retrieved from \(W_{proj}^l\), as shown in Figure~\ref{fig:frame}. Therefore, to achieve knowledge editing, we modify \(W_{proj}^l\). After identifying the component that needs modification, we further determine the specific layer for modifying this component.

To find the editing layer \( l^* \), our initial intuition is to identify the layer that produces the maximum activation strength for a specific knowledge, which is represented as \( \text{argmax}_l\ \|\sigma(\pmb x\cdot W_{fc}^l)\|_2 \). However, in practice, the optimization of \( \pmb\delta^l \) \cite{meng2022locating} after the FFN for aligning model's output to achieve knowledge updating varies across different layers, as detailed in Appendix~\ref{sec:appendix_delta_strength}. Specifically, \(\pmb \delta^l\) is calculated as follows:

\begin{equation}\label{equ:opt_delta}
\begin{gathered}
    \pmb \delta^l=\pmb v^{l}_*-W_{proj}^l\pmb k^l, \pmb v^{l}_*=\text{argmin}_{\pmb z}\mathcal L(\pmb z)\\
    \text{where } \mathcal L(\pmb z)=\frac{1}{N}\sum_{j=1}^N-\log\mathbb P_{f_{\theta}(\pmb m_i^l:=\pmb z)}[o^*|c_j+p]\\
    +D_{KL}(\mathbb P_{f_{\theta}(\pmb m_i^l:=\pmb z)}[y|p']||\mathbb P_{f_\theta}[y|p'])
\end{gathered}
\end{equation}
where \(\pmb m_i^l\) represents the output of the MLP in the \(l\)th layer on the \(i\)th token, which is the end of the subject, \(c_j\) denotes a token sequence randomly generated to simulate irrelevant context, \(p\) corresponds to the knowledge we intend to edit, and \(p'\) represents the essence of the subject. In simple terms, the first term in \(\mathcal L(\pmb z)\) is for knowledge updating, while the second term is for maintaining an understanding of the essence of the subject. Further details can be referenced in \citet{meng2022locating}.

As mentioned before, the importance of the hidden state outputted by different layers for editing specific knowledge is actually different. To comprehensively consider these two points, namely the activation strength of specific knowledge and the \(\pmb \delta^l\) optimized for knowledge editing, we define the editing layer (\textbf{wise-layer}) \( l^* \) as follows:

\begin{equation} \label{equ:wise_layer}
    l^*=\text{argmin}_l\ \left\|\frac{\pmb\delta^l}{\|W_{proj}^l\|_2\ \sigma(\pmb x\cdot W_{fc}^l)}\right\|_2
\end{equation} 
where the term \( \|W_{proj}^l\|_2 \) in the denominator can be regarded as normalization, allowing for comparison across layers. Ultimately, we determine that the target editing location is \(W_{proj}^{l^*}\) of the layer \(l^*\).

\subsection{How to Edit?} \label{how_to_edit}    

After determining the target editing location, the next step involves determining how to carry out an edit.

Same as Equation~\ref{equ:opt_delta}, we introduce a residual term \(\pmb{\delta}^{l^*} \in \mathbb{R}^d\) to the output of the FFN in editing layer \(l^*\), denoted as \(FFN^{l^*}(\pmb{x})+\pmb\delta^{l^*}\). We optimize this residual term to align the model's output with our expected results while not affecting irrelevant knowledge.

Subsequently, we allocate the optimized residual term $\pmb{\delta}^{l*}$ to $W_{proj}^{l^*}$ to accomplish knowledge editing:

\begin{equation} \label{equ:update}
    W_{proj}^{l^*}\leftarrow\frac{FFN^{l^*}(\pmb x)+\pmb \delta^{l^*}}{\sigma(\pmb x\cdot W_{fc}^{l^*})}
\end{equation}

Afterwards, we have completed one editing. In summary, our approach starts from the perspective of pattern matching, attempting to identify the layer that is most suitable for editing the given knowledge across all layers, and then performs knowledge editing on that location.

\section{Experiments} \label{experiments}

\subsection{Experimental Setting} \label{experimental_setting}

\noindent\textbf{Model} We utilize two widely employed autoregressive language models for knowledge editing: GPT-XL (1.5B) \cite{radford2019language} and GPT-J (6B)  \cite{wang2021gpt}. 

\noindent\textbf{Baselines} Regarding knowledge editing methods, we select the following approaches: KnowledgeEditor (KE) \cite{de2021editing} utilizes a bidirectional LSTM to predict weight updates for editing data points; KnowledgeNeuron (KN) \cite{dai2021knowledge} employs knowledge attribution to determine the positions of neurons, followed by parameter updates on these neurons to implement knowledge updates; MEND \cite{mitchell2021fast} uses low-rank decomposition of gradients to learn fine-tuning of language models; ROME \cite{meng2022locating} employs causal mediation analysis to identify the center of causal effects, followed by gradient descent parameter updates on the MLP at that layer; MEMIT \cite{meng2022mass} extends upon ROME by distributing residuals across multiple layers.

\noindent\textbf{Datasets, Metrics and Experiment Details} Due to space limitations, details of dataset, metrics, and experimental details are provided in Appendix~\ref{sec:appendix_experimental_details} for reference. 

\subsection{Main Results} \label{main_results}

As shown in Table~\ref{tab:main_results}, we present the knowledge editing results after 1024 edits on GPT-XL and GPT-J. The results indicate that current knowledge editing methods perform poorly in lifelong editing, far from the optimistic results reported in single editing. However, these methods have been directly transferred and used in many other scenarios (\citealp{ma2023untying}; \citealp{li2023pmet}; \citealp{anonymous2024badedit}; \citealp{wang2023easyedit}).

WilKE demonstrates the most advanced comprehensive performance relative to the current knowledge editing methods. Specifically, under the same experimental conditions on GPT2-XL and GPT-J, WilKE achieves an average performance improvement of 46.2\% and 67.8\%, respectively, relative to the state-of-the-art methods. 

To gain further insight, we have plotted the complete performance curves, and detailed results are presented in Appendix~\ref{sec:appendix_complete_performance_curves}.

\subsection{Ablation Study} \label{ablation_study}

Since the core of our method lies in selecting different editing layers based on various knowledge, as demonstrated in Equation~\ref{equ:wise_layer} in Section~\ref{where_to_edit}, we comprehensively consider three aspects: the optimization of $\pmb \delta$ for editing knowledge across different layers, the activation of specific knowledge across different layers $\sigma(\pmb x\cdot W_{fc}^l)$, and the $\|W_{proj}^l\|_2$ across different layers. Therefore, we sequentially ablate these three factors to demonstrate that considering these three factors collectively leads to a better editing layer.

\begin{figure}
    \centering
    \subfigure[Score with editing steps on GPT2-XL. ]{\includegraphics[width=0.4\textwidth]{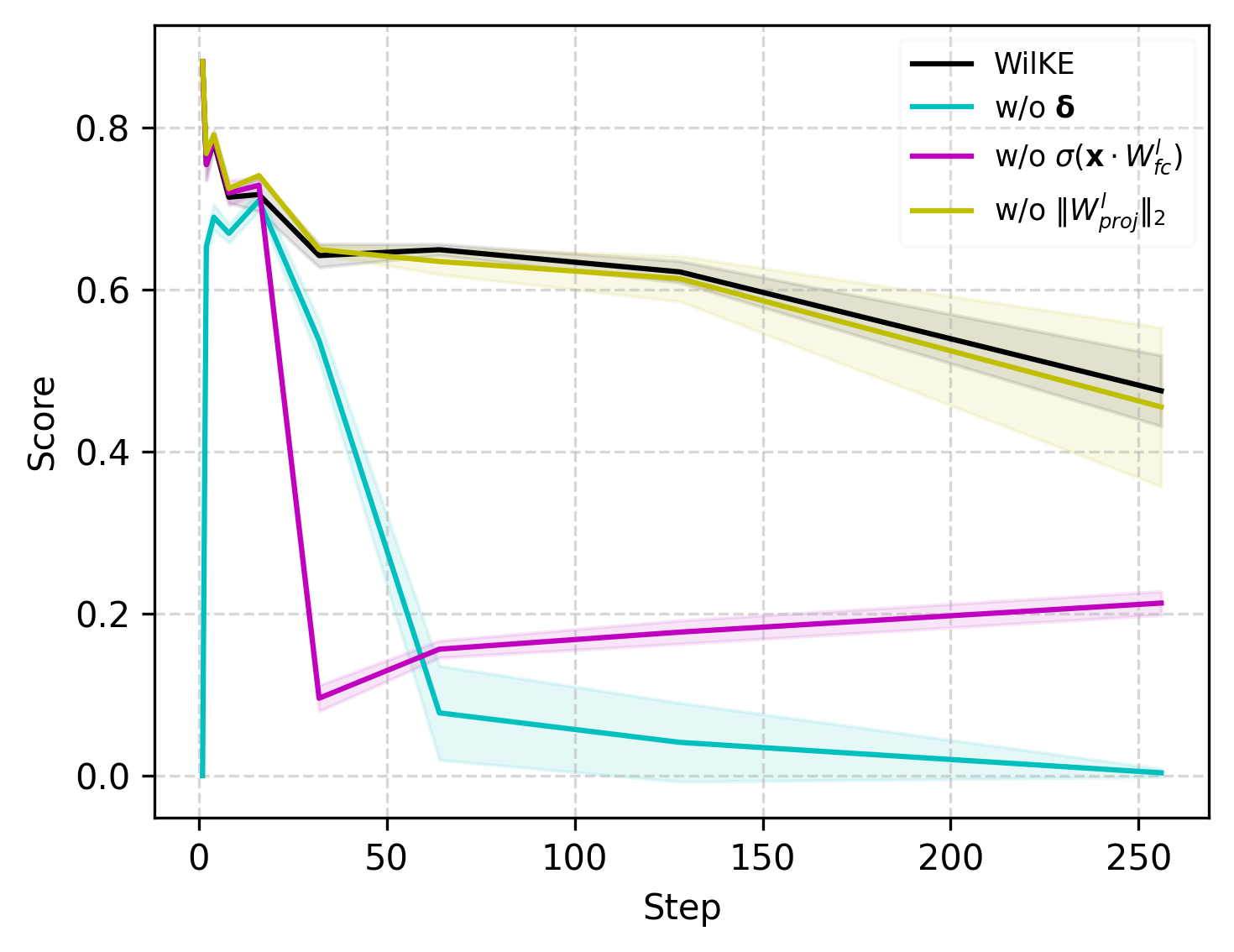}\label{fig:ablation_study_gpt2-xl}
  }
    \hfill
    \subfigure[Score with editing steps on GPT-J.]{\includegraphics[width=0.4\textwidth]{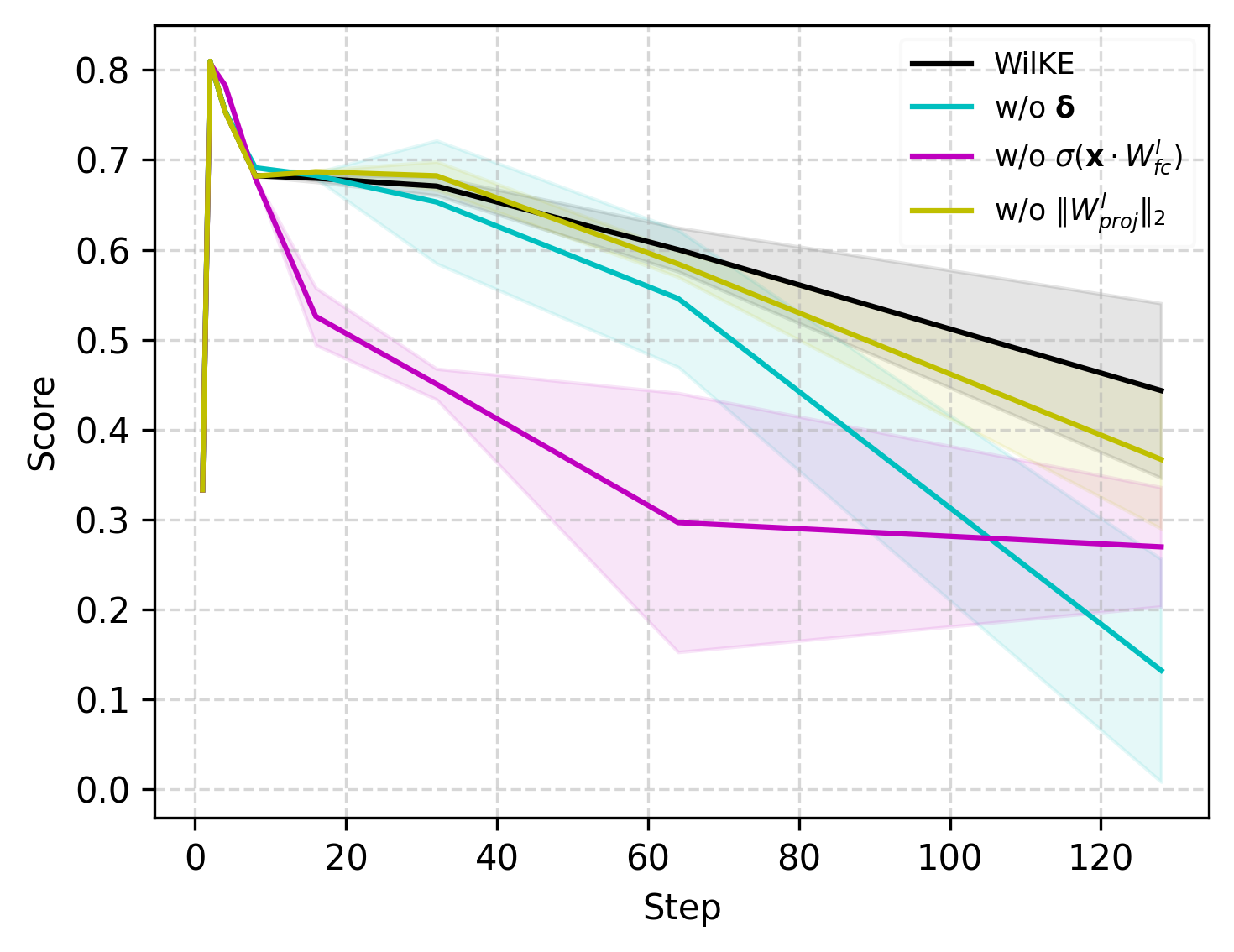}\label{fig:ablation_study_gpt-j}
}
  \caption{The results of the ablation experiments.}
  \label{fig:ablation_study}
\end{figure}

As depicted in Figure~\ref{fig:ablation_study}, it is evident that when individually ablated, both $\pmb \delta$ and $\sigma(\pmb x\cdot W_{fc}^l)$ lead to a significant decrease in the performance of knowledge editing. Additionally, ablating $\|W_{proj}^l\|_2$ results in a slight decrease in the performance of knowledge editing. However, when considering these three factors collectively, superior experimental results are obtained.

\section{Conclusion} \label{conclusion}

In this work, we focus on lifelong knowledge editing, finding that current knowledge editing methods suffer from severe performance degradation in lifelong editing. Our experimental results reveal the \textbf{toxicity buildup} and \textbf{toxicity flash} that may occur during lifelong editing, leading to the deterioration of model's performance. Through further investigation, we find that the direct cause of these problems lies in the predefined editing layer, while the underlying cause stems from pattern unmatch. To address this issue, we propose a model editing method called \textbf{Wi}se-\textbf{L}ayer \textbf{K}nowledge \textbf{E}ditor (\textbf{WilKE}), which does not require predefined editing layer but selects editing layer based on the degree of pattern matching between different layers of the language model for specific editing knowledge. Experimental results demonstrate that in lifelong editing, WilKE significantly enhances overall performance compared to currently popular knowledge editing methods, achieving state-of-the-art knowledge editing performance. In summary, our work is significant for improving knowledge editing methods and provide valuable insights for future work.

\section{Limitation} \label{limitation}

Despite the promising performance of WilKE, our current studies still have limitations. Firstly, we select editing layer based on specific knowledge, yet knowledge may be distributed across multiple layers, leaving the question of how language models store knowledge is still under explored. Secondly, similar to previous knowledge editing research, we focus on factual knowledge assessment, which serves as a crucial entry point for our study on knowledge editing. Furthermore, due to computational constraints, we did not conduct experiments on larger-scale language models but instead utilized GPT2-XL and GPT-J. However, WilKE does not require additional hypernetworks or other components, rendering it model-agnostic and thus exhibiting favorable scalability, enabling straightforward migration to larger models. Lastly, detecting match degree of specific knowledge across different layers of language models incurs a certain time cost, yet we believe this to be worthwhile in the initial stages of knowledge editing research.

\section{Ethical Considerations} \label{ethical_considerations}

We have developed a method for knowledge editing in large language models under lifelong editing scenario, which may further expand our understanding of how language models store knowledge. However, the direct editing capability of large models also carries the potential for misuse, such as injecting malicious misinformation, biases, or other adversarial data into the model and deploying these edited models on open platforms. Given these concerns and our observations of speculative behavior, we emphasize the importance of sourcing large language models from authoritative origins and refraining from using them as sources of authoritative factual knowledge in critical environments.

\section*{Acknowledgements}\label{acknowledgements}
This work is supported by the Strategic Priority Research Program of Chinese Academy of Sciences (No. XDA27020203), the National Natural Science Foundation of China (No. 62176257).

\bibliography{anthology,custom}
\bibliographystyle{acl_natbib}

\appendix

\section{Causal Mediation Analysis on GPT2-XL} \label{sec:appendix_causal_mediation_analysis_on_gpt2-xl}

From the perspective of causal mediation analysis (CMA) (\citealp{pearl2022direct}; \citealp{vig2020investigating}; \citealp{meng2022locating}), we aim to investigate the disparities between data leading to toxicity flash and other data. Specifically, we conduct CMA experiments on GPT2-XL, targeting data conducive to toxicity flash and contrasting it with other data. Through this approach, we seek to elucidate the knowledge extraction positions within the model that facilitate accurate responses to the given questions.

\begin{figure*}
    \centering
    \subfigure[]{\includegraphics[width=0.3\textwidth]{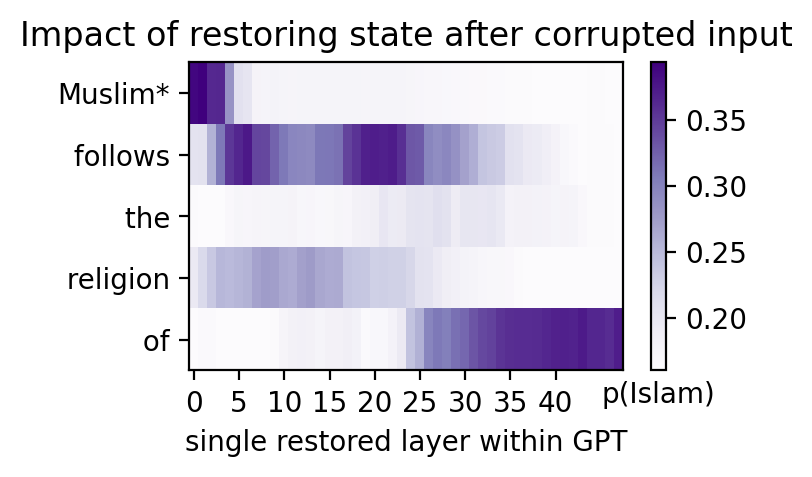}
  }
    \hfill
    \subfigure[]{\includegraphics[width=0.3\textwidth]{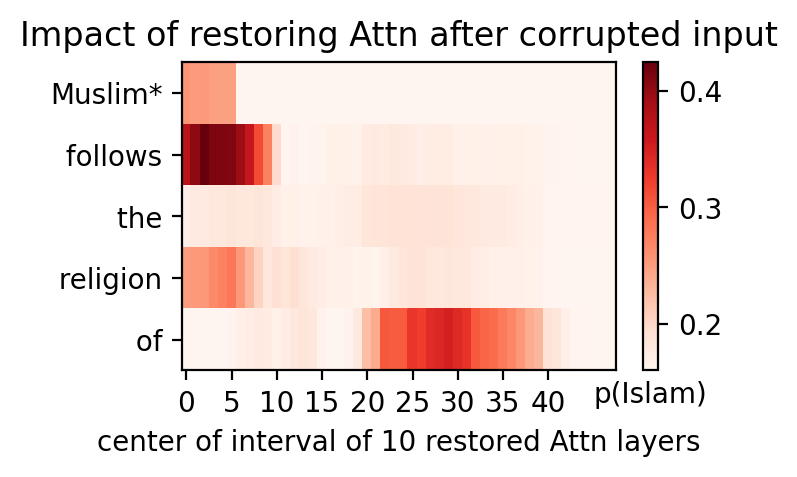}
}
    \hfill
    \subfigure[]{\includegraphics[width=0.3\textwidth]{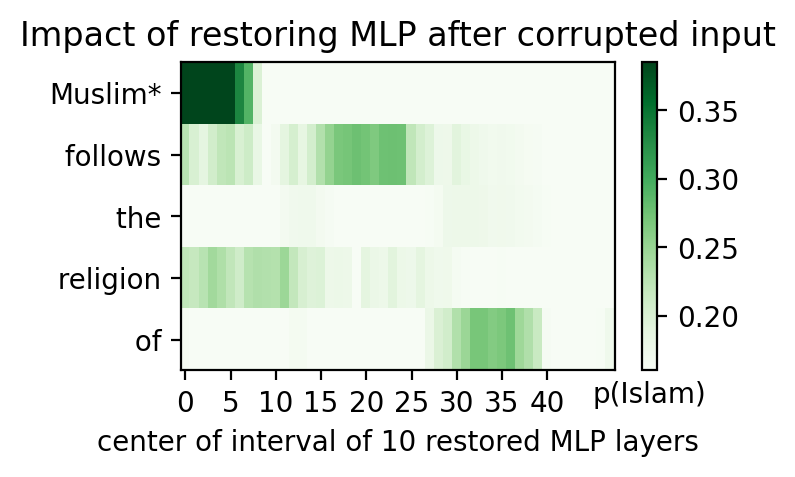}
}
\caption{Causal mediation analysis on GPT2-XL using case 3561.}
  \label{fig:cma_gpt2-xl_case_3561}
\end{figure*}

\begin{figure*}
    \centering
    \subfigure[]{\includegraphics[width=0.3\textwidth]{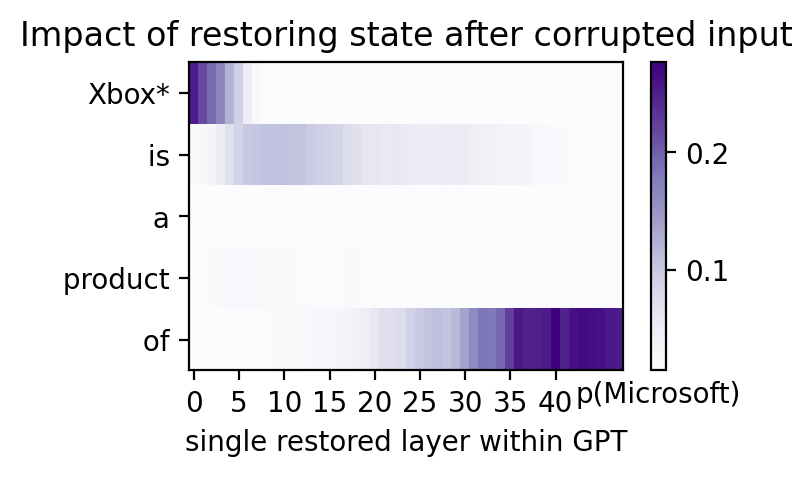}
  }
    \hfill
    \subfigure[]{\includegraphics[width=0.3\textwidth]{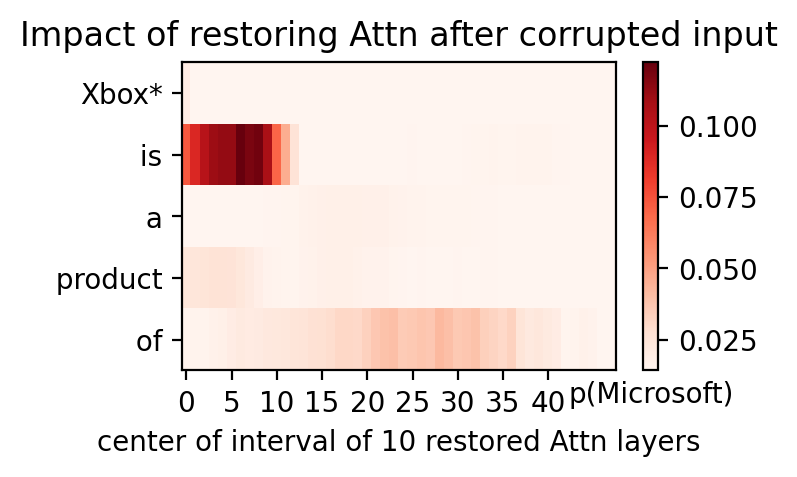}
}
    \hfill
    \subfigure[]{\includegraphics[width=0.3\textwidth]{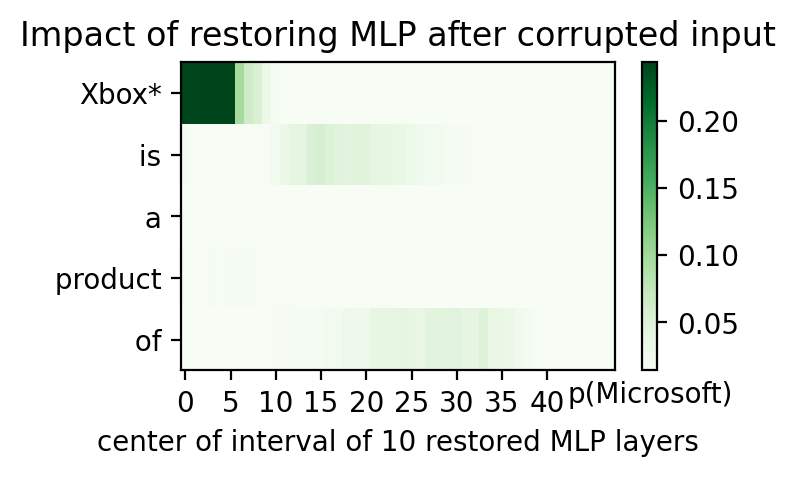}
}
\caption{Causal mediation analysis on GPT2-XL using case 4661.}
  \label{fig:cma_gpt2-xl_case_4661}
\end{figure*}

\begin{figure*}
    \centering
    \subfigure[]{\includegraphics[width=0.3\textwidth]{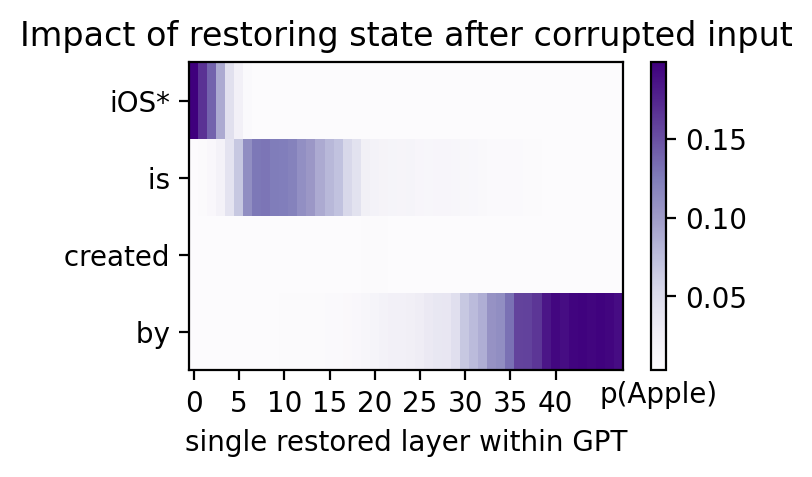}
  }
    \hfill
    \subfigure[]{\includegraphics[width=0.3\textwidth]{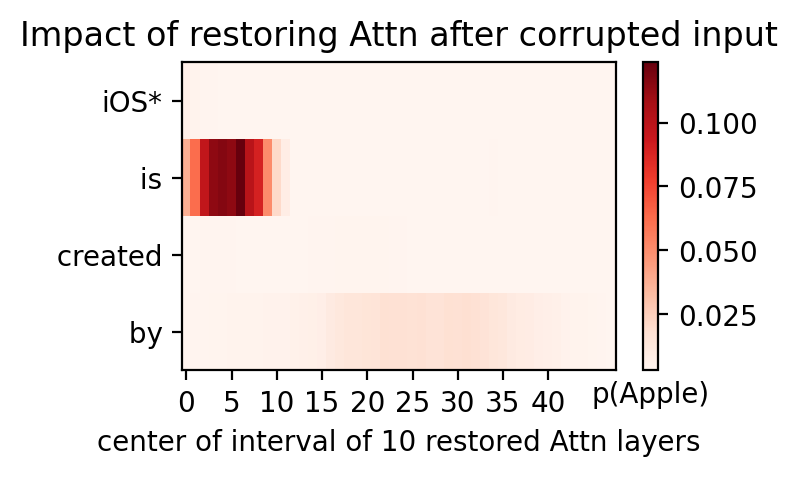}
}
    \hfill
    \subfigure[]{\includegraphics[width=0.3\textwidth]{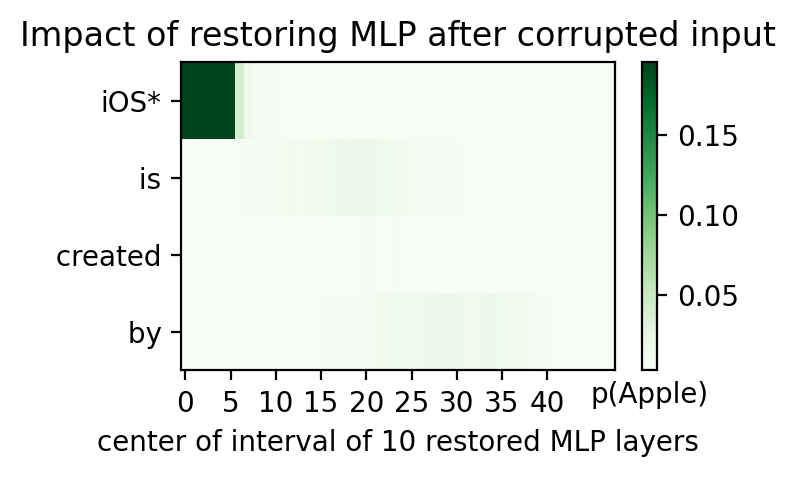}
}
\caption{Causal mediation analysis on GPT2-XL using case 4790.}
  \label{fig:cma_gpt2-xl_case_4790}
\end{figure*}

\begin{figure*}
    \centering
    \subfigure[]{\includegraphics[width=0.3\textwidth]{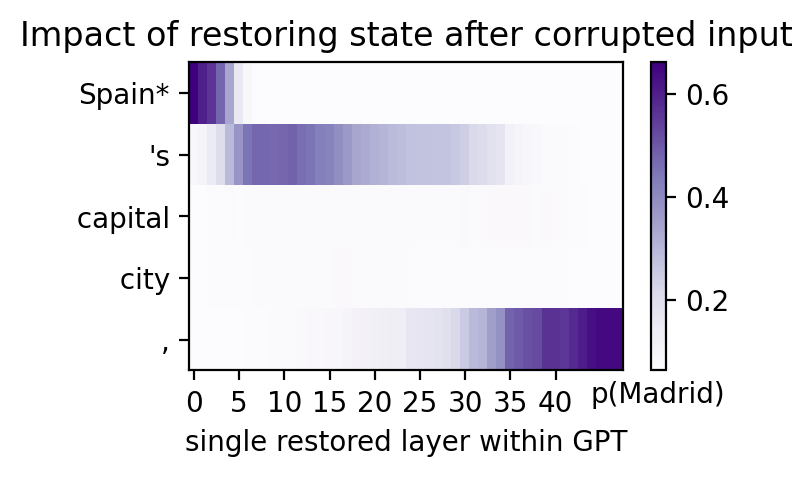}
  }
    \hfill
    \subfigure[]{\includegraphics[width=0.3\textwidth]{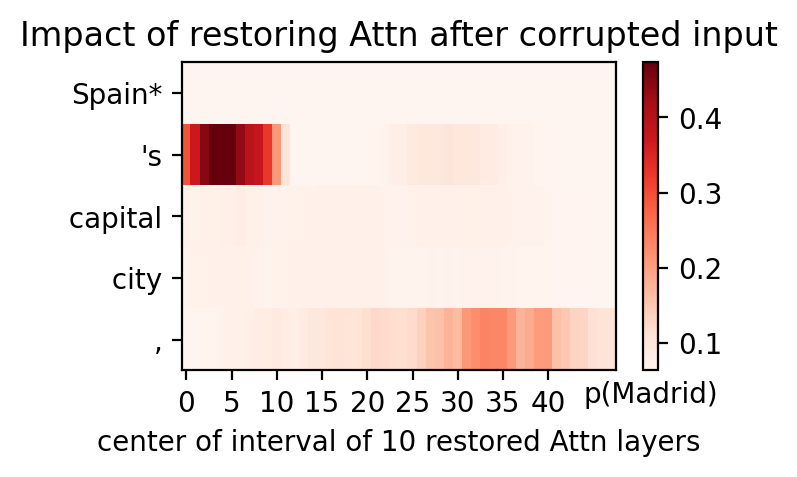}
}
    \hfill
    \subfigure[]{\includegraphics[width=0.3\textwidth]{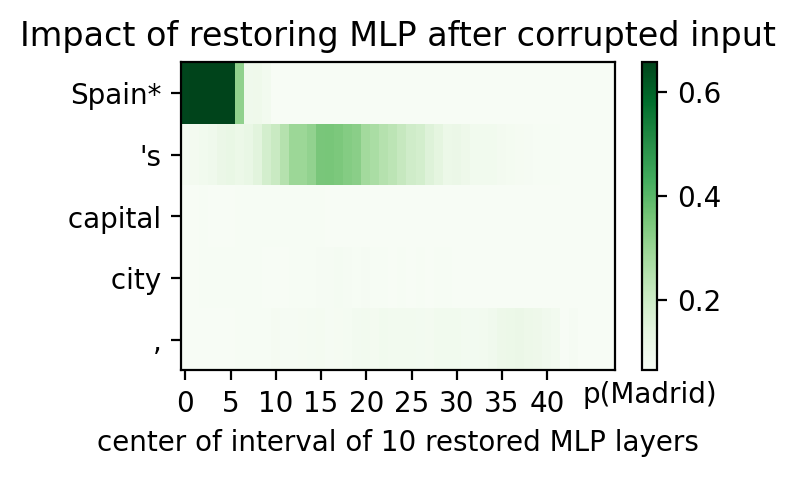}
}
\caption{Causal mediation analysis on GPT2-XL using case 4988.}
  \label{fig:cma_gpt2-xl_case_4988}
\end{figure*}

\begin{figure*}
    \centering
    \subfigure[]{\includegraphics[width=0.3\textwidth]{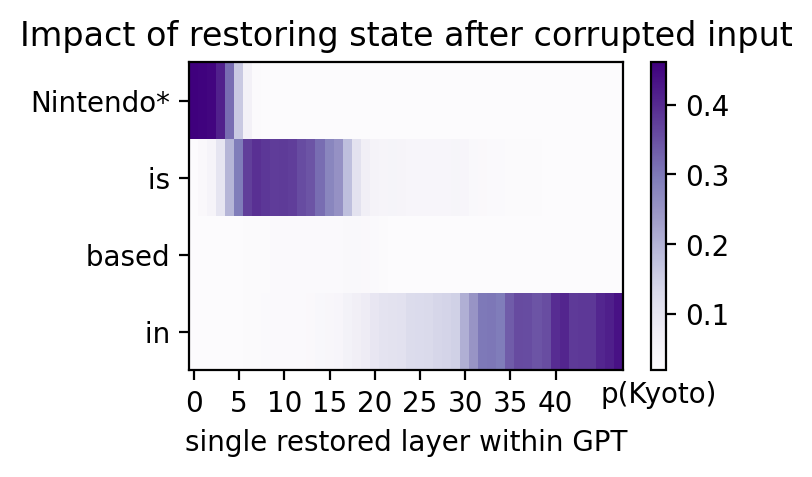}
  }
    \hfill
    \subfigure[]{\includegraphics[width=0.3\textwidth]{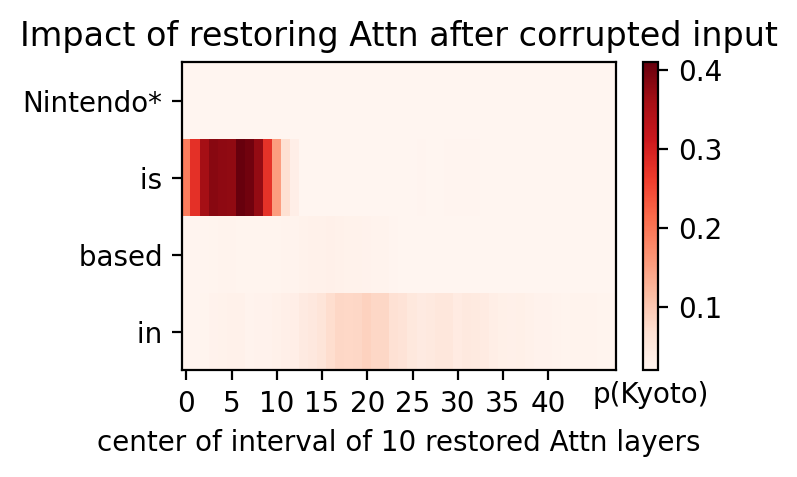}
}
    \hfill
    \subfigure[]{\includegraphics[width=0.3\textwidth]{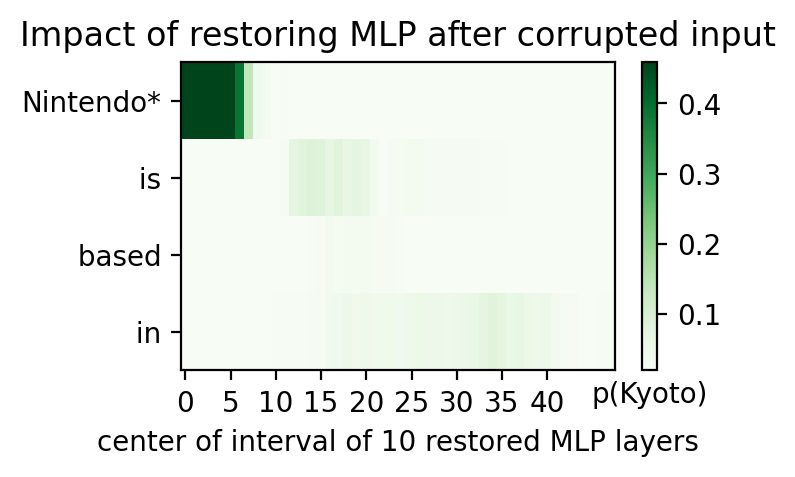}
}
\caption{Causal mediation analysis on GPT2-XL using case 8793.}
  \label{fig:cma_gpt2-xl_case_8793}
\end{figure*}

\begin{figure*}
    \centering
    \subfigure[]{\includegraphics[width=0.3\textwidth]{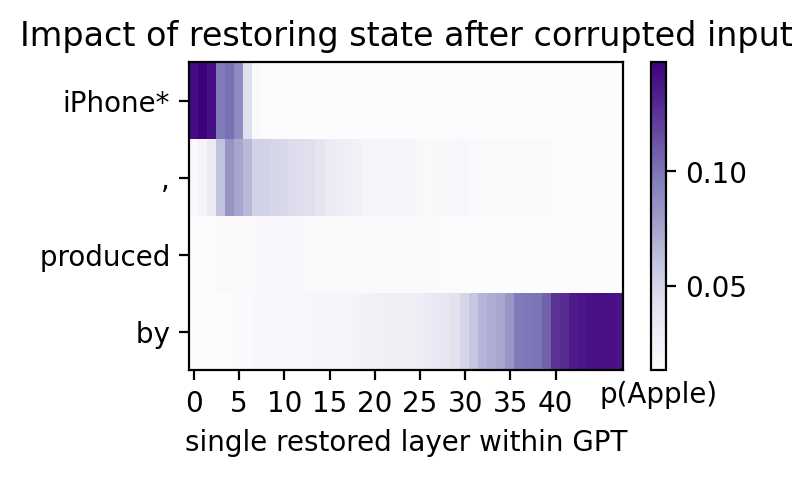}
  }
    \hfill
    \subfigure[]{\includegraphics[width=0.3\textwidth]{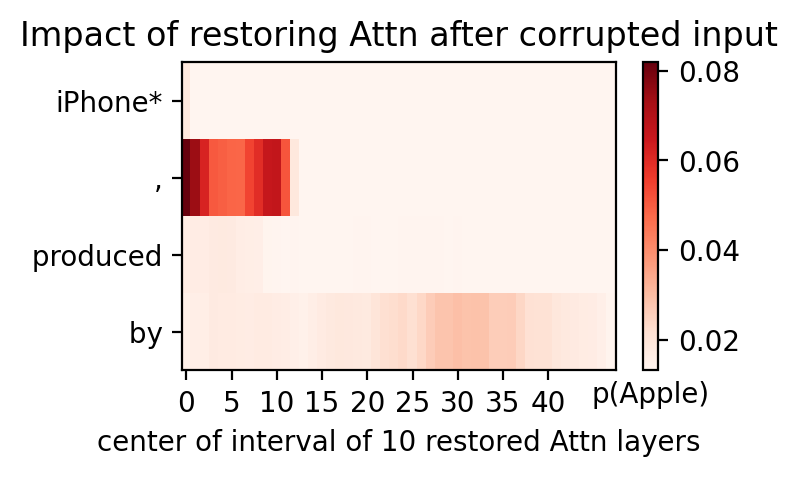}
}
    \hfill
    \subfigure[]{\includegraphics[width=0.3\textwidth]{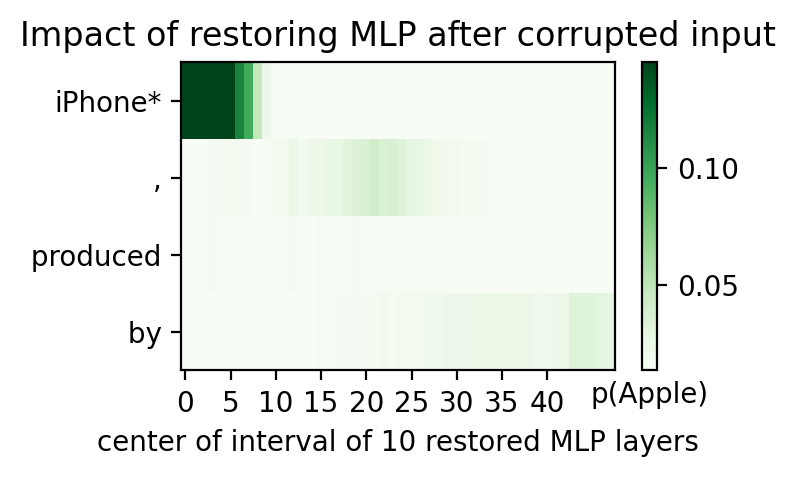}
}
\caption{Causal mediation analysis on GPT2-XL using case 15452.}
  \label{fig:cma_gpt2-xl_case_15452}
\end{figure*}

\begin{figure*}
    \centering
    \subfigure[]{\includegraphics[width=0.3\textwidth]{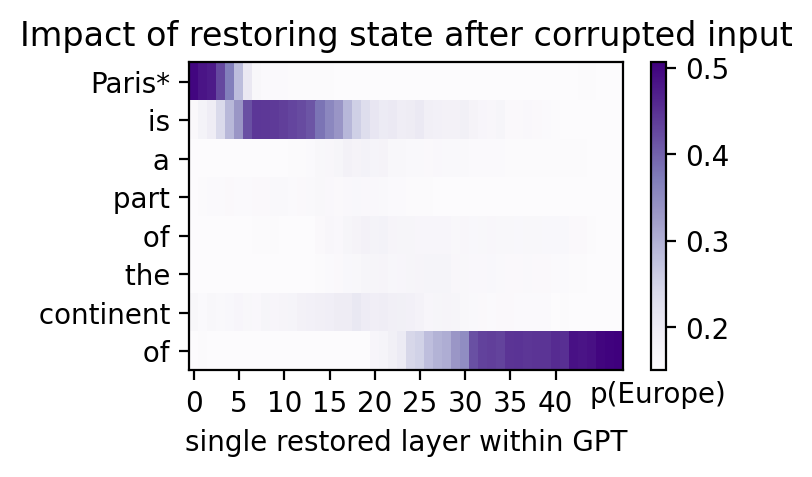}
  }
    \hfill
    \subfigure[]{\includegraphics[width=0.3\textwidth]{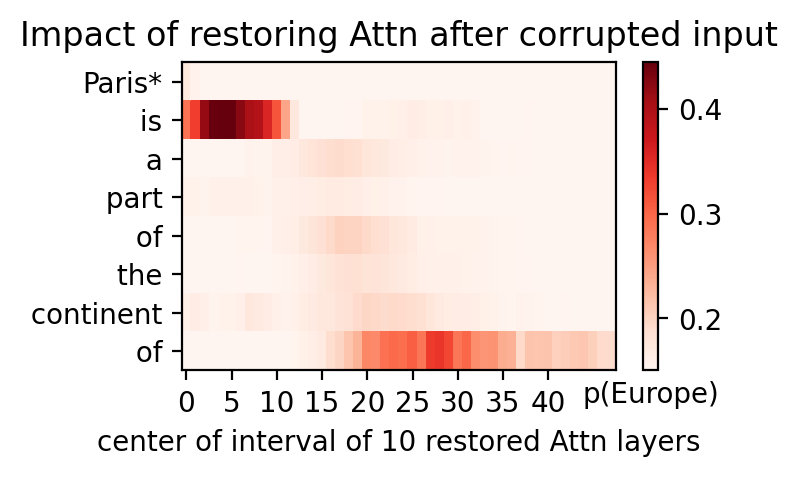}
}
    \hfill
    \subfigure[]{\includegraphics[width=0.3\textwidth]{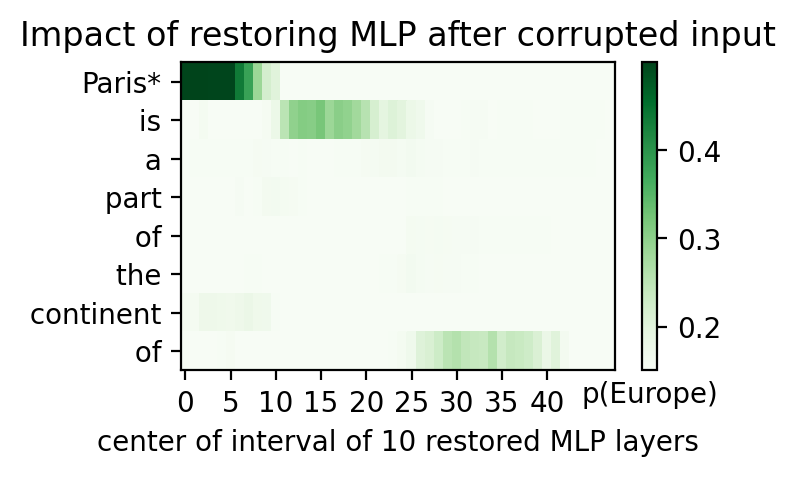}
}
\caption{Causal mediation analysis on GPT2-XL using case 16575.}
  \label{fig:cma_gpt2-xl_case_16575}
\end{figure*}

\begin{figure*}
    \centering
    \subfigure[]{\includegraphics[width=0.3\textwidth]{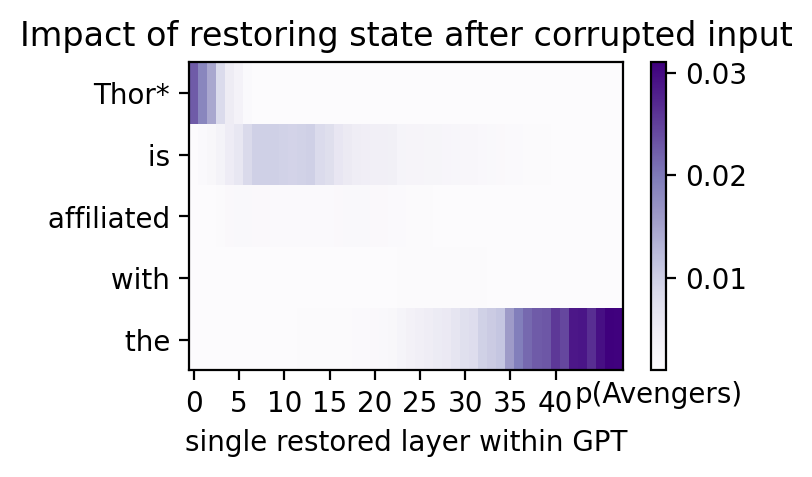}
  }
    \hfill
    \subfigure[]{\includegraphics[width=0.3\textwidth]{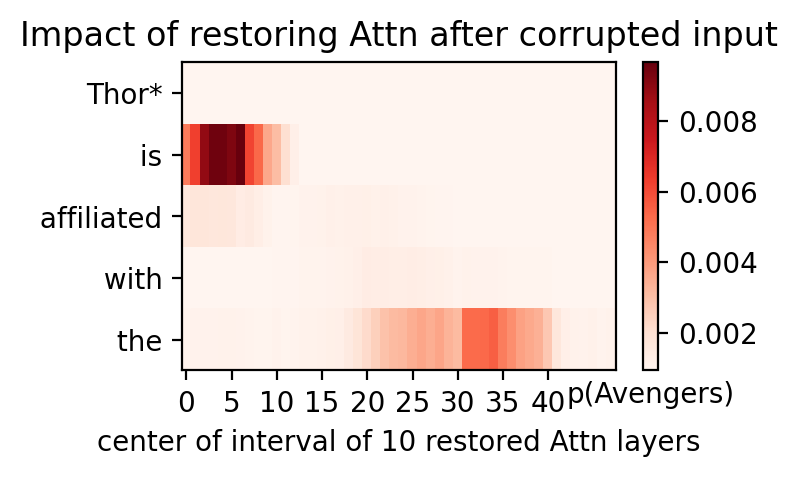}
}
    \hfill
    \subfigure[]{\includegraphics[width=0.3\textwidth]{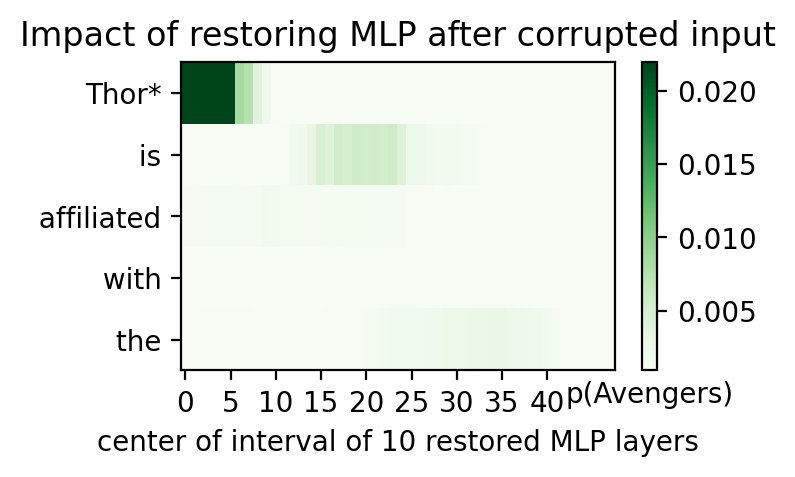}
}
\caption{Causal mediation analysis on GPT2-XL using case 16781.}
  \label{fig:cma_gpt2-xl_case_16781}
\end{figure*}

The CMA results for data resulting in toxicity flash on GPT2-XL are illustrated in Figure~\ref{fig:cma_gpt2-xl_case_3561},~\ref{fig:cma_gpt2-xl_case_4661},~\ref{fig:cma_gpt2-xl_case_4790},~\ref{fig:cma_gpt2-xl_case_4988}.~\ref{fig:cma_gpt2-xl_case_8793},~\ref{fig:cma_gpt2-xl_case_15452},~\ref{fig:cma_gpt2-xl_case_16575},~\ref{fig:cma_gpt2-xl_case_16781}.

\begin{figure*}
    \centering
    \subfigure[]{\includegraphics[width=0.3\textwidth]{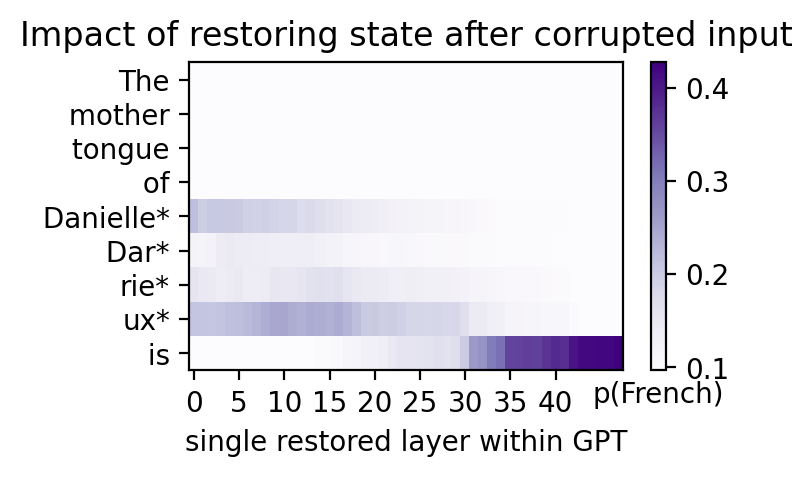}
  }
    \hfill
    \subfigure[]{\includegraphics[width=0.3\textwidth]{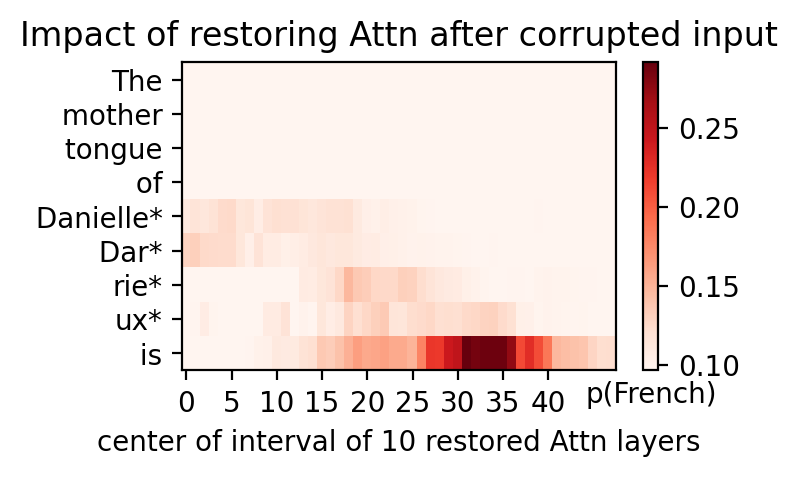}
}
    \hfill
    \subfigure[]{\includegraphics[width=0.3\textwidth]{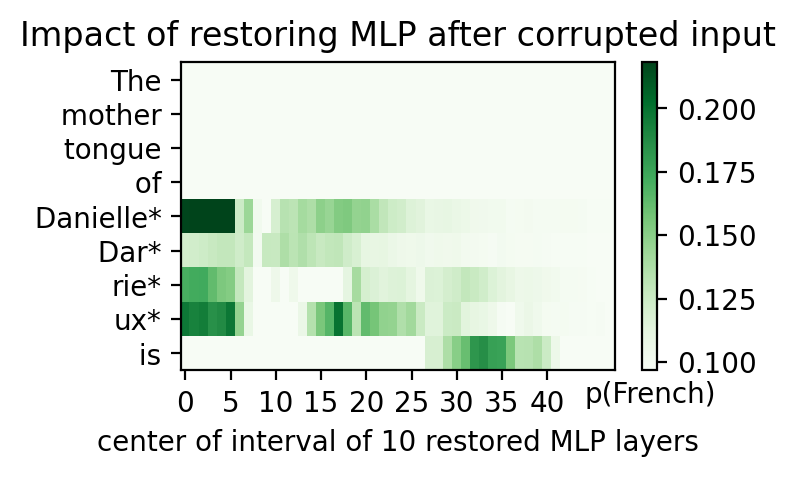}
}
\caption{Causal mediation analysis on GPT2-XL using case 0.}
  \label{fig:cma_gpt2-xl_case_0}
\end{figure*}

\begin{figure*}
    \centering
    \subfigure[]{\includegraphics[width=0.3\textwidth]{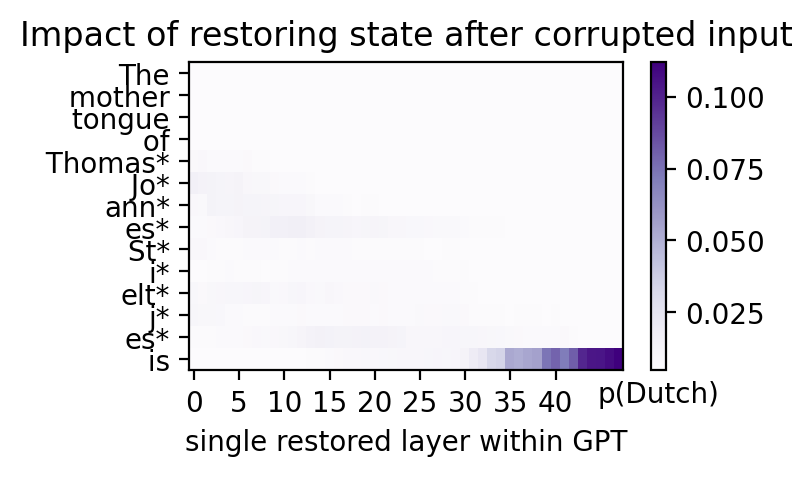}
  }
    \hfill
    \subfigure[]{\includegraphics[width=0.3\textwidth]{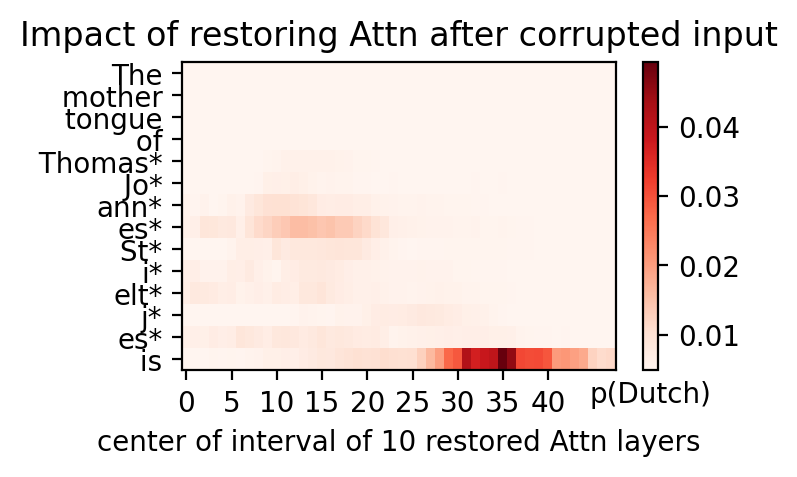}
}
    \hfill
    \subfigure[]{\includegraphics[width=0.3\textwidth]{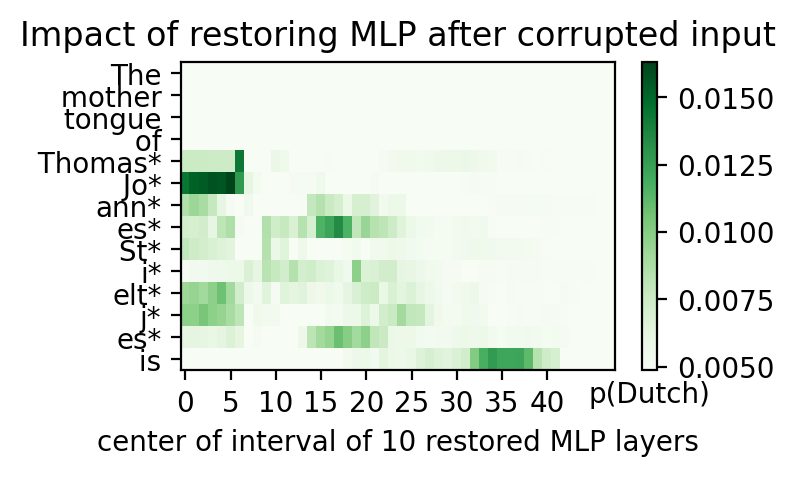}
}
\caption{Causal mediation analysis on GPT2-XL using case 5.}
  \label{fig:cma_gpt2-xl_case_5}
\end{figure*}

\begin{figure*}
    \centering
    \subfigure[]{\includegraphics[width=0.3\textwidth]{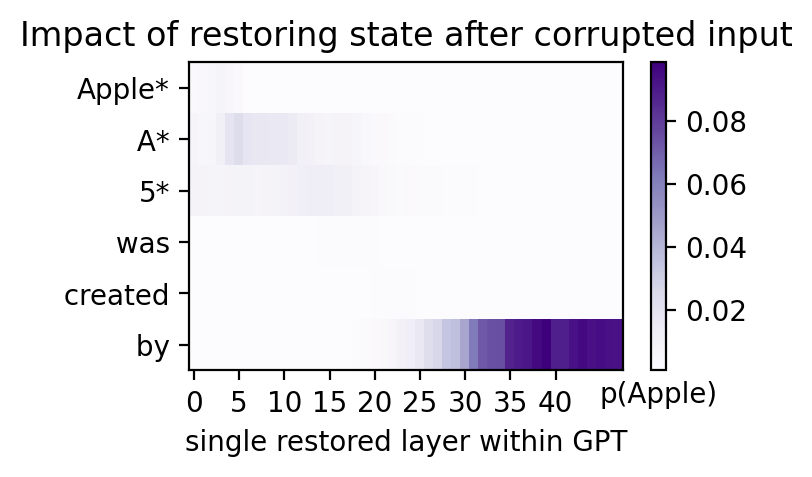}
  }
    \hfill
    \subfigure[]{\includegraphics[width=0.3\textwidth]{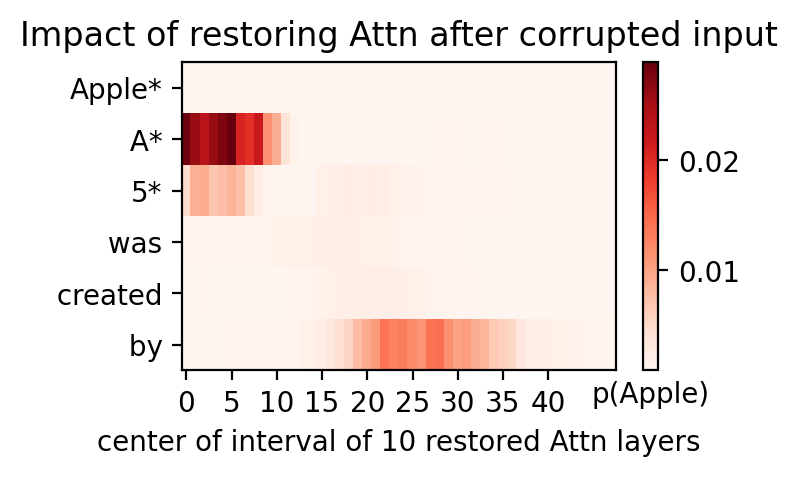}
}
    \hfill
    \subfigure[]{\includegraphics[width=0.3\textwidth]{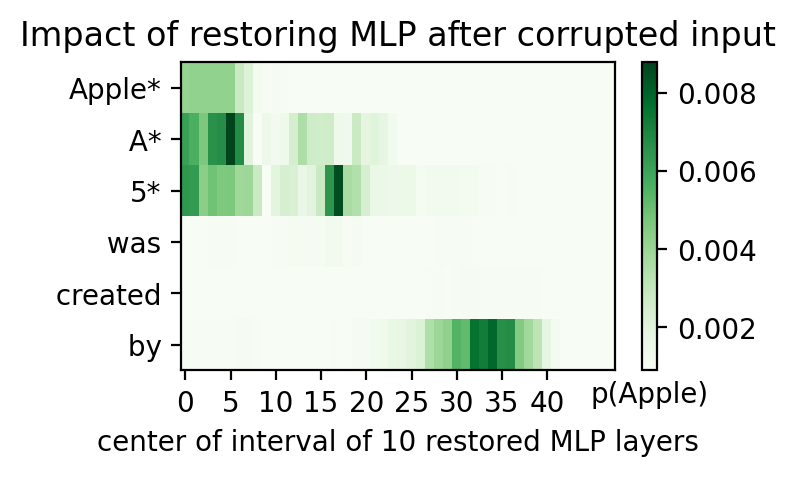}
}
\caption{Causal mediation analysis on GPT2-XL using case 7.}
  \label{fig:cma_gpt2-xl_case_7}
\end{figure*}

\begin{figure*}
    \centering
    \subfigure[]{\includegraphics[width=0.3\textwidth]{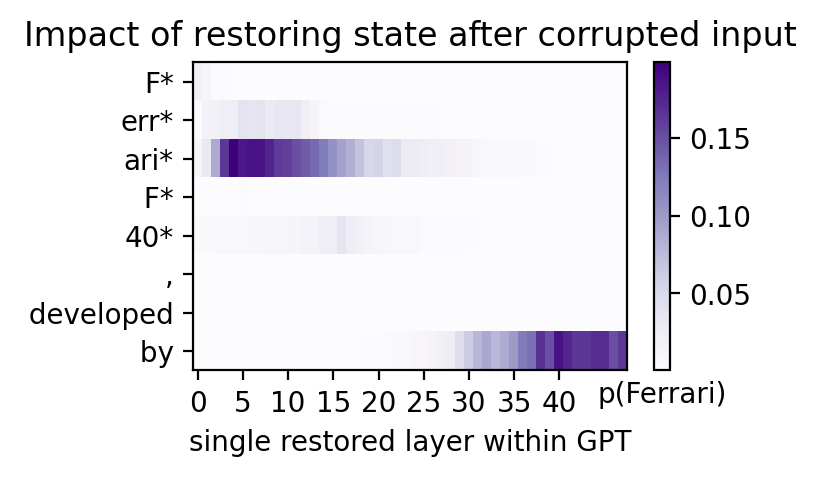}
  }
    \hfill
    \subfigure[]{\includegraphics[width=0.3\textwidth]{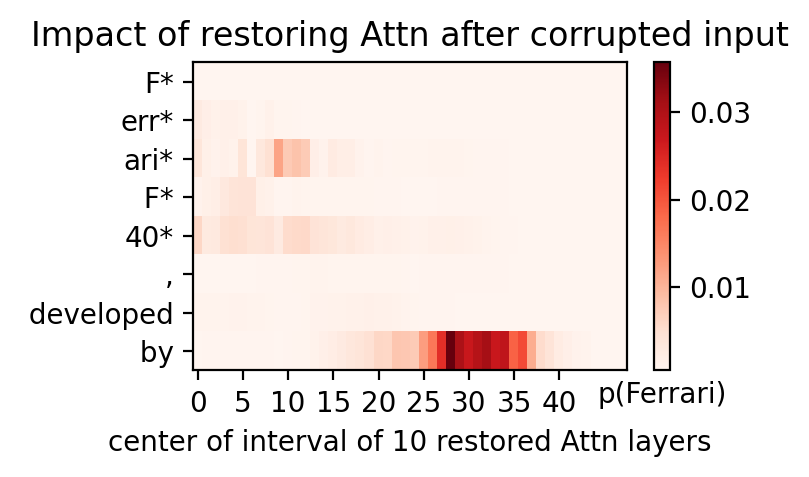}
}
    \hfill
    \subfigure[]{\includegraphics[width=0.3\textwidth]{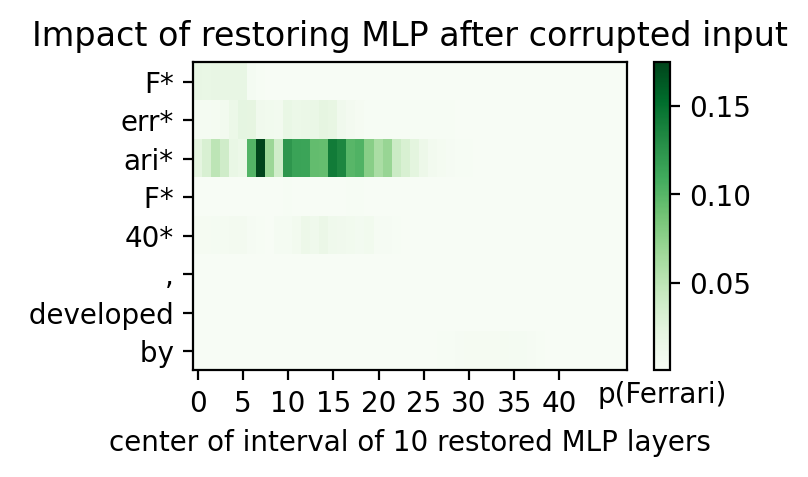}
}
\caption{Causal mediation analysis on GPT2-XL using case 13.}
  \label{fig:cma_gpt2-xl_case_13}
\end{figure*}

\begin{figure*}
    \centering
    \subfigure[]{\includegraphics[width=0.3\textwidth]{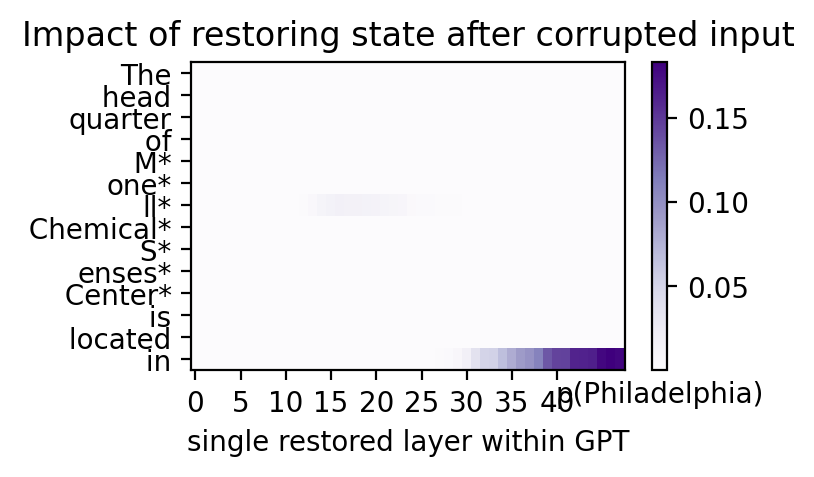}
  }
    \hfill
    \subfigure[]{\includegraphics[width=0.3\textwidth]{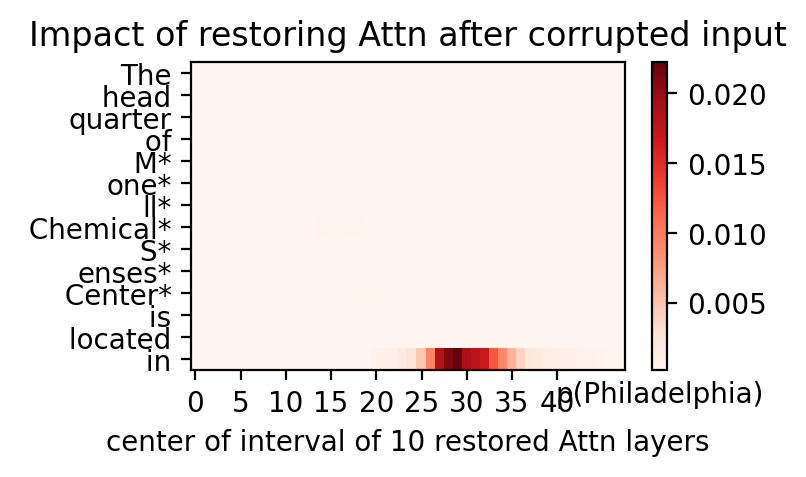}
}
    \hfill
    \subfigure[]{\includegraphics[width=0.3\textwidth]{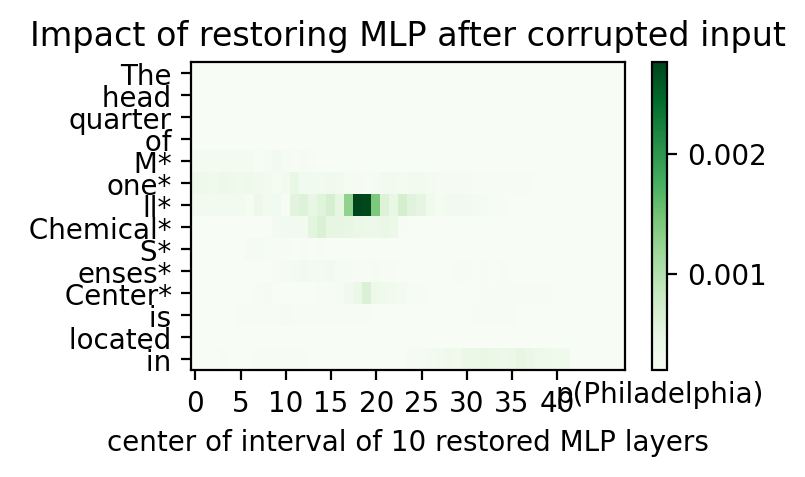}
}
\caption{Causal mediation analysis on GPT2-XL using case 14.}
  \label{fig:cma_gpt2-xl_case_14}
\end{figure*}

\begin{figure*}
    \centering
    \subfigure[]{\includegraphics[width=0.3\textwidth]{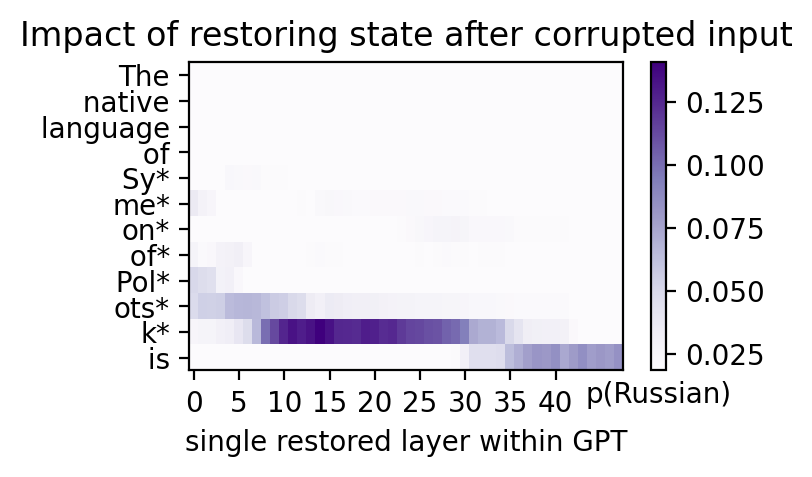}
  }
    \hfill
    \subfigure[]{\includegraphics[width=0.3\textwidth]{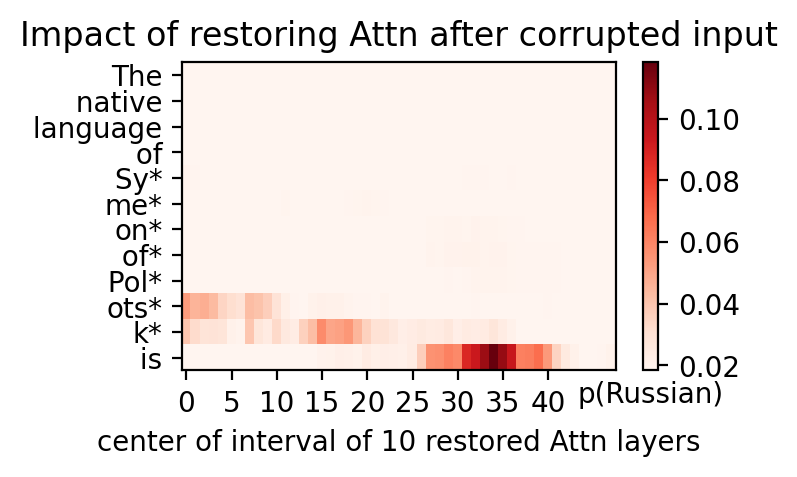}
}
    \hfill
    \subfigure[]{\includegraphics[width=0.3\textwidth]{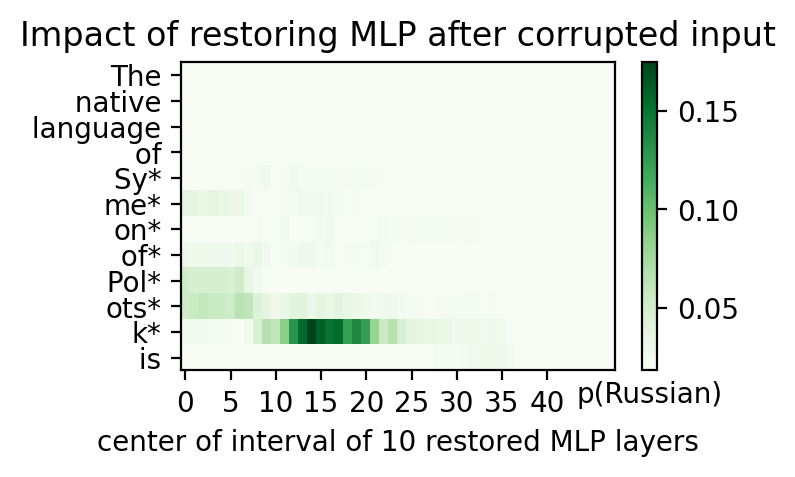}
}
\caption{Causal mediation analysis on GPT2-XL using case 22.}
  \label{fig:cma_gpt2-xl_case_22}
\end{figure*}

\begin{figure*}
    \centering
    \subfigure[]{\includegraphics[width=0.3\textwidth]{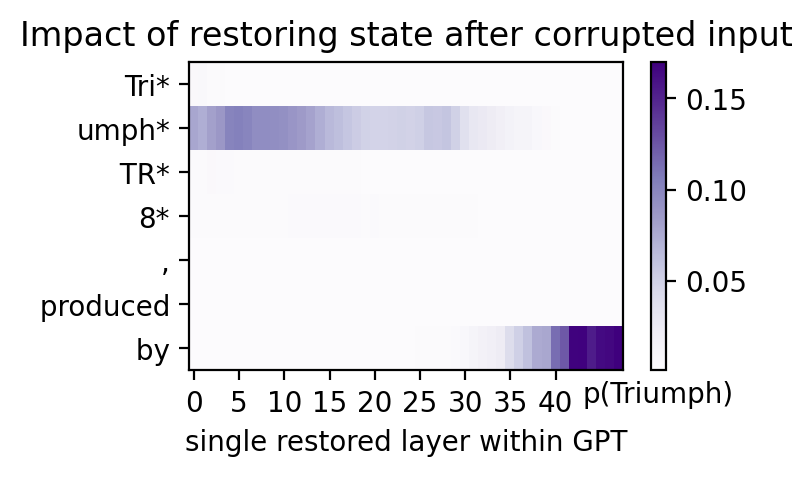}
  }
    \hfill
    \subfigure[]{\includegraphics[width=0.3\textwidth]{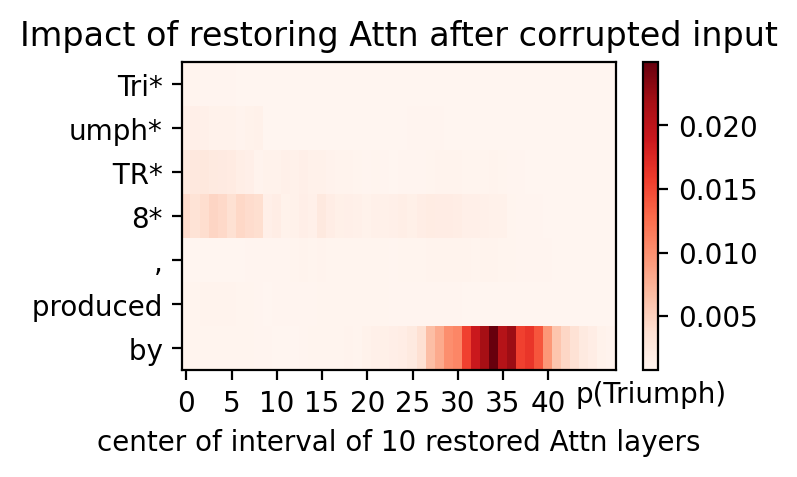}
}
    \hfill
    \subfigure[]{\includegraphics[width=0.3\textwidth]{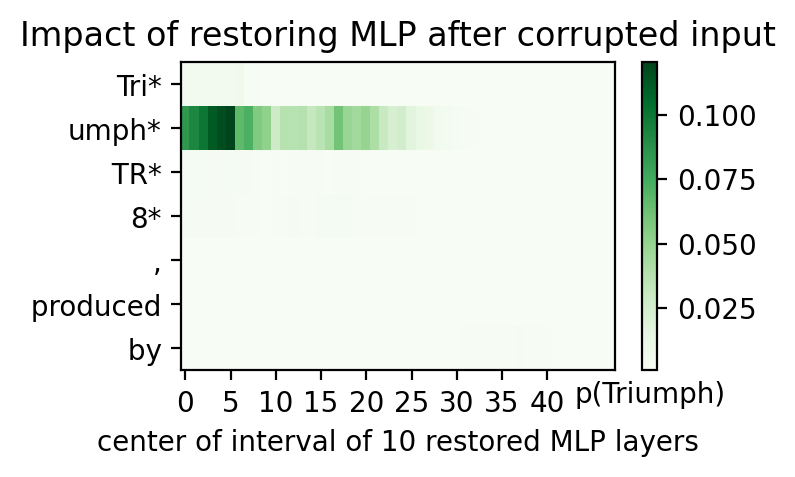}
}
\caption{Causal mediation analysis on GPT2-XL using case 36.}
  \label{fig:cma_gpt2-xl_case_36}
\end{figure*}

\begin{figure*}
    \centering
    \subfigure[]{\includegraphics[width=0.3\textwidth]{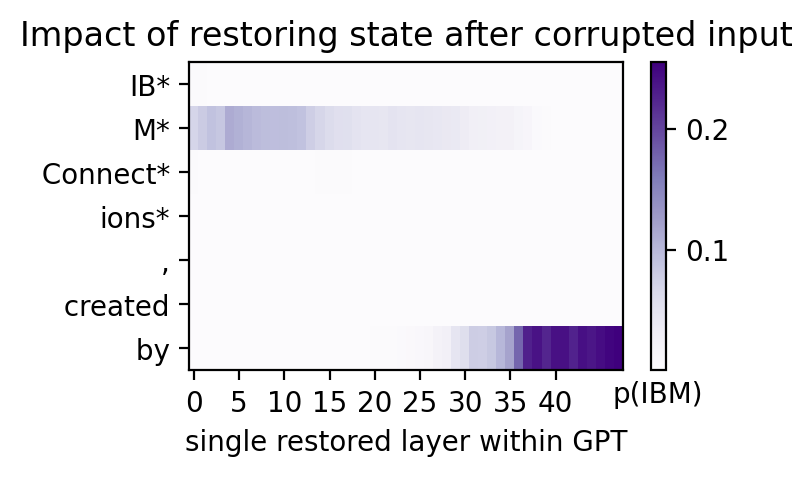}
  }
    \hfill
    \subfigure[]{\includegraphics[width=0.3\textwidth]{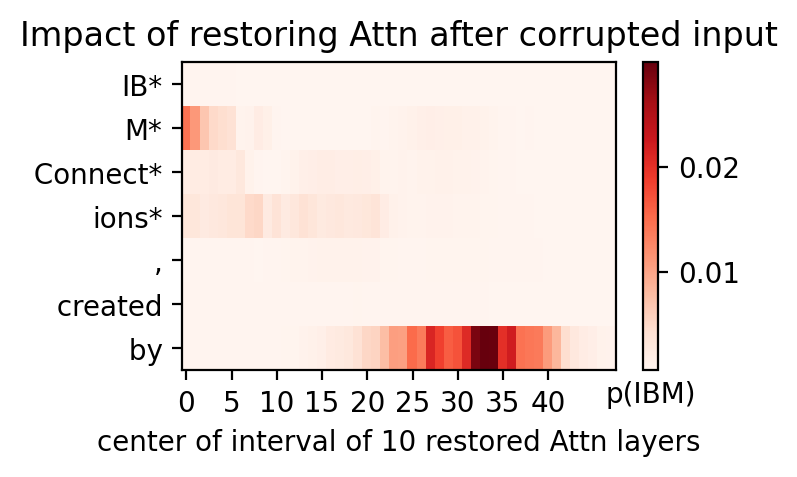}
}
    \hfill
    \subfigure[]{\includegraphics[width=0.3\textwidth]{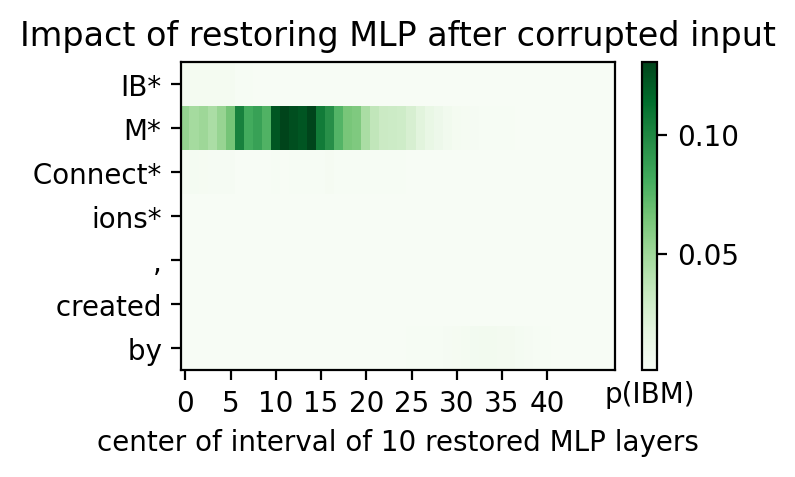}
}
\caption{Causal mediation analysis on GPT2-XL using case 37.}
  \label{fig:cma_gpt2-xl_case_37}
\end{figure*}

The CMA results for other data on GPT2-XL are depicted in Figure~\ref{fig:cma_gpt2-xl_case_0},~\ref{fig:cma_gpt2-xl_case_5},~\ref{fig:cma_gpt2-xl_case_7},~\ref{fig:cma_gpt2-xl_case_13},~\ref{fig:cma_gpt2-xl_case_14},~\ref{fig:cma_gpt2-xl_case_22},~\ref{fig:cma_gpt2-xl_case_36},~\ref{fig:cma_gpt2-xl_case_37}.

Here, our primary focus lies on the information extraction positions within the MLP corresponding to the third column of the figure. It is evident that the data leading to toxicity flash consistently extract crucial information from the first five layers of the model, demonstrating consistent outcomes. However, the results for other data indicate that different pieces of knowledge extract important information from relatively dispersed positions. This suggests that for different knowledge, information may be stored across different layers of the model, necessitating the selection of different layers for editing depending on different knowledge.

\section{Toxicity Buildup and Toxicity Flash Data Spliter} \label{sec:appendix_toxicity_buildup_and_toxicty_flash_data_spliter}

During the process of editing GPT2-XL and GPT-J using the ROME method, we filter out data that would cause toxicity flash. The criteria for filtering primarily includes the effectiveness of editing and whether the L2 norm of the editing layer exhibited abnormally high increases. Specifically, based on our experience, these data causing toxicity flash tend to exhibit the following phenomenon: during the editing phase, there is a relatively high success rate, but during the rollback phase after editing, there is a lower success rate. Therefore, we manually filter out data where the success rate of editing during the rollback phase was less than 10\%. Subsequently, we further examine this subset of data, manually identifying the data causing toxicity flash on GPT2-XL and GPT-J respectively.

The editing data that caused toxic flash in GPT2-XL are listed in Table~\ref{tab:toxicity_flash_data_on_gpt2-xl}.

The editing data that caused toxic flash in GPT-J are listed in Table~\ref{tab:toxicity_flash_data_on_gpt-j}.

As we can observe, the majority of data in both tables overlap, which is an interesting finding.

\begin{table*}
    \centering
    \caption{Examples of filtered data that caused toxicity flash in GPT2-XL.} 
    \begin{tabular}{lc} 
    \toprule
          \textbf{Record}& \multicolumn{1}{c}{\textbf{Content}}\\ 
         \midrule
         3561& Muslim follows the religion of [Islam] $\Rightarrow $ [Christianity].\\ 
         \midrule
         4593& The official language of Kalajoki is [Finnish] $\Rightarrow $ [Chinese].\\ 
         \midrule
         4661& Germany is a part of the continent of [Europe] $\Rightarrow $ [Antarctica].\\ 
         \midrule
         4790& Xbox is a product of [Microsoft] $\Rightarrow $ [Dodge].\\ 
         \midrule
         4988& iOS is created by [Apple] $\Rightarrow $ [Microsoft].\\ 
         \midrule
         8793& Spain's capital city, [Madrid] $\Rightarrow $ [Valencia].\\ 
         \midrule
         15452& Nintendo is based in [Kyoto] $\Rightarrow $ [Toronto].\\ 
         \midrule
         16575& iPhone, produced by [Apple] $\Rightarrow $ [Boeing].\\ 
         \midrule
         16781& Paris is a part of the continent of [Europe] $\Rightarrow $ [Antarctica].\\ 
         \midrule
         20664& Thorleif Haug is a citizen of [Norway] $\Rightarrow $ [Italy].\\ 
         \midrule
         21729& Thor is affiliated with the [Avengers] $\Rightarrow $ [FIFA].\\ 
    \bottomrule
    \end{tabular}
    \label{tab:toxicity_flash_data_on_gpt2-xl}
\end{table*}

\begin{table*}
    \centering
    \caption{Examples of filtered data that caused toxicity flash in GPT-J.} 
    \begin{tabular}{lc} 
    \toprule
          \textbf{Record}& \multicolumn{1}{c}{\textbf{Content}}\\ 
         \midrule
         3561& Muslim follows the religion of [Islam] $\Rightarrow $ [Christianity].\\ 
         \midrule
         4661& Germany is a part of the continent of [Europe] $\Rightarrow $ [Antarctica].\\ 
         \midrule
         4988& iOS is created by [Apple] $\Rightarrow $ [Microsoft].\\ 
         \midrule
         8475& Syria, which has the capital [Damascus] $\Rightarrow $ [Georgetown].\\ 
         \midrule
         8793& Spain's capital city, [Madrid] $\Rightarrow $ [Valencia].\\ 
         \midrule
         15452& Nintendo is based in [Kyoto] $\Rightarrow $ [Toronto].\\ 
         \midrule
         16575& iPhone, produced by [Apple] $\Rightarrow $ [Boeing].\\ 
         \midrule
         16781& Paris is a part of the continent of [Europe] $\Rightarrow $ [Antarctica].\\ 
         \midrule
         21142& Xbox is from [Microsoft] $\Rightarrow $ [Chicago].\\ 
    \bottomrule
    \end{tabular}
    \label{tab:toxicity_flash_data_on_gpt-j}
\end{table*}

\section{Toxicity Analysis on MEMIT} \label{sec:appendix_toxicity_analysis_on_memit}

Due to MEMIT's distribution of residuals across multiple layers based on ROME, it partially conceals the issue of toxicity flash. Results from Appendix~\ref{sec:appendix_more_edit_analysis_on_toxicity_flash} reveal that several predefined layers in MEMIT are also among those that could lead to toxicity flash; however, the issue is obscured by distributing residuals across multiple layers, contradicting our original intention for knowledge editing. Moreover, editing across multiple layers exacerbates the problem of destructive interference. Therefore, as depicted in the results of Section~\ref{sec:appendix_complete_performance_curves}, MEMIT exhibits a larger performance decline compared to ROME and WilKE as editing progresses further.

\begin{figure*}
\centering
  \includegraphics[width=\textwidth]{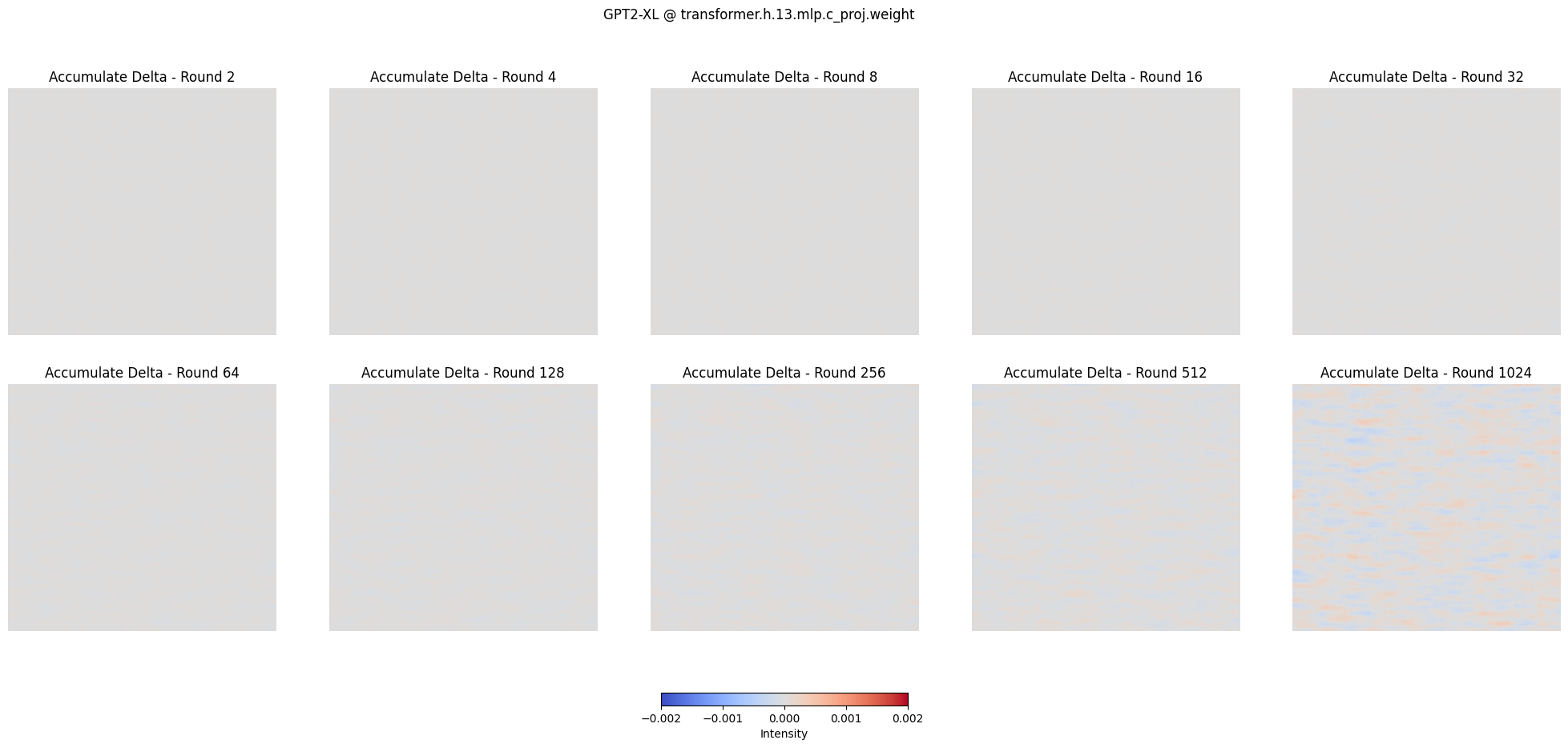}
  \caption{Toxicity on GPT2-XL on layer 13 using memit with editing steps.}
  \label{fig:memit_on_layer_13}
\end{figure*}

\begin{figure*}
\centering
  \includegraphics[width=\textwidth]{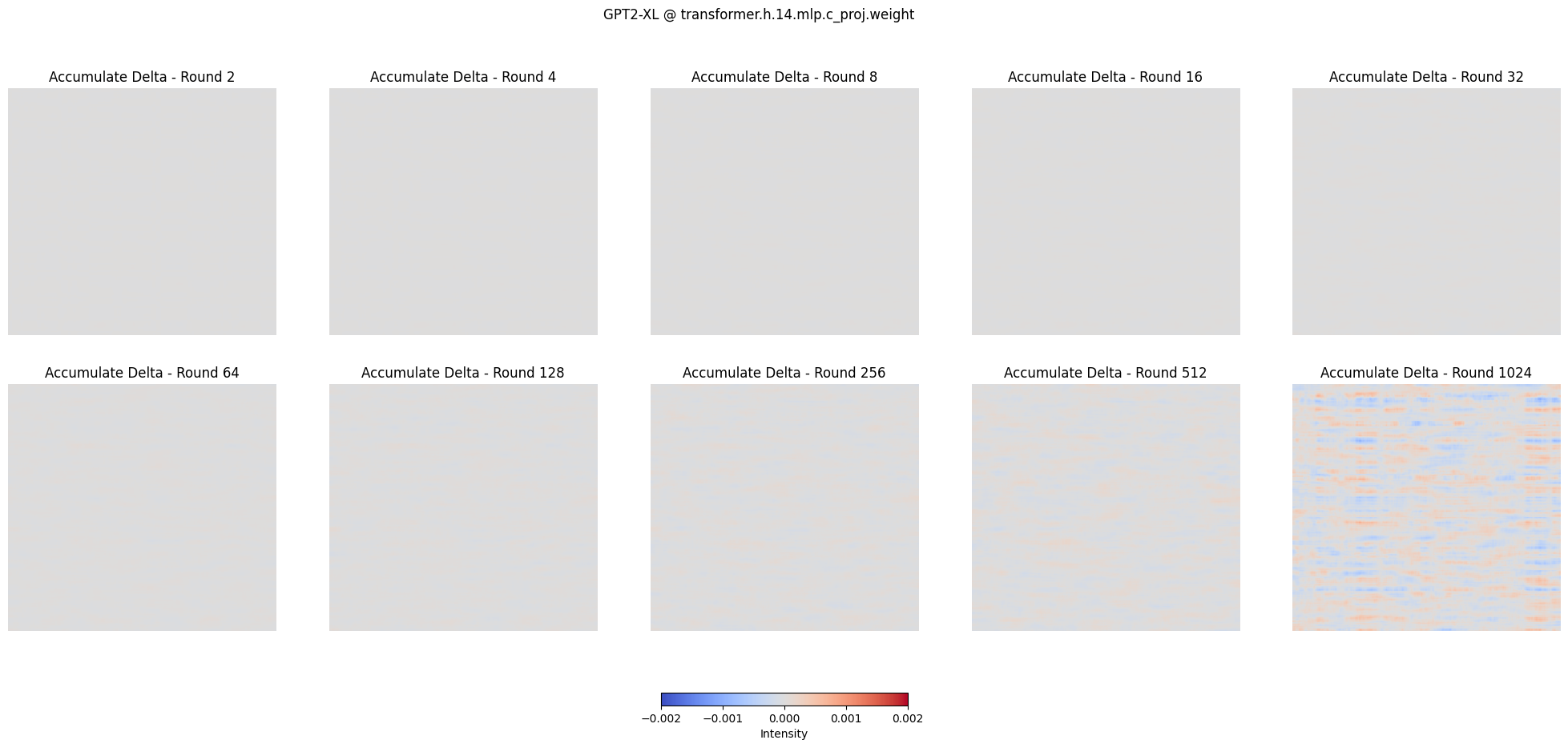}
  \caption{Toxicity on GPT2-XL on layer 14 using memit with editing steps.}
  \label{fig:memit_on_layer_14}
\end{figure*}

\begin{figure*}
\centering
  \includegraphics[width=\textwidth]{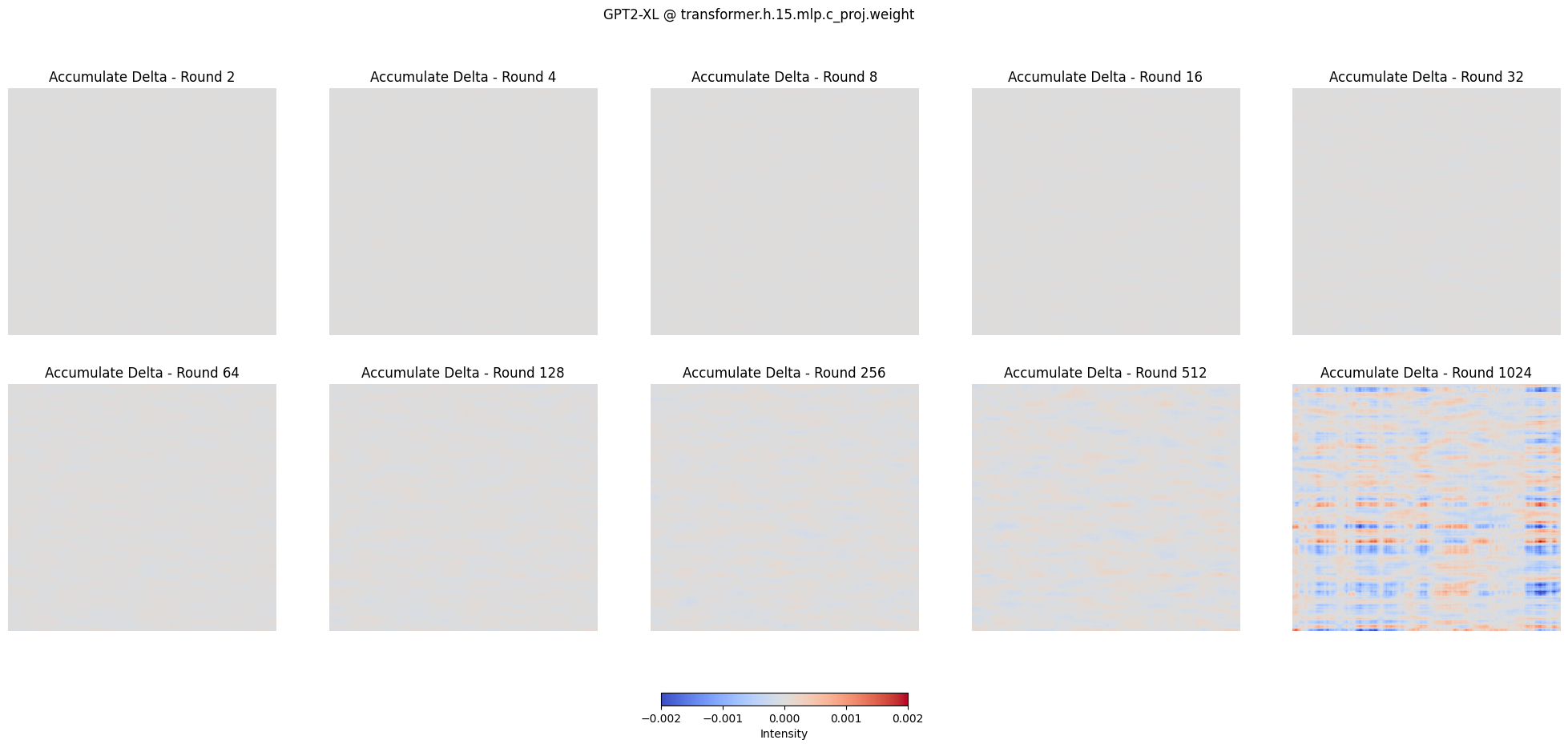}
  \caption{Toxicity on GPT2-XL on layer 15 using memit with editing steps.}
  \label{fig:memit_on_layer_15}
\end{figure*}

\begin{figure*}
\centering
  \includegraphics[width=\textwidth]{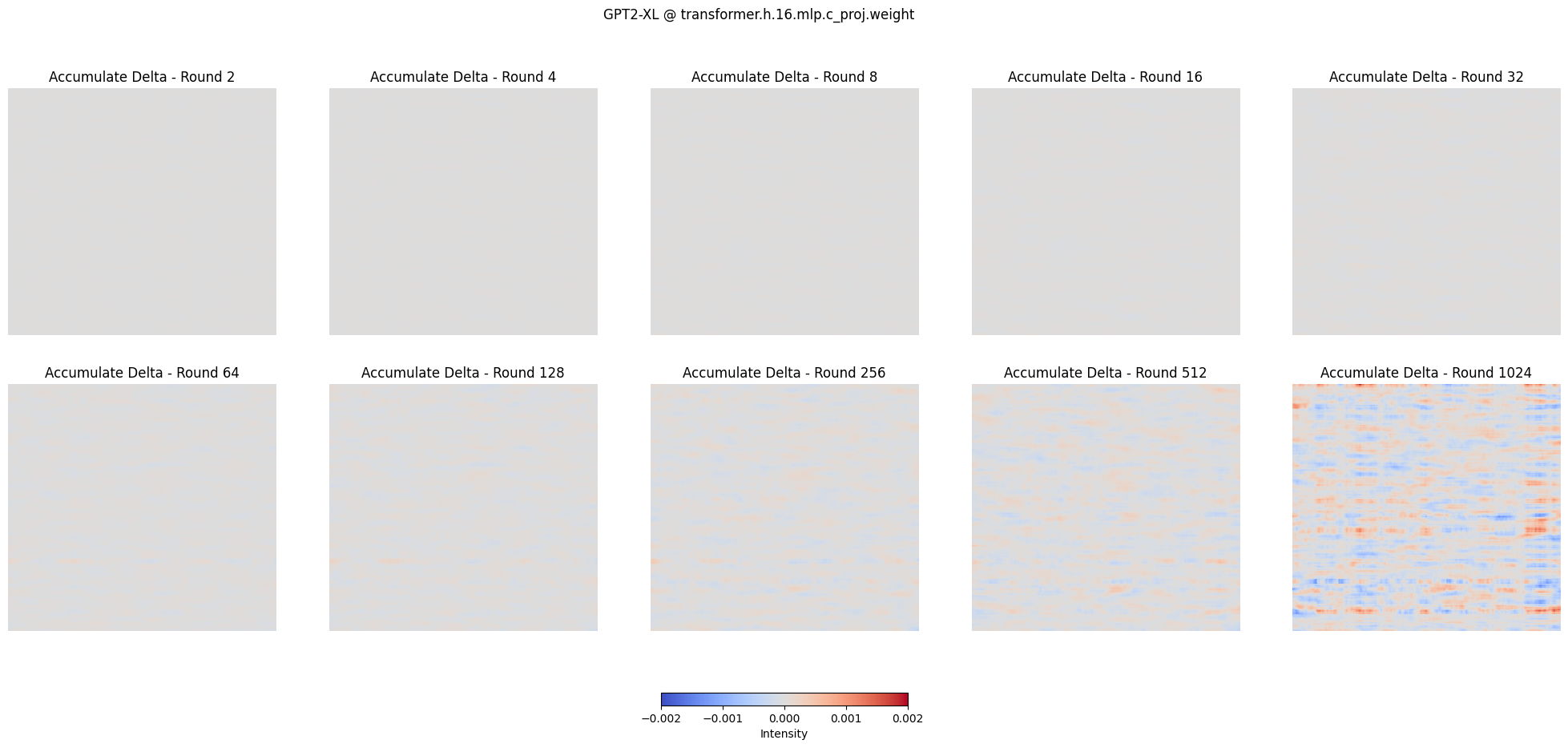}
  \caption{Toxicity on GPT2-XL on layer 16 using memit with editing steps.}
  \label{fig:memit_on_layer_16}
\end{figure*}

\begin{figure*}
\centering
  \includegraphics[width=\textwidth]{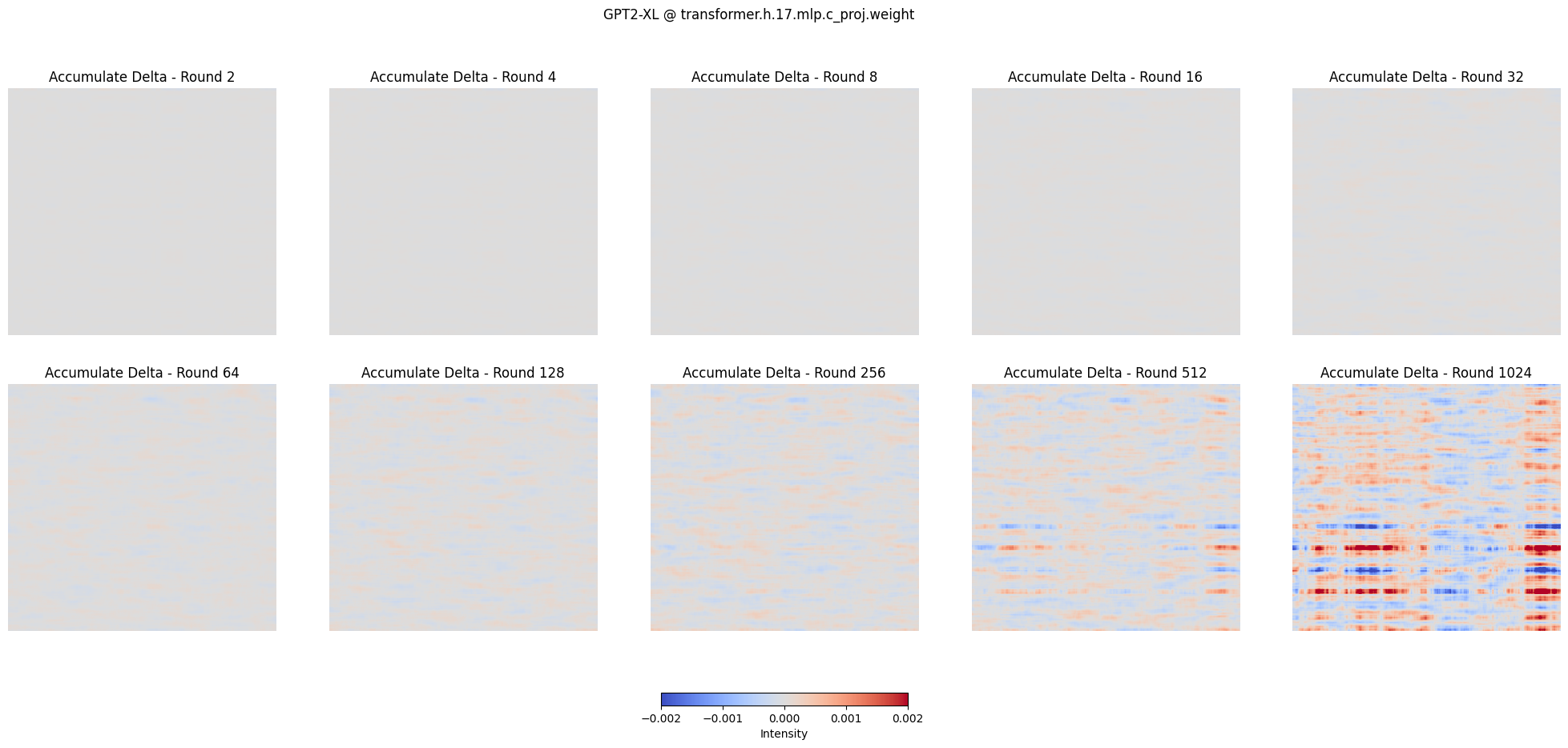}
  \caption{Toxicity on GPT2-XL on layer 17 using memit with editing steps.}
  \label{fig:memit_on_layer_17}
\end{figure*}

As editing progresses, the toxicity buildup effects within the predefined editing layers of MEMIT are illustrated in Figure~\ref{fig:memit_on_layer_13},~\ref{fig:memit_on_layer_14},~\ref{fig:memit_on_layer_15},~\ref{fig:memit_on_layer_16},~\ref{fig:memit_on_layer_17}. 

Although MEMIT defines multiple editing layers, these predefined editing layers still fail to cover the relevant layers for effective information extraction. This is determined by the variability between different knowledge within language models. Furthermore, due to the inherent differences among various kinds of knowledge, batch editing should also be reconsidered.

\section{Pattern Unmatch} \label{sec:appendix_pattern_unmatch}

In this section, we can proceed to a more formal description of pattern unmatch in Section~\ref{toxicity_flash}. This phenomenon occurs when there is partial data for which the activation value \(\sigma(\pmb x\cdot W_{fc}^l)\) in the predefined editing layer of ROME is extremely small (across several orders of magnitude, detailed results can be found in Appendix~\ref{sec:appendix_activation_strength}). However, in reality, the difference between $FFN(\pmb x)^l+\pmb\delta$ and other layers cannot be considered as the dominant factor (refer to detailed results in Appendix~\ref{sec:appendix_delta_strength}). Therefore, according to Equation~\ref{equ:update}, the extremely small activation value $\sigma(\pmb x\cdot W_{fc}^l)$ in the denominator becomes the primary cause of toxicity flash.

\subsection{Activation Strength} \label{sec:appendix_activation_strength}

\begin{figure*}
  \centering
  \subfigure[Activation strength over layers on case 3561.]{\includegraphics[width=0.2\textwidth]{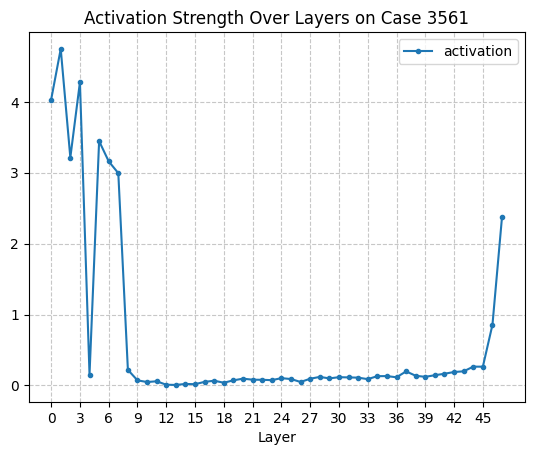}}
  \hfill
  \subfigure[Activation strength over layers on case 4661.]{\includegraphics[width=0.2\textwidth]{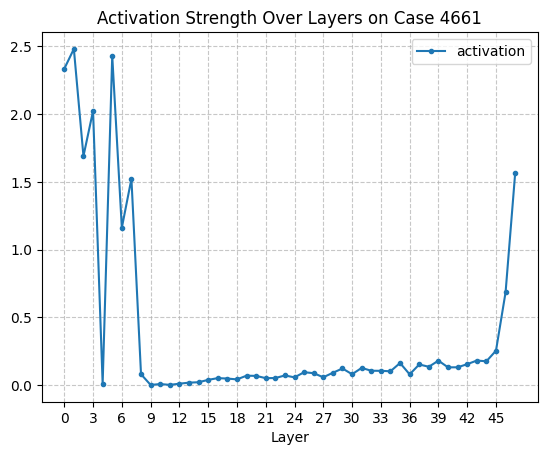}}
  \hfill
  \subfigure[Activation strength over layers on case 4790.]{\includegraphics[width=0.2\textwidth]{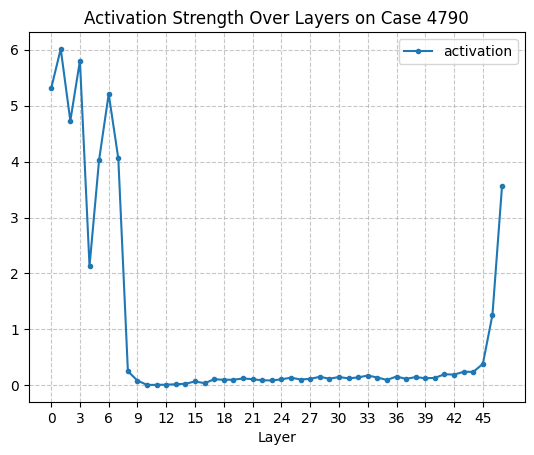}}
  \hfill
  \subfigure[Activation strength over layers on case 4988.]{\includegraphics[width=0.2\textwidth]{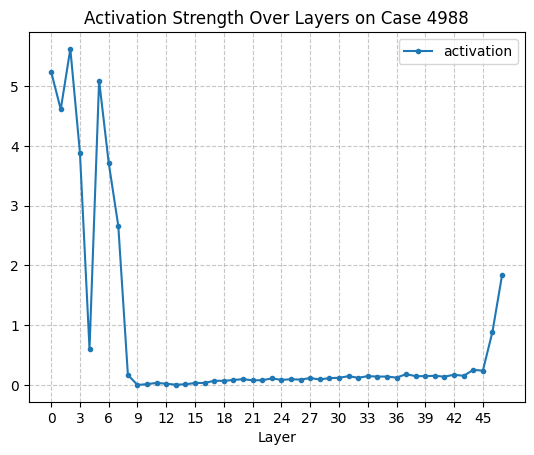}}
  \hfill
  \subfigure[Activation strength over layers on case 8793.]{\includegraphics[width=0.2\textwidth]{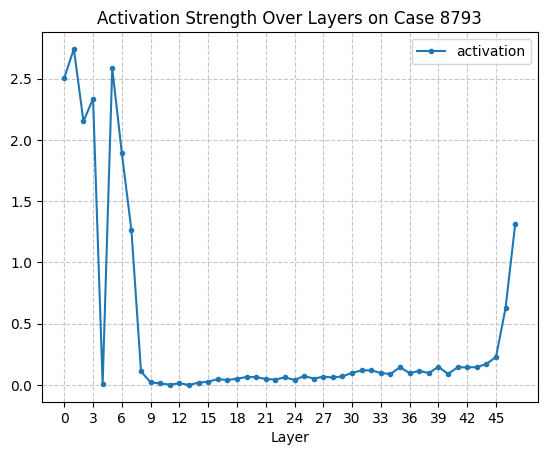}}
  \hfill
  \subfigure[Activation strength over layers on case 15452.]{\includegraphics[width=0.2\textwidth]{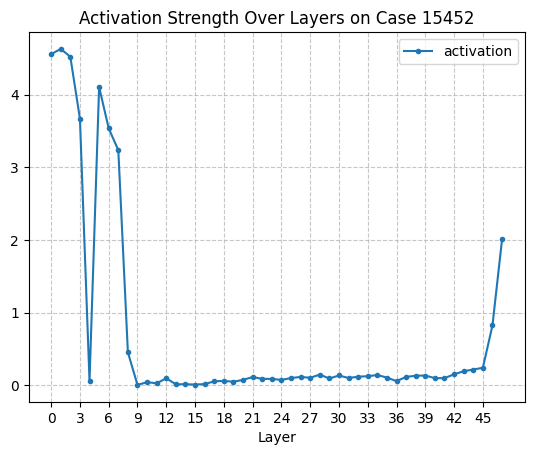}}
  \hfill
  \subfigure[Activation strength over layers on case 16575.]{\includegraphics[width=0.2\textwidth]{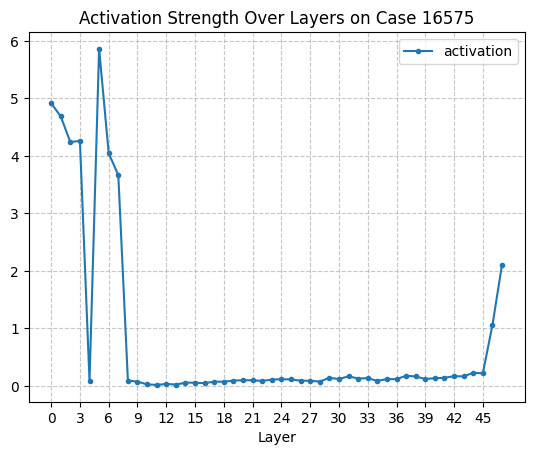}}
  \hfill
  \subfigure[Activation strength over layers on case 16781.]{\includegraphics[width=0.2\textwidth]{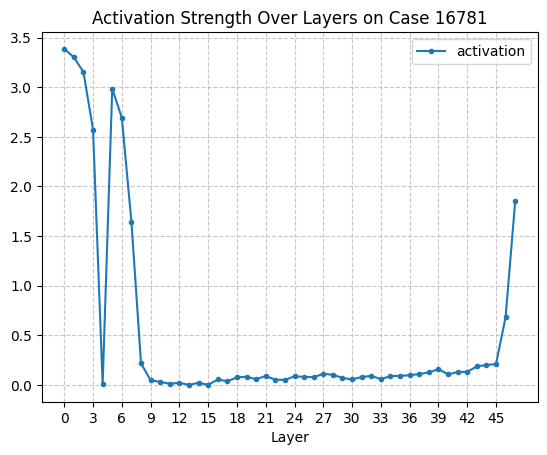}}

  \caption{Activation strength distribution on GPT2-XL among different layers.}
  \label{fig:activation_strength_on_gpt2-xl_flash}
\end{figure*}

The distribution of activation strength for data causing toxicity flash on GPT2-XL is depicted in Figure~\ref{fig:activation_strength_on_gpt2-xl_flash}.

\begin{figure*}
  \centering
  \subfigure[Activation strength over layers on case 0.]{\includegraphics[width=0.2\textwidth]{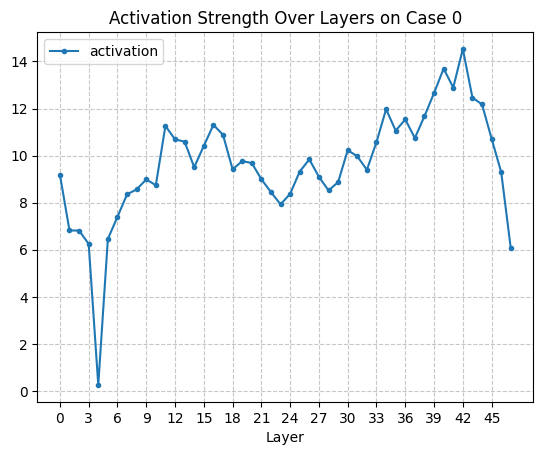}}
  \hfill
  \subfigure[Activation strength over layers on case 5.]{\includegraphics[width=0.2\textwidth]{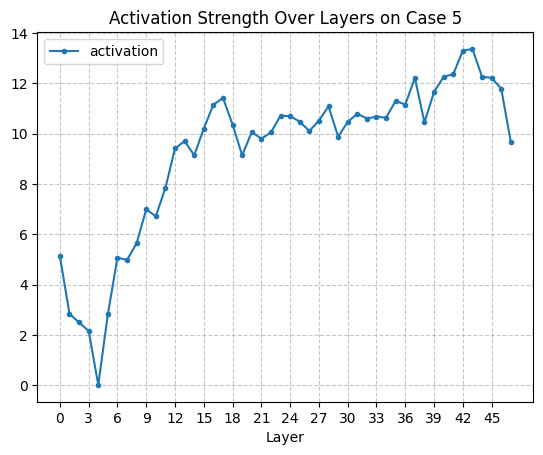}}
  \hfill
  \subfigure[Activation strength over layers on case 7.]{\includegraphics[width=0.2\textwidth]{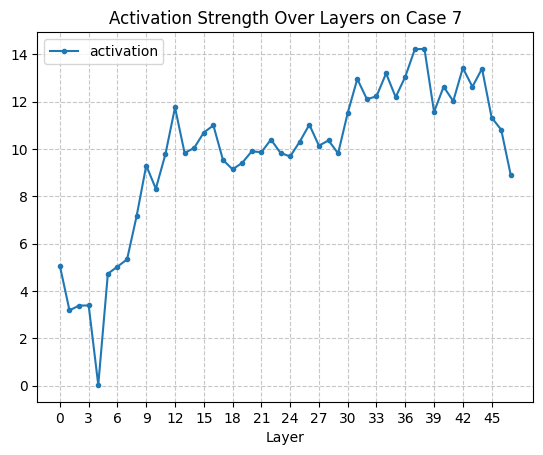}}
  \hfill
  \subfigure[Activation strength over layers on case 13.]{\includegraphics[width=0.2\textwidth]{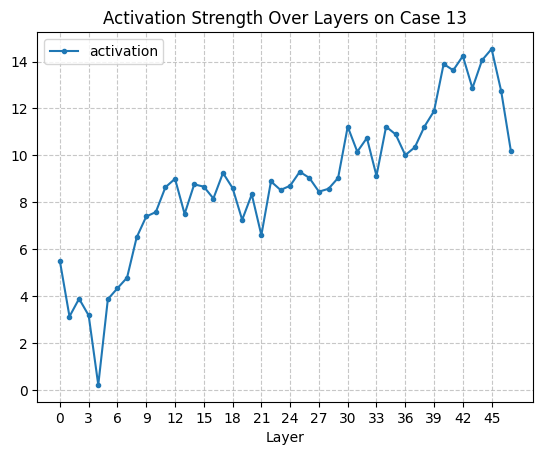}}
  \hfill
  \subfigure[Activation strength over layers on case 14.]{\includegraphics[width=0.2\textwidth]{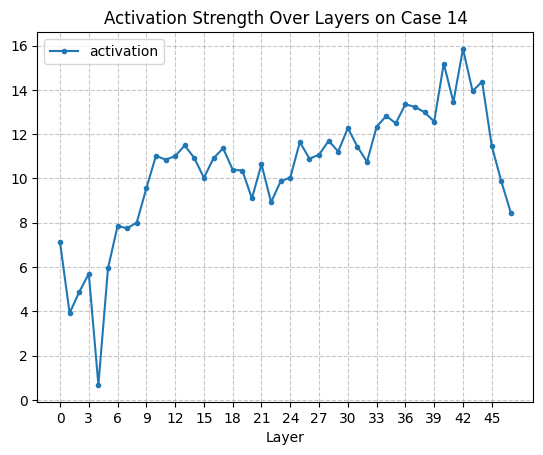}}
  \hfill
  \subfigure[Activation strength over layers on case 22.]{\includegraphics[width=0.2\textwidth]{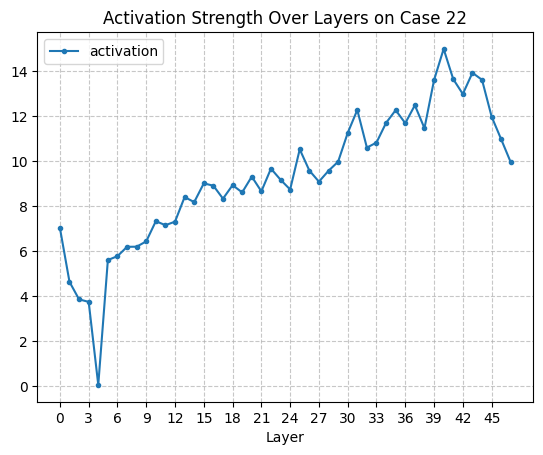}}
  \hfill
  \subfigure[Activation strength over layers on case 36.]{\includegraphics[width=0.2\textwidth]{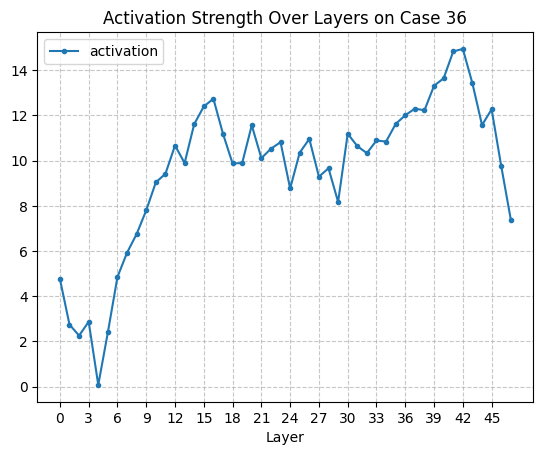}}
  \hfill
  \subfigure[Activation strength over layers on case 37.]{\includegraphics[width=0.2\textwidth]{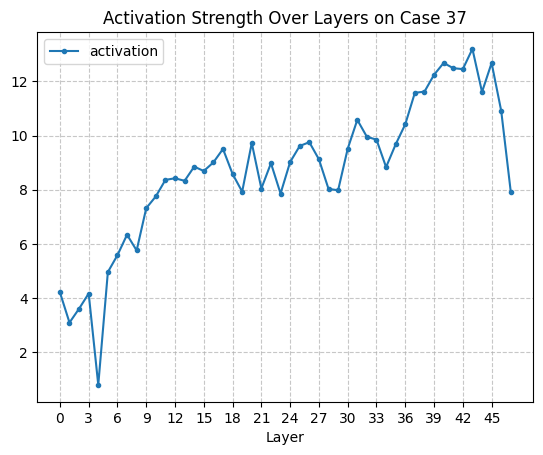}}

  \caption{Activation strength distribution on GPT2-XL among different layers.}
  \label{fig:activation_strength_on_gpt2-xl_other}
\end{figure*}

The distribution of activation strength for other data on GPT2-XL is shown in Figure~\ref{fig:activation_strength_on_gpt2-xl_other}.

\begin{figure*}
  \centering
  \subfigure[Activation strength over layers on case 3561.]{\includegraphics[width=0.2\textwidth]{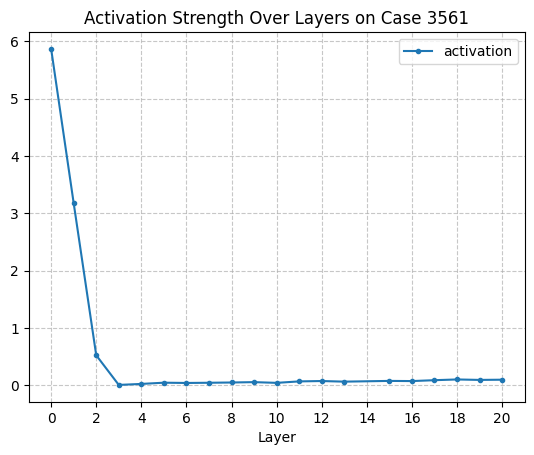}}
  \hfill
  \subfigure[Activation strength over layers on case 4661.]{\includegraphics[width=0.2\textwidth]{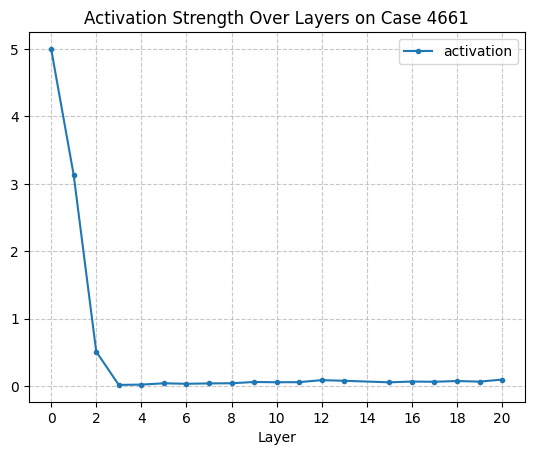}}
  \hfill
  \subfigure[Activation strength over layers on case 4988.]{\includegraphics[width=0.2\textwidth]{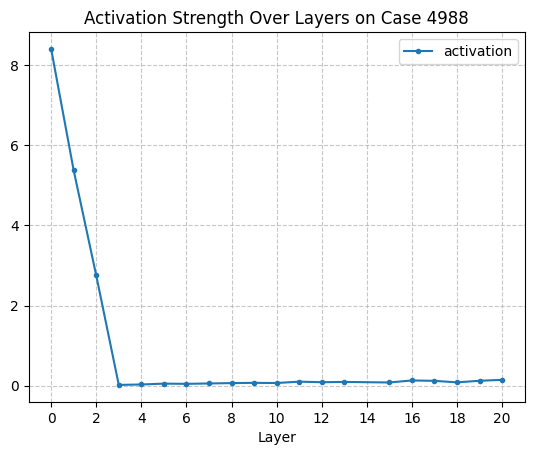}}
  \hfill
  \subfigure[Activation strength over layers on case 8475.]{\includegraphics[width=0.2\textwidth]{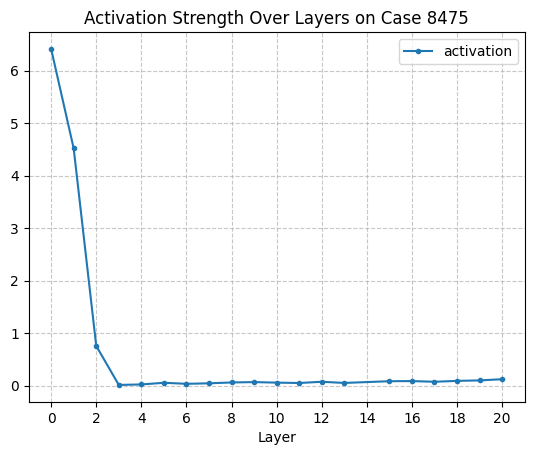}}
  \hfill
  \subfigure[Activation strength over layers on case 8793.]{\includegraphics[width=0.2\textwidth]{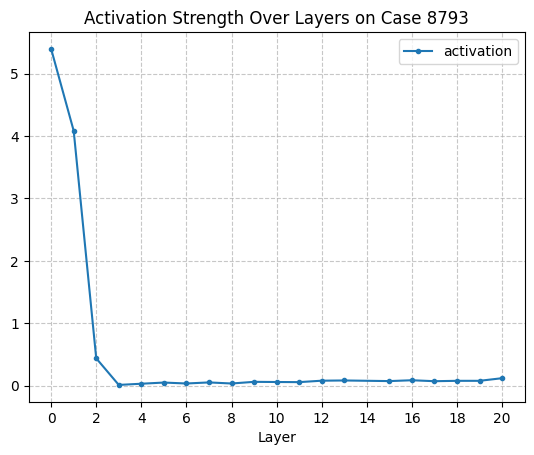}}
  \hfill
  \subfigure[Activation strength over layers on case 15452.]{\includegraphics[width=0.2\textwidth]{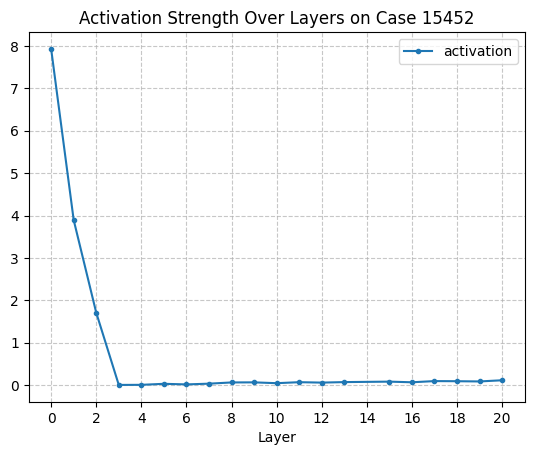}}
  \hfill
  \subfigure[Activation strength over layers on case 16575.]{\includegraphics[width=0.2\textwidth]{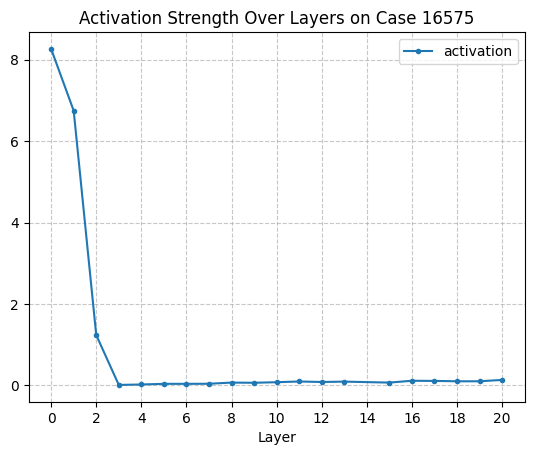}}
  \hfill
  \subfigure[Activation strength over layers on case 16781.]{\includegraphics[width=0.2\textwidth]{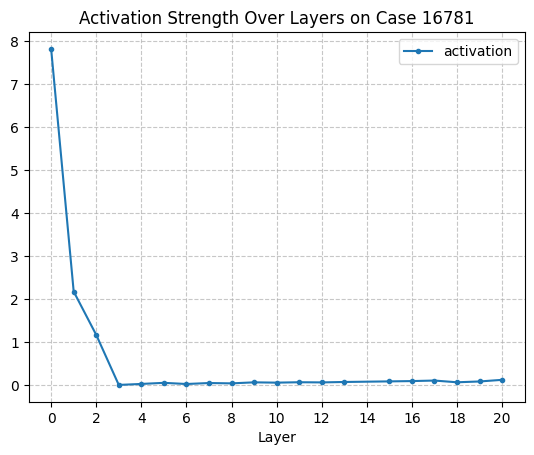}}

  \caption{Activation strength distribution on GPT-J among different layers.}
  \label{fig:activation_strength_on_gpt-j_flash}
\end{figure*}

The distribution of activation strength for data causing toxicity flash on GPT-J is depicted in Figure~\ref{fig:activation_strength_on_gpt-j_flash}.

\begin{figure*}
  \centering
  \subfigure[Activation strength over layers on case 0.]{\includegraphics[width=0.2\textwidth]{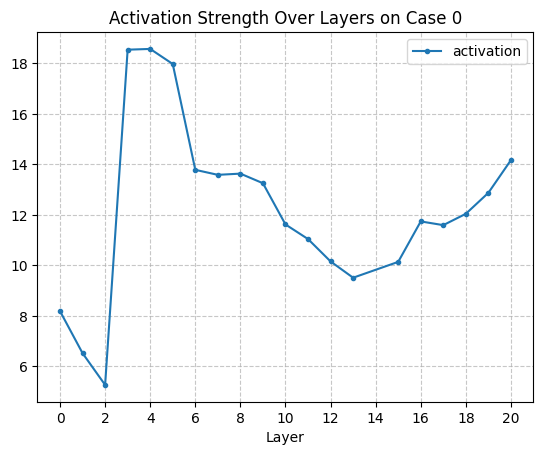}}
  \hfill
  \subfigure[Activation strength over layers on case 5.]{\includegraphics[width=0.2\textwidth]{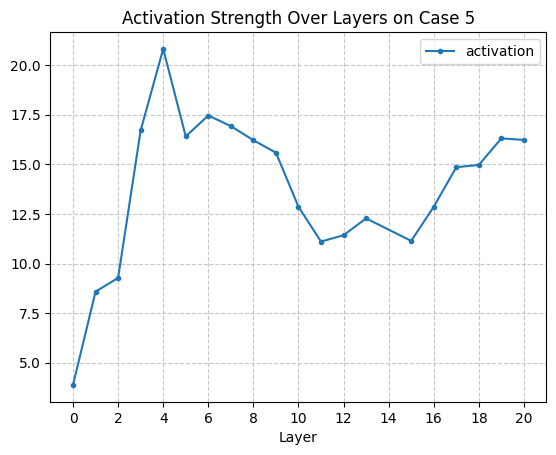}}
  \hfill
  \subfigure[Activation strength over layers on case 7.]{\includegraphics[width=0.2\textwidth]{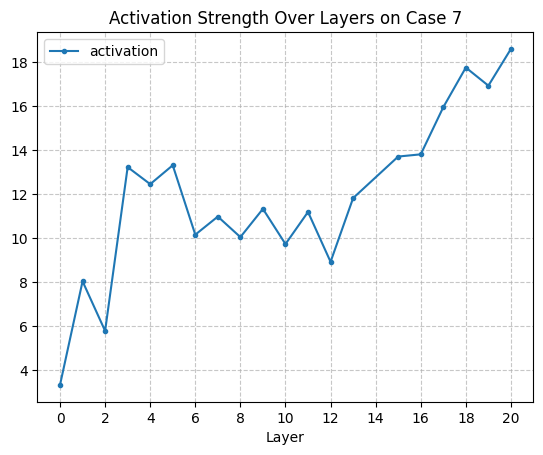}}
  \hfill
  \subfigure[Activation strength over layers on case 14.]{\includegraphics[width=0.2\textwidth]{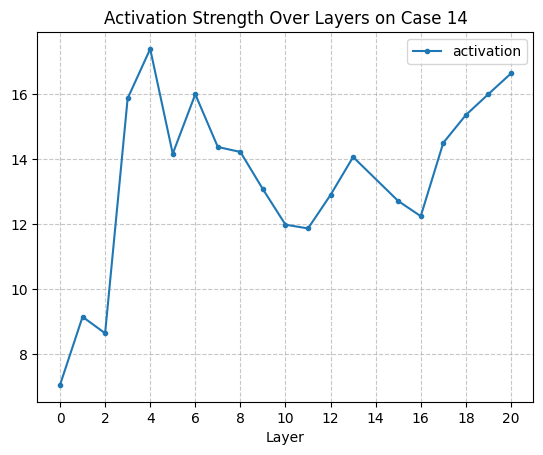}}
  \hfill
  \subfigure[Activation strength over layers on case 29.]{\includegraphics[width=0.2\textwidth]{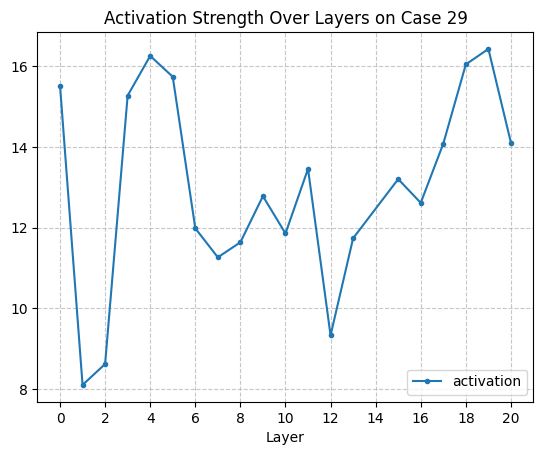}}
  \hfill
  \subfigure[Activation strength over layers on case 52.]{\includegraphics[width=0.2\textwidth]{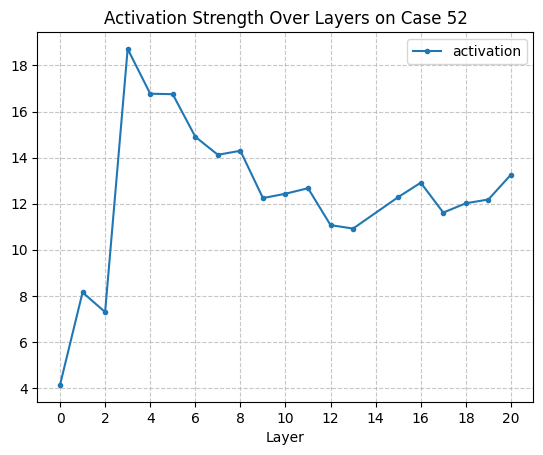}}
  \hfill
  \subfigure[Activation strength over layers on case 54.]{\includegraphics[width=0.2\textwidth]{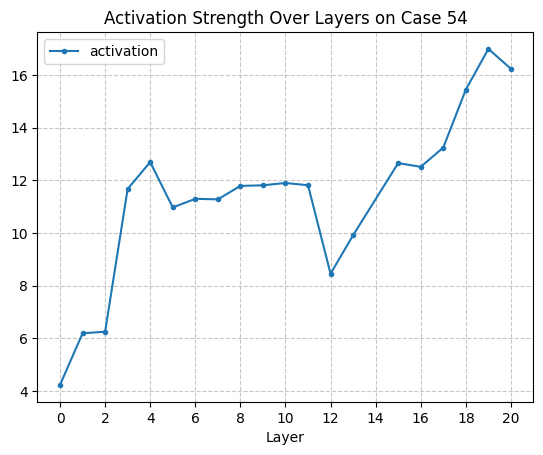}}
  \hfill
  \subfigure[Activation strength over layers on case 56.]{\includegraphics[width=0.2\textwidth]{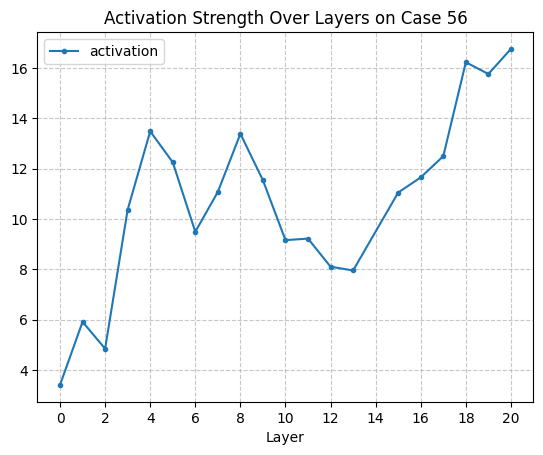}}

  \caption{Activation strength distribution on GPT-J among different layers.}
  \label{fig:activation_strength_on_gpt-j_other}
\end{figure*}

The distribution of activation strength for other data on GPT-J is shown in Figure~\ref{fig:activation_strength_on_gpt-j_other}.

\subsection{Delta Strength} \label{sec:appendix_delta_strength}

\begin{figure*}
  \centering
  \subfigure[Delta strength over layers on case 3561.]{\includegraphics[width=0.2\textwidth]{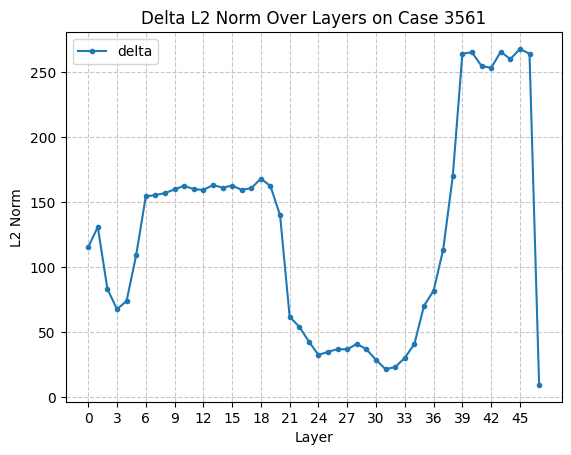}}
  \hfill
  \subfigure[Delta strength over layers on case 4661.]{\includegraphics[width=0.2\textwidth]{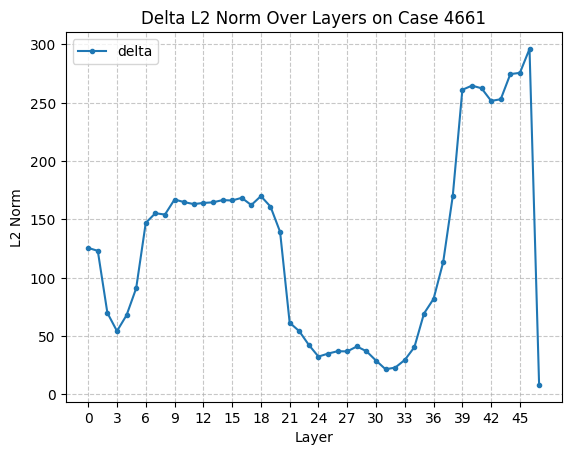}}
  \hfill
  \subfigure[Delta strength over layers on case 4790.]{\includegraphics[width=0.2\textwidth]{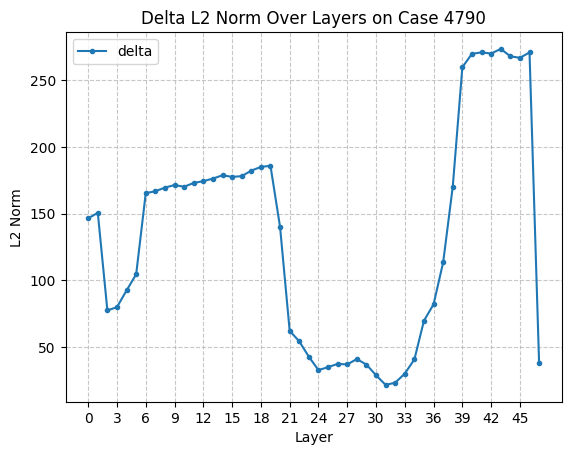}}
  \hfill
  \subfigure[Delta strength over layers on case 4988.]{\includegraphics[width=0.2\textwidth]{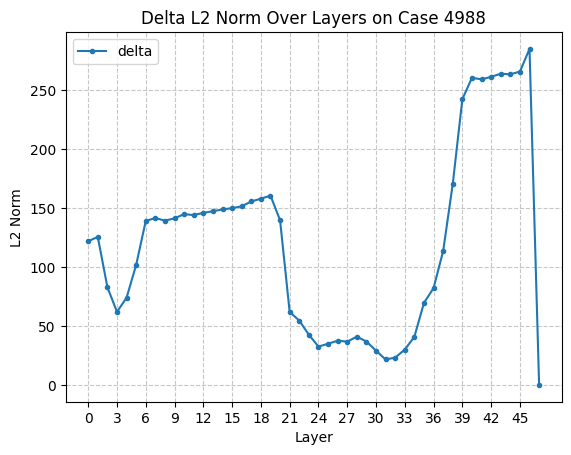}}
  \hfill
  \subfigure[Delta strength over layers on case 8793.]{\includegraphics[width=0.2\textwidth]{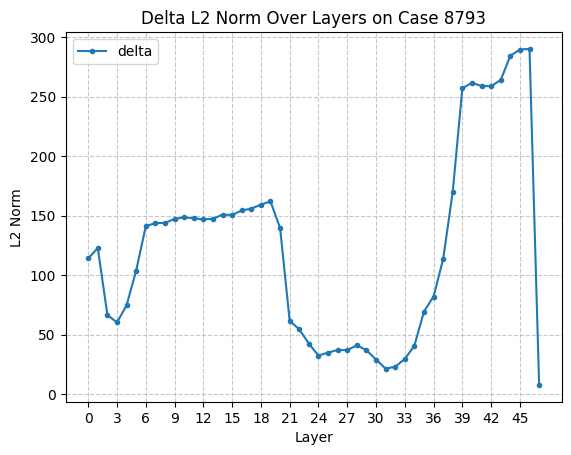}}
  \hfill
  \subfigure[Delta strength over layers on case 15452.]{\includegraphics[width=0.2\textwidth]{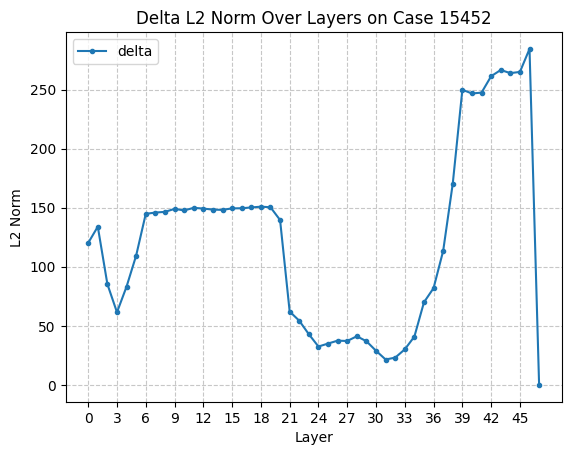}}
  \hfill
  \subfigure[Delta strength over layers on case 16575.]{\includegraphics[width=0.2\textwidth]{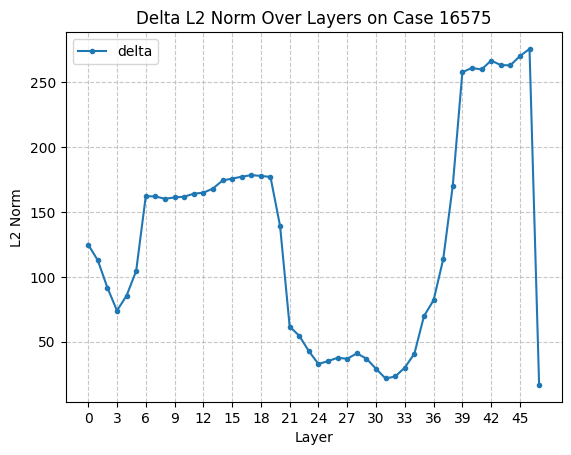}}
  \hfill
  \subfigure[Delta strength over layers on case 16781.]{\includegraphics[width=0.2\textwidth]{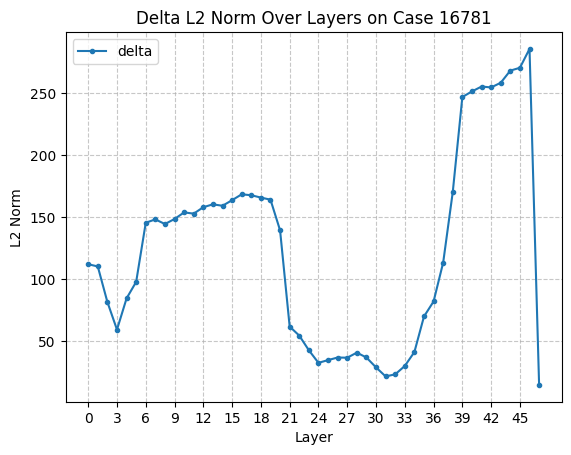}}

  \caption{Delta strength distribution on GPT2-XL among different layers.}
  \label{fig:delta_strength_on_gpt2-xl_flash}
\end{figure*}

The distribution of delta strength for data causing toxicity flash on GPT2-XL is depicted in Figure~\ref{fig:delta_strength_on_gpt2-xl_flash}.

\begin{figure*}
  \centering
  \subfigure[Delta strength over layers on case 0.]{\includegraphics[width=0.2\textwidth]{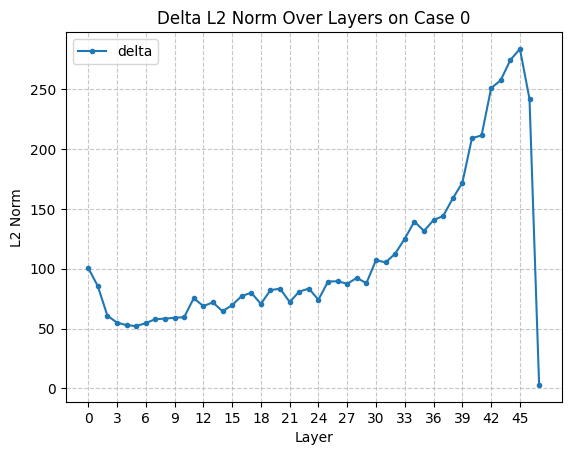}}
  \hfill
  \subfigure[Delta strength over layers on case 5.]{\includegraphics[width=0.2\textwidth]{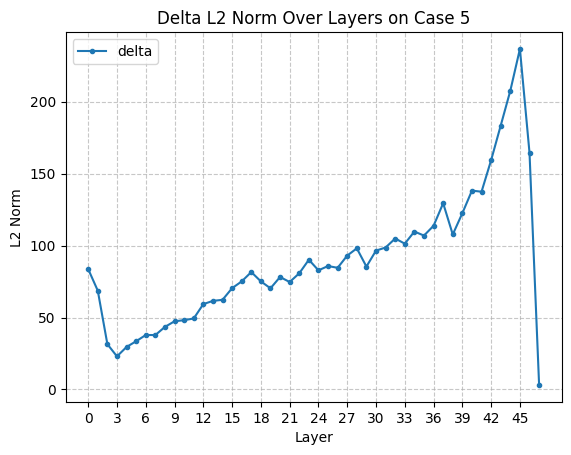}}
  \hfill
  \subfigure[Delta strength over layers on case 7.]{\includegraphics[width=0.2\textwidth]{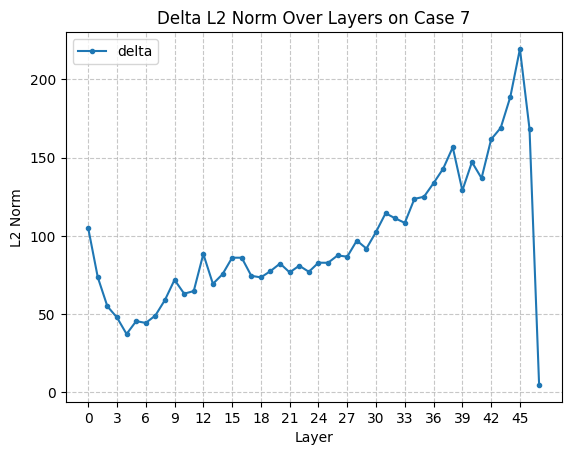}}
  \hfill
  \subfigure[Delta strength over layers on case 13.]{\includegraphics[width=0.2\textwidth]{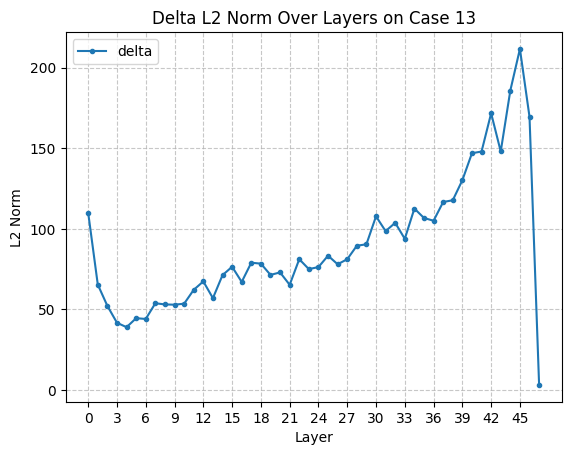}}
  \hfill
  \subfigure[Delta strength over layers on case 14.]{\includegraphics[width=0.2\textwidth]{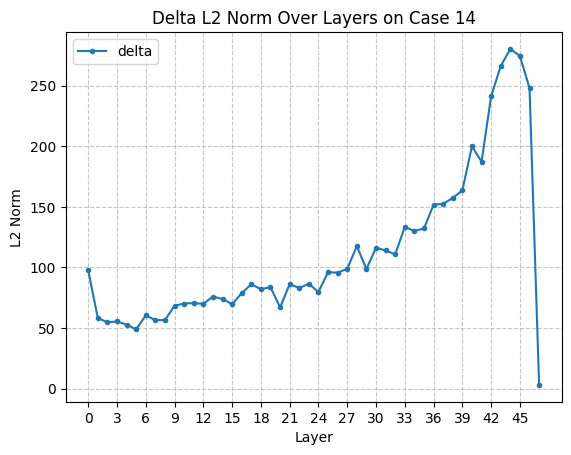}}
  \hfill
  \subfigure[Delta strength over layers on case 22.]{\includegraphics[width=0.2\textwidth]{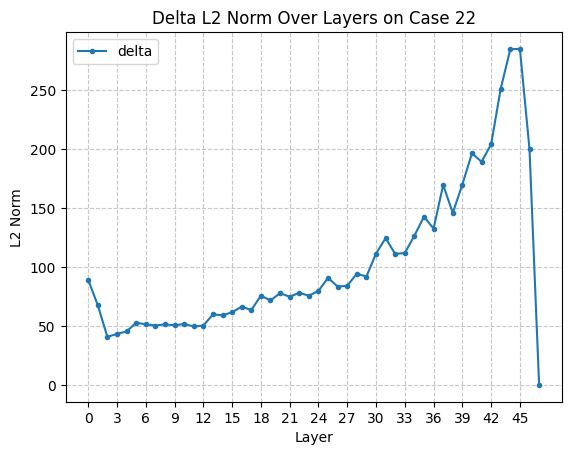}}
  \hfill
  \subfigure[Delta strength over layers on case 36.]{\includegraphics[width=0.2\textwidth]{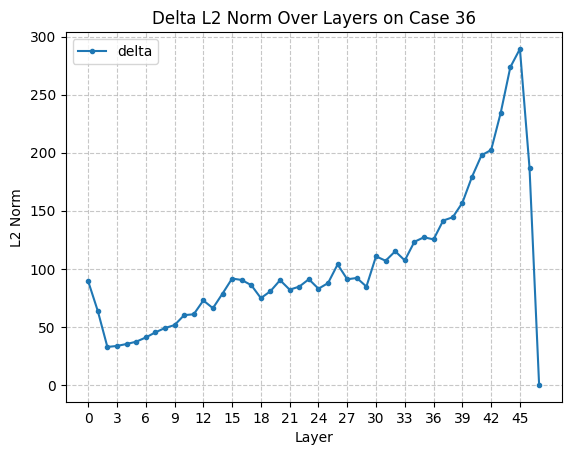}}
  \hfill
  \subfigure[Delta strength over layers on case 37.]{\includegraphics[width=0.2\textwidth]{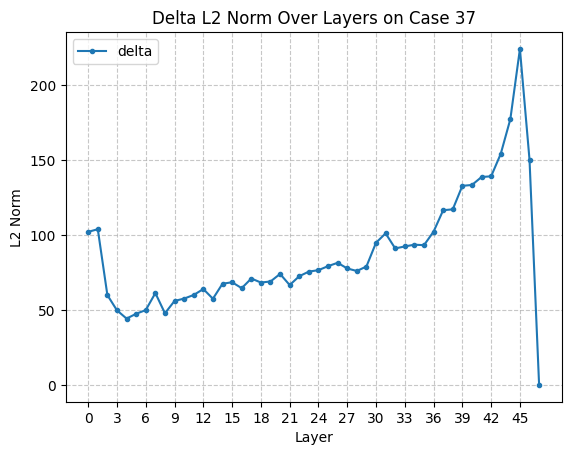}}

  \caption{Delta strength distribution on GPT2-XL among different layers.}
  \label{fig:delta_strength_on_gpt2-xl_other}
\end{figure*}

The distribution of delta strength for other data on GPT2-XL is shown in Figure~\ref{fig:delta_strength_on_gpt2-xl_other}.

\begin{figure*}
  \centering
  \subfigure[Delta strength over layers on case 3561.]{\includegraphics[width=0.2\textwidth]{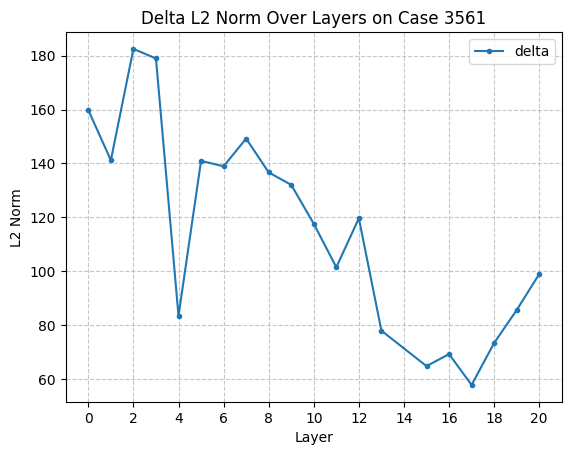}}
  \hfill
  \subfigure[Delta strength over layers on case 4661.]{\includegraphics[width=0.2\textwidth]{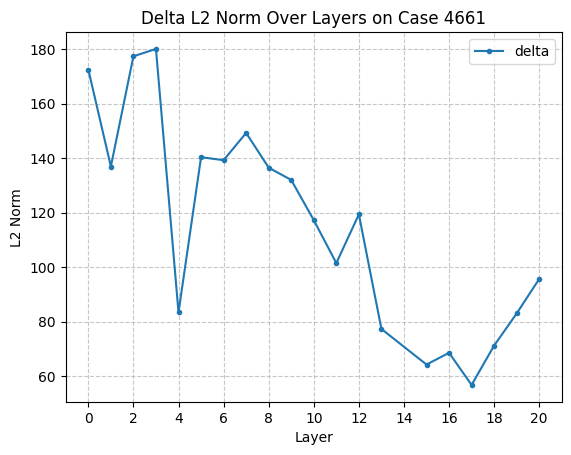}}
  \hfill
  \subfigure[Delta strength over layers on case 4988.]{\includegraphics[width=0.2\textwidth]{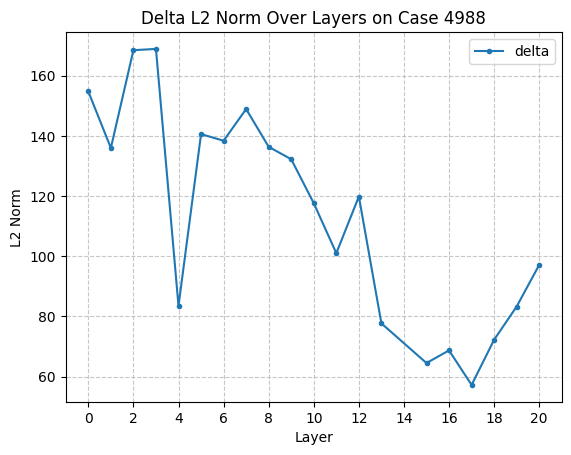}}
  \hfill
  \subfigure[Delta strength over layers on case 8475.]{\includegraphics[width=0.2\textwidth]{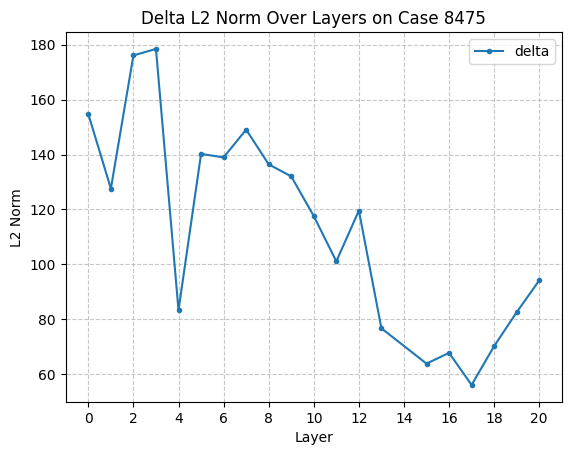}}
  \hfill
  \subfigure[Delta strength over layers on case 8793.]{\includegraphics[width=0.2\textwidth]{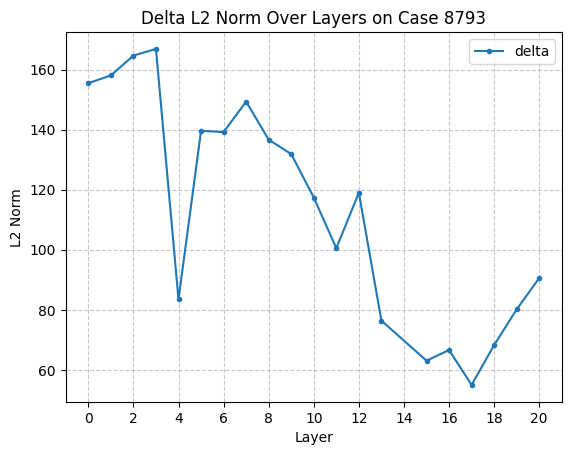}}
  \hfill
  \subfigure[Delta strength over layers on case 15452.]{\includegraphics[width=0.2\textwidth]{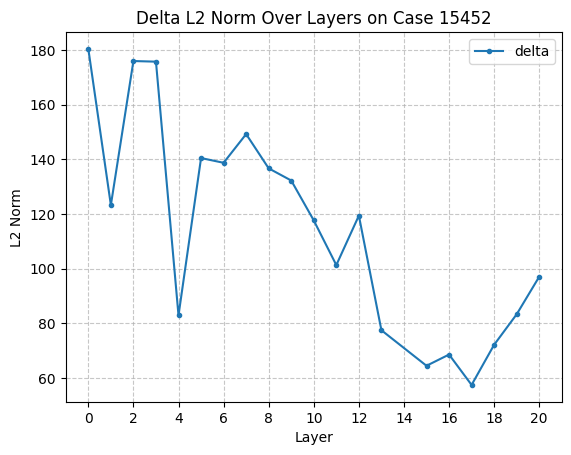}}
  \hfill
  \subfigure[Delta strength over layers on case 16575.]{\includegraphics[width=0.2\textwidth]{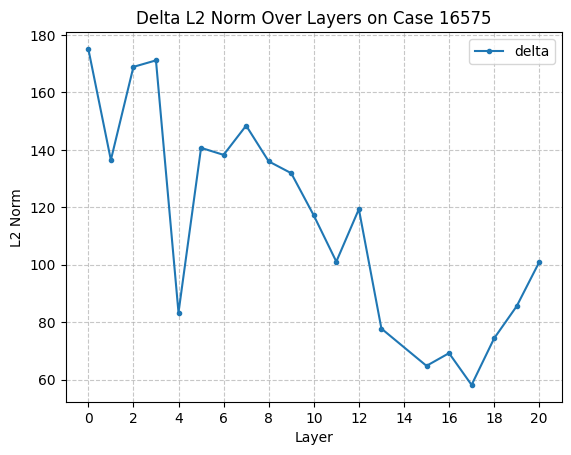}}
  \hfill
  \subfigure[Delta strength over layers on case 16781.]{\includegraphics[width=0.2\textwidth]{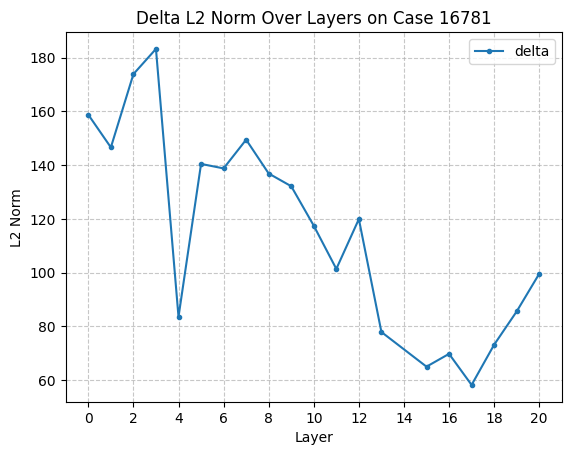}}

  \caption{Delta strength distribution on GPT-J among different layers.}
  \label{fig:delta_strength_on_gpt-j_flash}
\end{figure*}

The distribution of delta strength for data causing toxicity flash on GPT-J is depicted in Figure~\ref{fig:delta_strength_on_gpt-j_flash}.

\begin{figure*}
  \centering
  \subfigure[Delta strength over layers on case 0.]{\includegraphics[width=0.2\textwidth]{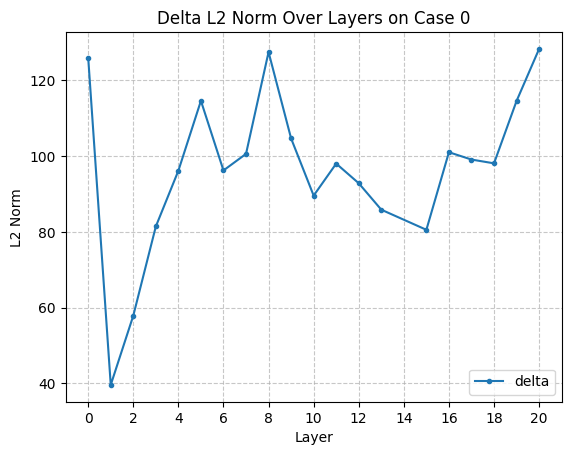}}
  \hfill
  \subfigure[Delta strength over layers on case 5.]{\includegraphics[width=0.2\textwidth]{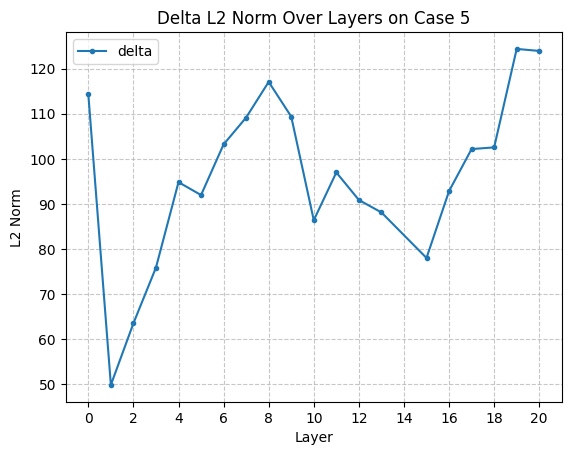}}
  \hfill
  \subfigure[Delta strength over layers on case 7.]{\includegraphics[width=0.2\textwidth]{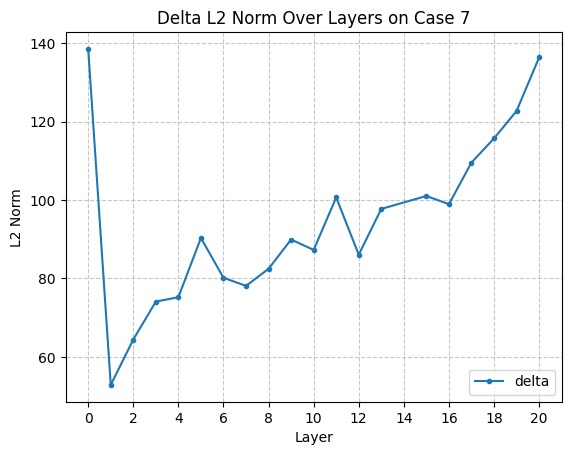}}
  \hfill
  \subfigure[Delta strength over layers on case 14.]{\includegraphics[width=0.2\textwidth]{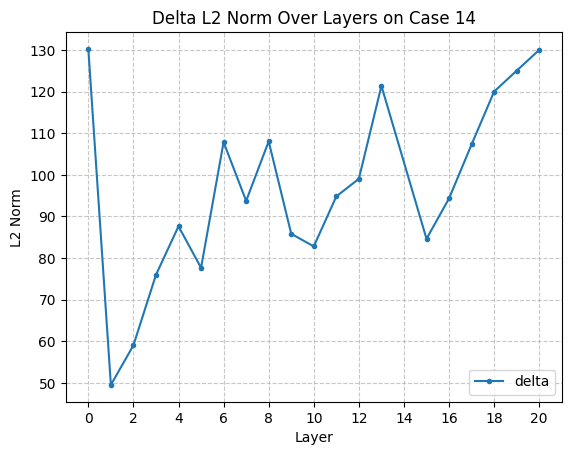}}
  \hfill
  \subfigure[Delta strength over layers on case 29.]{\includegraphics[width=0.2\textwidth]{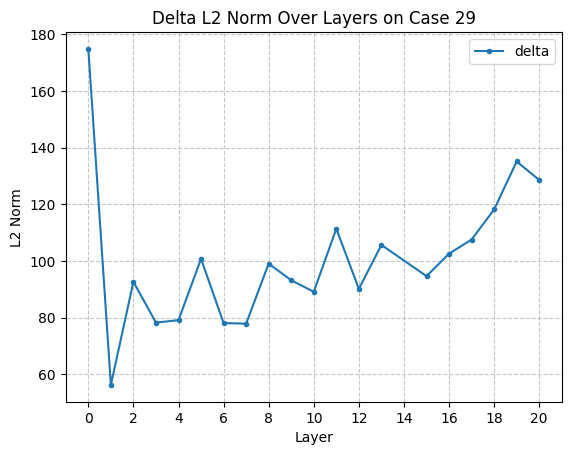}}
  \hfill
  \subfigure[Delta strength over layers on case 52.]{\includegraphics[width=0.2\textwidth]{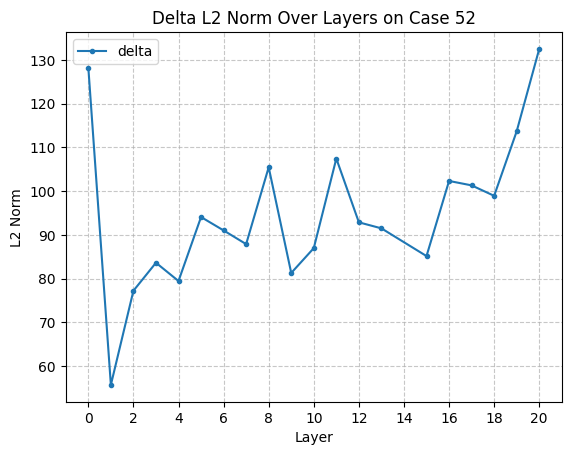}}
  \hfill
  \subfigure[Delta strength over layers on case 54.]{\includegraphics[width=0.2\textwidth]{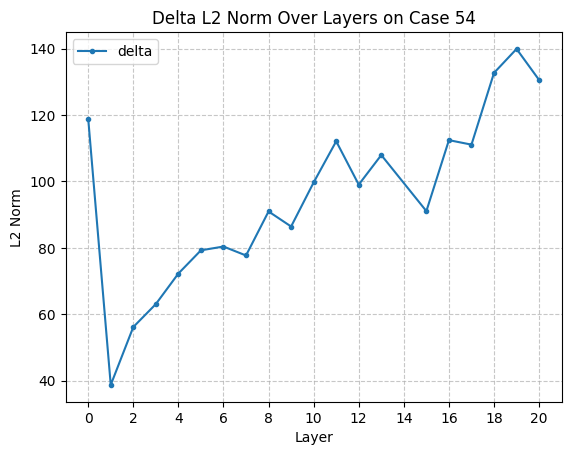}}
  \hfill
  \subfigure[Delta strength over layers on case 56.]{\includegraphics[width=0.2\textwidth]{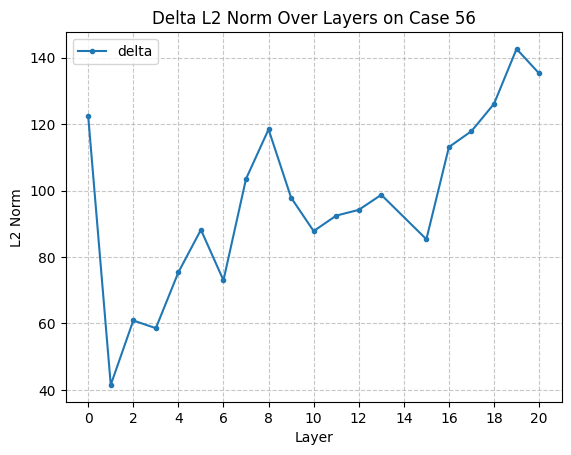}}

  \caption{Delta strength distribution on GPT-J among different layers.}
  \label{fig:delta_strength_on_gpt-j_other}
\end{figure*}

The distribution of delta strength for other data on GPT-J is shown in Figure~\ref{fig:delta_strength_on_gpt-j_other}.

\section{More Edit Analysis on Toxicity Flash } \label{sec:appendix_more_edit_analysis_on_toxicity_flash}

In this section, we present the experimental results on additional data described in Section~\ref{toxicity_flash}.

\begin{figure*}
  \centering
  \subfigure[Toxicity distribution on case 3561.]{\includegraphics[width=0.2\textwidth]{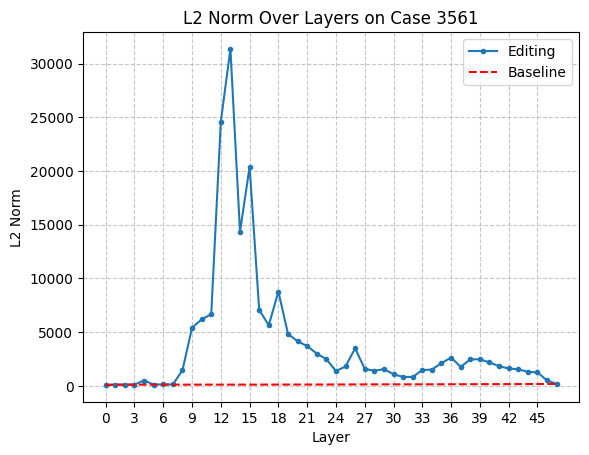}}
  \hfill
  \subfigure[Toxicity distribution on case 4661.]{\includegraphics[width=0.2\textwidth]{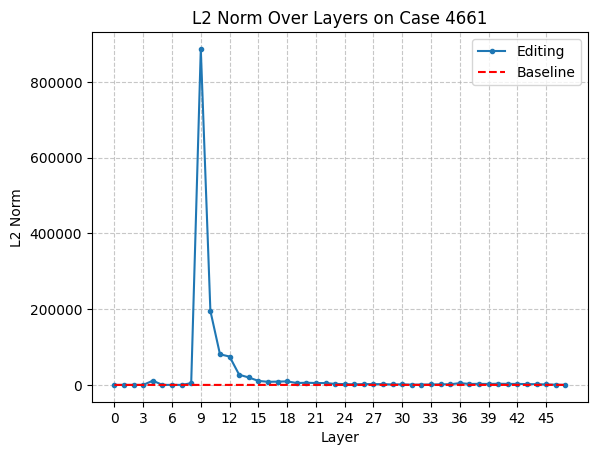}}
  \hfill
  \subfigure[Toxicity distribution on case 4790.]{\includegraphics[width=0.2\textwidth]{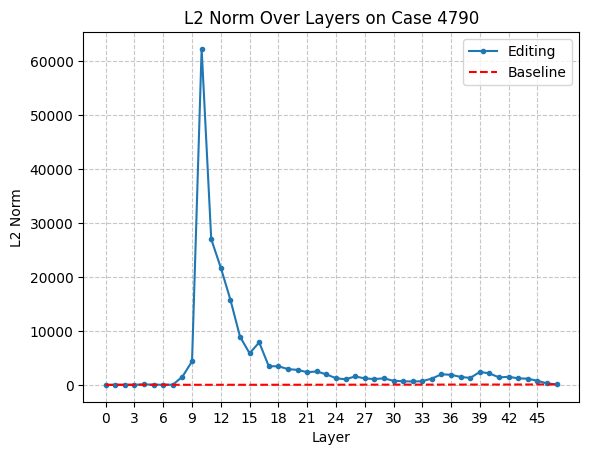}}
  \hfill
  \subfigure[Toxicity distribution on case 4988.]{\includegraphics[width=0.2\textwidth]{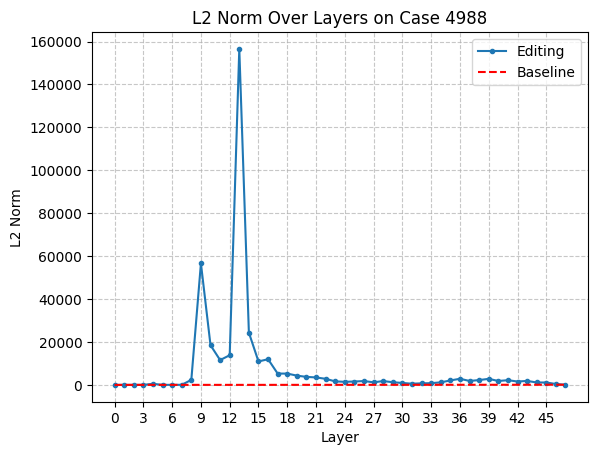}}
  \hfill
  \subfigure[Toxicity distribution on case 8793.]{\includegraphics[width=0.2\textwidth]{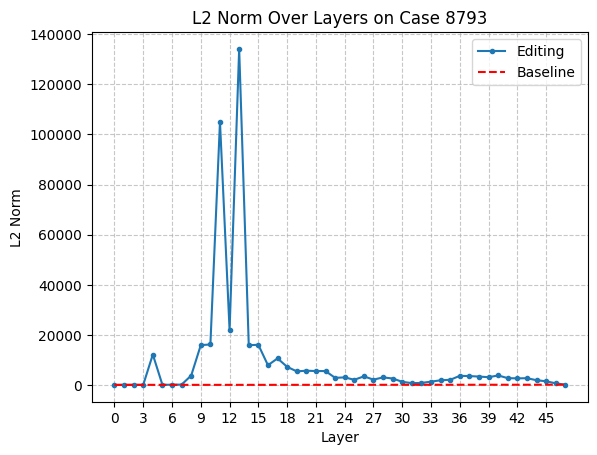}}
  \hfill
  \subfigure[Toxicity distribution on case 15452.]{\includegraphics[width=0.2\textwidth]{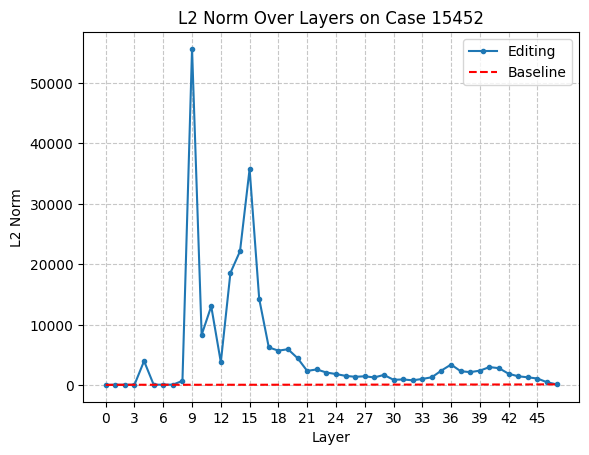}}
  \hfill
  \subfigure[Toxicity distribution on case 16575.]{\includegraphics[width=0.2\textwidth]{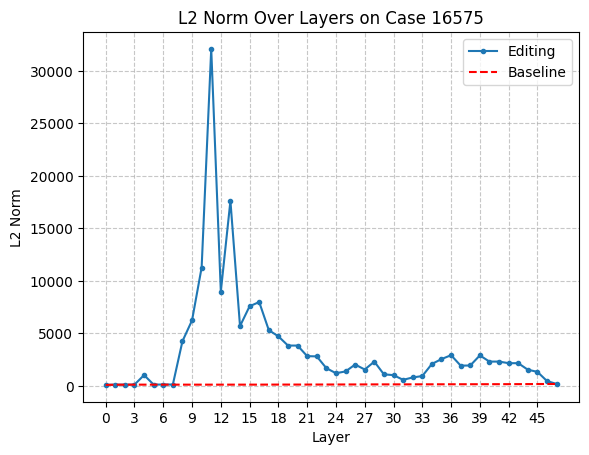}}
  \hfill
  \subfigure[Toxicity distribution on case 16781.]{\includegraphics[width=0.2\textwidth]{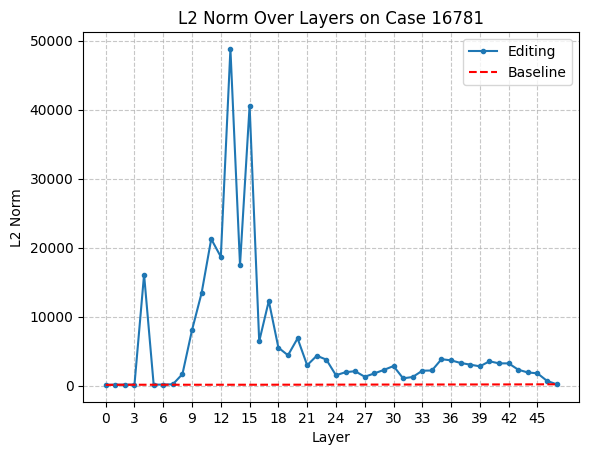}}

  \caption{Toxicity distribution on GPT2-XL among different layers. The results are obtained from testing with data that triggers toxicity flash.}
  \label{fig:toxicity_distribution_on_flash_data_gpt2-xl}
\end{figure*}

The distribution of toxicity across various layers during the editing of GPT2-XL, leading to toxicity flash, is depicted in Figure~\ref{fig:toxicity_distribution_on_flash_data_gpt2-xl}.

\begin{figure*}
  \centering
  \subfigure[Toxicity distribution on case 0.]{\includegraphics[width=0.2\textwidth]{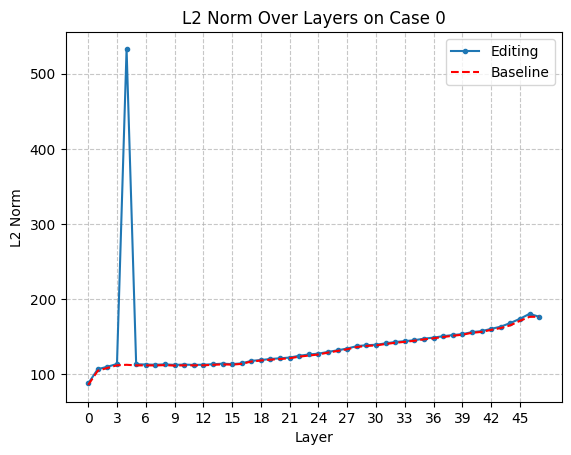}
  }
  \hfill
  \subfigure[Toxicity distribution on case 5.]{\includegraphics[width=0.2\textwidth]{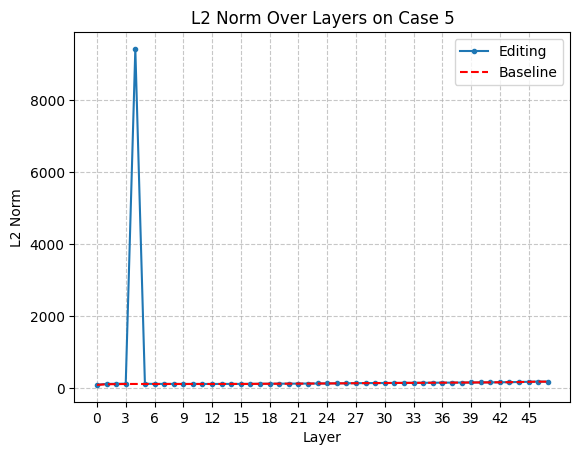}}
  \hfill
  \subfigure[Toxicity distribution on case 7.]{\includegraphics[width=0.2\textwidth]{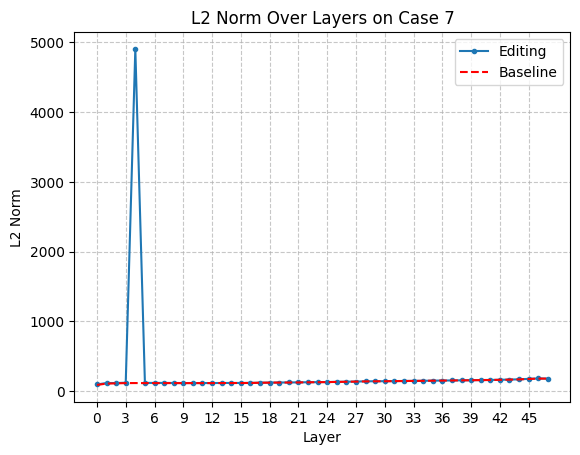}}
  \hfill
  \subfigure[Toxicity distribution on case 13.]{\includegraphics[width=0.2\textwidth]{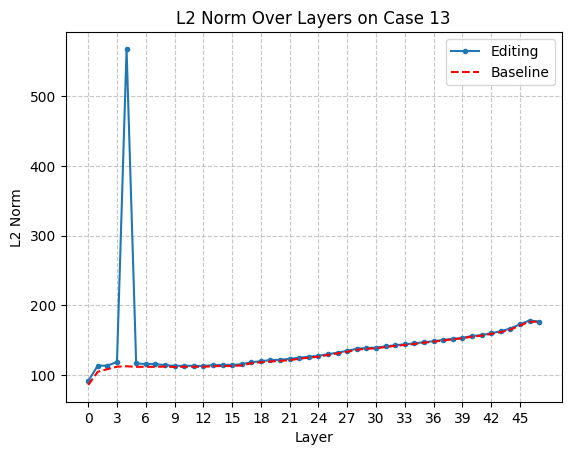}}
  \hfill
  \subfigure[Toxicity distribution on case 22.]{\includegraphics[width=0.2\textwidth]{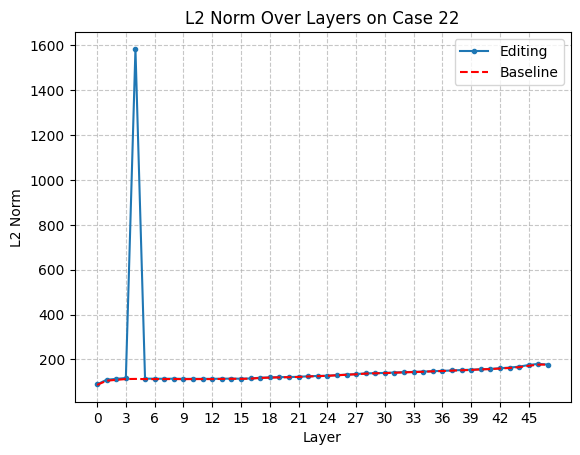}}
  \hfill
  \subfigure[Toxicity distribution on case 36.]{\includegraphics[width=0.2\textwidth]{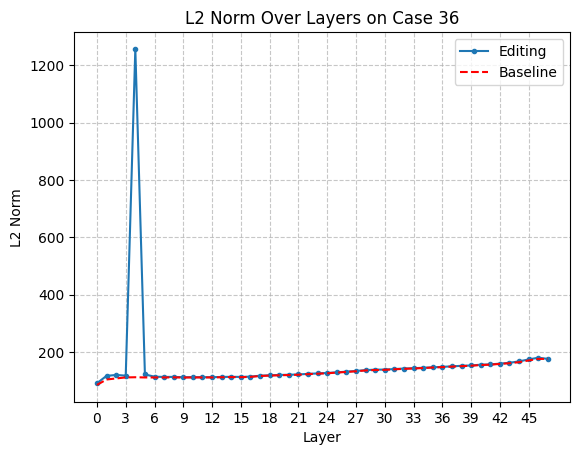}}
  \hfill
  \subfigure[Toxicity distribution on case 37.]{\includegraphics[width=0.2\textwidth]{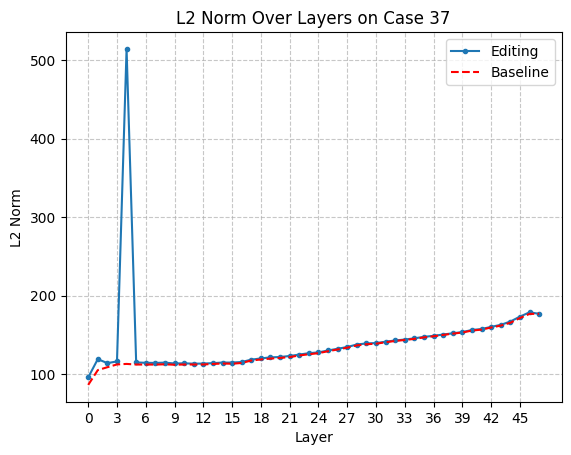}}
  \hfill
  \subfigure[Toxicity distribution on case 48.]{\includegraphics[width=0.2\textwidth]{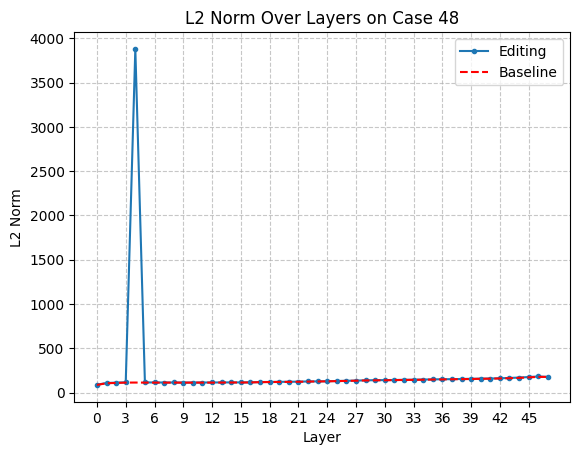}}

  \caption{Toxicity distribution on GPT2-XL among different layers. The results are obtained from testing with other normal data.}
  \label{fig:toxicity_distribution_on_normal_data_gpt2-xl}
\end{figure*}

The distribution of toxicity across various layers during the editing of GPT2-XL, not leading to toxicity flash, is depicted in Figure~\ref{fig:toxicity_distribution_on_normal_data_gpt2-xl}.

\begin{figure*}
  \centering
  \subfigure[Toxicity distribution on case 3561.]{\includegraphics[width=0.2\textwidth]{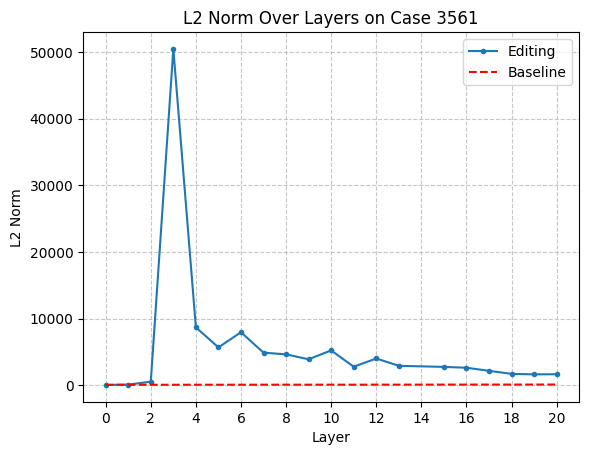}}
  \hfill
  \subfigure[Toxicity distribution on case 4661.]{\includegraphics[width=0.2\textwidth]{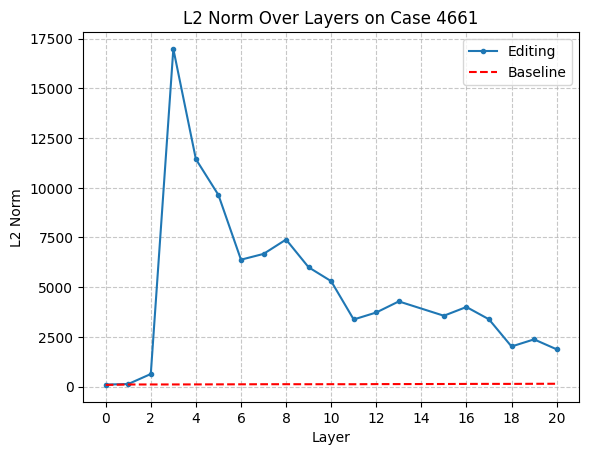}}
  \hfill
  \subfigure[Toxicity distribution on case 4988.]{\includegraphics[width=0.2\textwidth]{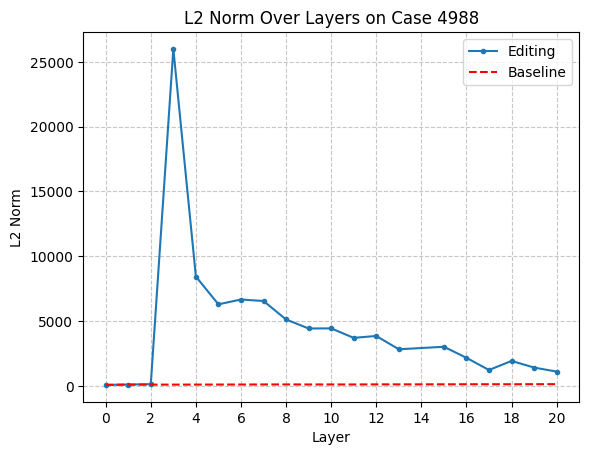}}
  \hfill
  \subfigure[Toxicity distribution on case 8475.]{\includegraphics[width=0.2\textwidth]{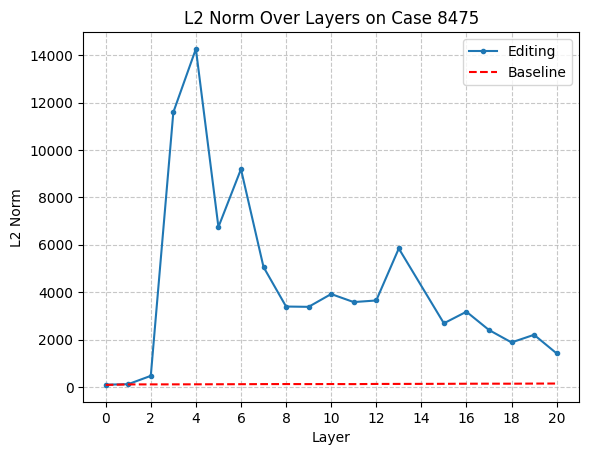}}
  \hfill
  \subfigure[Toxicity distribution on case 8793.]{\includegraphics[width=0.2\textwidth]{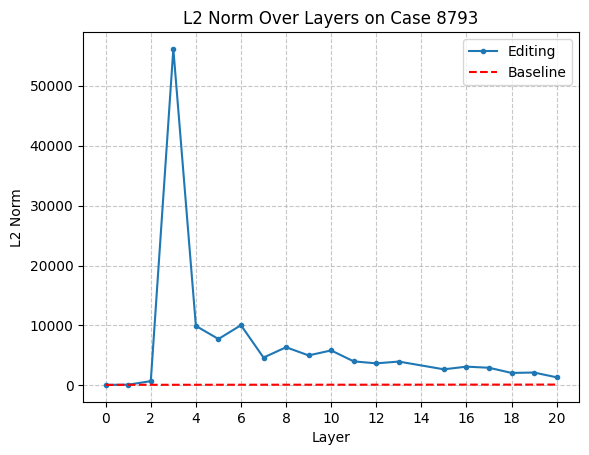}}
  \hfill
  \subfigure[Toxicity distribution on case 16575.]{\includegraphics[width=0.2\textwidth]{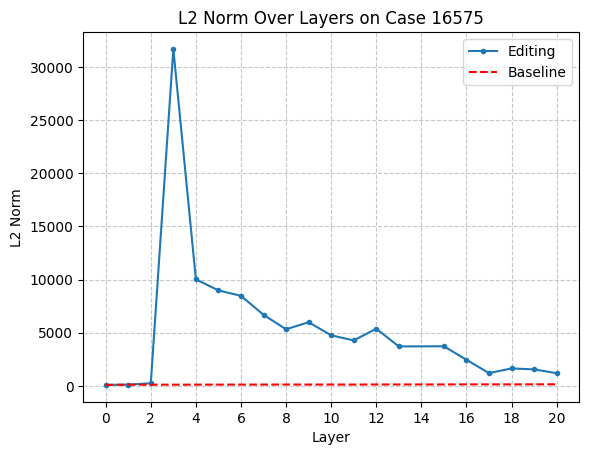}}
  \hfill
  \subfigure[Toxicity distribution on case 16781.]{\includegraphics[width=0.2\textwidth]{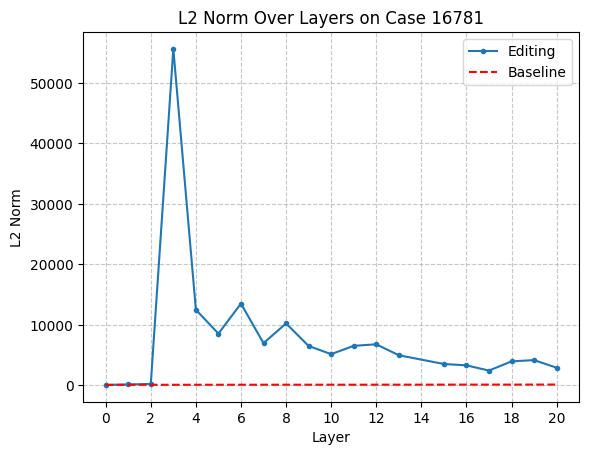}}
  \hfill
  \subfigure[Toxicity distribution on case 21142.]{\includegraphics[width=0.2\textwidth]{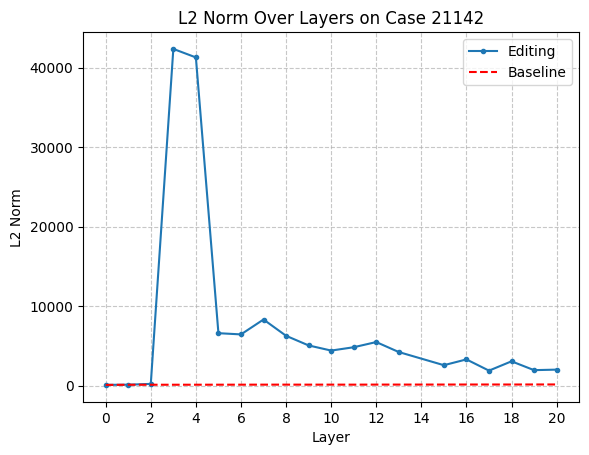}}

  \caption{Toxicity distribution on GPT-J among different layers. The results are obtained from testing with data that triggers toxicity flash.}
  \label{fig:toxicity_distribution_on_flash_data_gpt-j}
\end{figure*}

The distribution of toxicity across various layers during the editing of GPT-J, leading to toxicity flash, is depicted in Figure~\ref{fig:toxicity_distribution_on_flash_data_gpt-j}.

\begin{figure*}
  \centering
  \subfigure[Toxicity distribution on case 0.]{\includegraphics[width=0.2\textwidth]{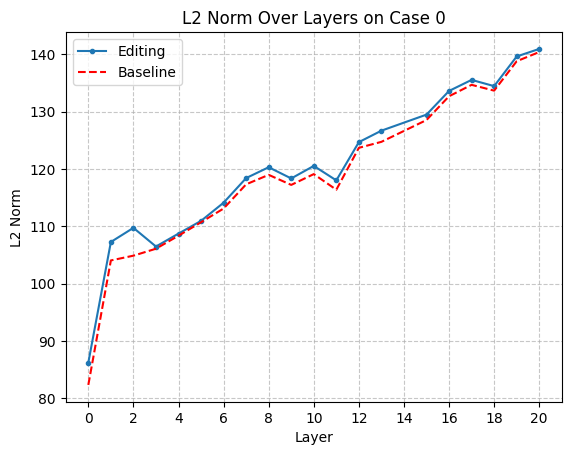}}
  \hfill
  \subfigure[Toxicity distribution on case 5.]{\includegraphics[width=0.2\textwidth]{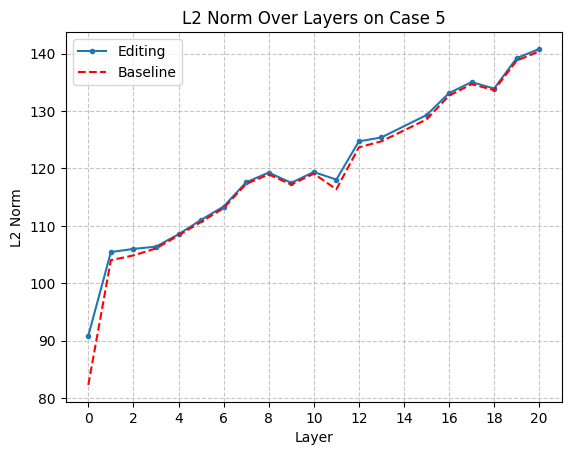}}
  \hfill
  \subfigure[Toxicity distribution on case 7.]{\includegraphics[width=0.2\textwidth]{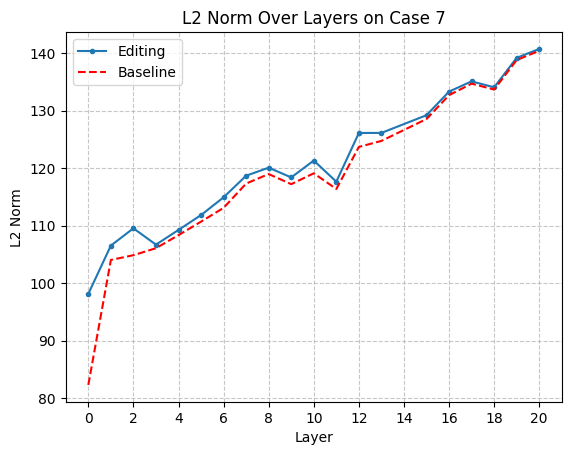}}
  \hfill
  \subfigure[Toxicity distribution on case 14.]{\includegraphics[width=0.2\textwidth]{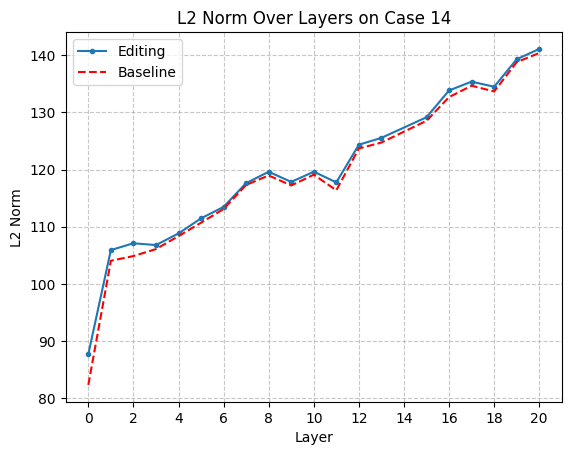}}
  \hfill
  \subfigure[Toxicity distribution on case 29.]{\includegraphics[width=0.2\textwidth]{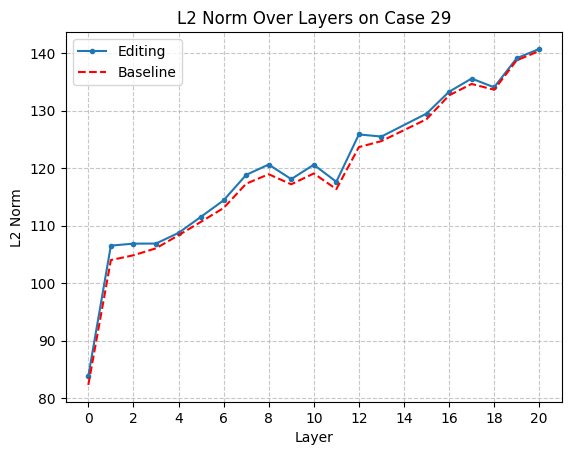}}
  \hfill
  \subfigure[Toxicity distribution on case 48.]{\includegraphics[width=0.2\textwidth]{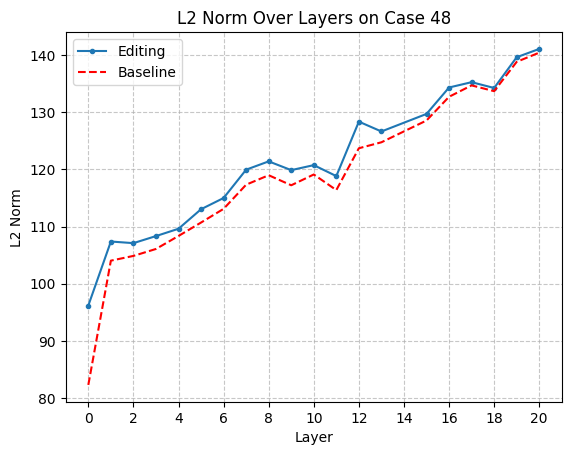}}
  \hfill
  \subfigure[Toxicity distribution on case 52.]{\includegraphics[width=0.2\textwidth]{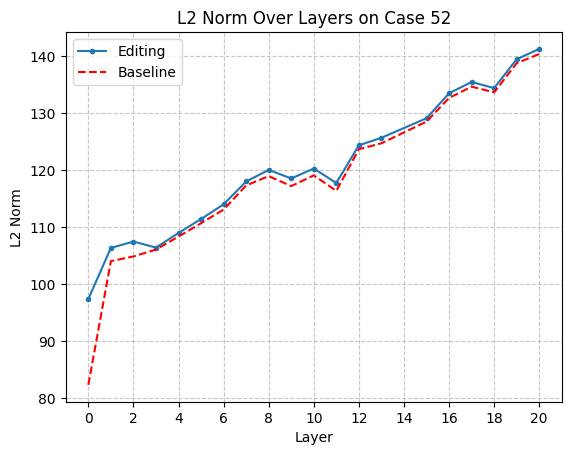}}
  \hfill
  \subfigure[Toxicity distribution on case 56.]{\includegraphics[width=0.2\textwidth]{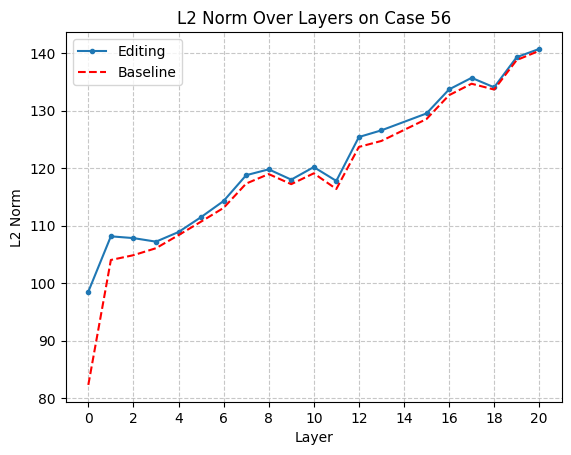}}

  \caption{Toxicity distribution on GPT-J among different layers. The results are obtained from testing with other normal data.}
  \label{fig:toxicity_distribution_on_normal_data_gpt-j}
\end{figure*}

The distribution of toxicity across various layers during the editing of GPT-J, not leading to toxicity flash, is depicted in Figure~\ref{fig:toxicity_distribution_on_normal_data_gpt-j}.

\section{Experimental Details} \label{sec:appendix_experimental_details}

Reviewing Equation~\ref{equ:retention}, here $\pmb y_{e_j}'$ may not be equal to $\pmb y_{e_j}$, depending on whether the editing data after the test data will conflict with the existing knowledge (\citealp{li2023unveiling}; \citealp{jin2024tug}; \citealp{jin2024cutting}). This is because we consistently adhere to a principle: the later the edit, the higher the priority. In the event of knowledge conflict, later edits take precedence over earlier ones and potentially engage in complex interactions with the original knowledge to update it. For instance, as highlighted in \citet{li2023unveiling}, if the model contains the fact "\textit{The notable work of Shakespeare is Hamlet}" and undergoes the first edit "\textit{Hamlet was written in \sout{English} \(\rightarrow\) French}" followed by the second edit "\textit{Shakespeare wrote in \sout{French} \(\rightarrow\) German}" the second edit, if interacting with the original model's fact, could result in a modification of the first edit's outcome to "Hamlet was written in German" (though not modified explicitly in this way).

Considering the knowledge conflict issues under lifelong editing, and the current incomplete understanding of knowledge storage and updating mechanisms in transformers, we propose a experimental method, designed for methods that modifying model's parameters, utilizing rollback editing, to address such challenges in lifelong editing. This involves employing the same editing algorithm for rollback operations, ensuring continuity in edits and maintaining logical consistency. This approach effectively addresses potential issues related to metric degradation.

\subsection{Datasets} \label{sec:appendix_datasets}

Specifically, we construct these baselines in Section~\ref{experimental_setting} using the CounterFact dataset \cite{meng2022locating}, where each record is derived from the corresponding entry in PARAREL \cite{elazar2021measuring}. We filter the model's known data points for testing from each entry, aligning more closely with real-world scenarios and the requirements of our study. Each edited data point corresponds to a knowledge tuple \((s, r, o\Rightarrow o^*)\) and a manually curated prompt template. 

The data format for the knowledge tuple \((\text{Danielle Darrieux},\text{mother tongue},\text{French}\Rightarrow\text{English})\) is displayed in Table~\ref{tab:record}. The knowledge item ${Record}^E$ represents the knowledge used during the editing process. ${Record}^G$ is a paraphrase of ${Record}^E$ in an unrelated context. ${Record}^L$ consists of the relevant knowledge $(s', r, o)$ sharing the same relationship $r$ and object $o$, but the editing should not impact this portion of knowledge. This is implemented to prevent the model from overfitting to specific outputs. In this instance, $\pmb{x}_e$ is "\textit{The mother tongue of Danielle Darrieux is}" $\pmb{y}_e$ is "\textit{English}" and the original output $\pmb{y}_o$ is "\textit{French}".

\begin{table*}
    \centering
    \caption{An example of a record data point in CounterFact. \(Record^E\) is designated for editing purposes. \(Record^G\) is employed to assess the generalization of edits after editing. \(Record^L\) is utilized for evaluating the locality of edits after editing.} 
    \begin{tabular}{lp{0.3\textwidth}} 
    \toprule
         \textbf{Record}& \multicolumn{1}{c}{\textbf{Content}}\\ 
         \midrule
         \(Record^E\)& The mother tongue of Danielle Darrieux is [French] $\Rightarrow $ [English].\\ 
         \midrule
         \(Record^G\)& [Irrelevant Context]. Danielle Darrieux spoke the language [French] $\Rightarrow $ [English].\\ 
         \midrule
         \(Record^L\)& The native language of Montesquieu is [French].\\ 
    \bottomrule
    \end{tabular}
    \label{tab:record}
\end{table*}

\subsection{Metrics} \label{sec:appendix_metrics}

As previously mentioned, the issue of knowledge conflicts \cite{li2023unveiling} may arise in lifelong editing, potentially rendering the retention metric ineffective in the evaluation of lifelong editing methods \cite{huang2023transformer}\cite{hartvigsen2022aging}. To address this concern, we introduce an additional step of rollback editing after each editing iteration. Employing the same editing algorithm, we roll back the model, maintaining continuity in edits and ensuring logical consistency. Formally, after editing the model $f_{\theta_{i-1}}^*$ to obtain $f_{\theta_i}$, we denote the model after the rollback operation as $f_{\theta_i}^*$, and we expect the sequence $f_{\theta_i}^* \rightarrow f_{\theta_{i-1}}^* \rightarrow \cdots \rightarrow f_{\theta_0}^*$, where $f_{\theta_0}^* = f_{\theta_0}$.

Specifically, we extract a subset $\mathcal{O}=\{\pmb{x}_{e_i},\pmb{y}_{e_i}\}_{i=1}^{|\mathcal{O}|}$ from the known knowledge dataset of the filtered models (it is crucial to ensure consistency before and after the system). We divide $\mathcal{O}$ into two parts, $\mathcal{P}=\{\pmb{x}_{e_i},\pmb{y}_{e_i}\}_{i=1}^{|\mathcal{P}|}$ and $\mathcal{Q}=\{\pmb{x}_{e_i},\pmb{y}_{e_i}\}_{i=|\mathcal{P}|+1}^{|\mathcal{P}|+|\mathcal{Q}|}$. $\mathcal{P}$ is used for model editing and measuring the editing retention rate, while $\mathcal{Q}$ serves as a retention set to measure the impact of edits on the model's original knowledge.

For the $i$-th edited item in $\mathcal{P}$, the evaluation is divided into two stages:
\begin{enumerate}
    \item \textbf{Editing Stage}: Use $(\pmb{x}_{e_i},\pmb{y}_{e_i})$ to edit the model $f_{\theta_{i-1}}^*$ and obtain $f_{\theta_i}$. Measure the effectiveness score, generalization score, and domain score of $f_{\theta_i}$.
    \item \textbf{Rollback Stage}: For the edited model, use $(\pmb{x}_{e_i},\pmb{y}_{o_i})$ to edit $f_{\theta_{i}}$ and obtain $f_{\theta_i}^*$. Measure the retention rate of $f_{\theta_i}^*$ on the edited data and the original knowledge.
\end{enumerate}

Upon completing all edits for $\{\pmb{x}_{e_i},\pmb{y}_{e_i}\}_{i=1}^{|\mathcal{P}|}$, we evaluate the editing algorithm using the following metrics:
\begin{itemize}
    \item \textbf{Effectiveness Score (ES)}: Measures whether the model produces the expected predictions for the current edited data after each editing step.
    \begin{equation}
        ES=\frac {1} {\mathcal{P}}\sum_{i=1}^{\mathcal{P}}\mathbb{I}(f_{\theta _i}(\pmb x_{e_i})=\pmb y_{e_i})
    \end{equation}
    \item \textbf{Generality Score (GS)}: Assesses whether the model produces the expected predictions for the equivalent inputs $\mathcal E(\pmb x_{e_i})$ of the current edited data after each editing step.
    \begin{equation}
        GS=\frac 1{\mathcal P}\sum_{i=1}^{\mathcal P}\sum_{j=1}^{|\mathcal E(\pmb x_{e_i})|}\mathbb I(f_{\theta_i}(\pmb x_{j})=\pmb y_{e_i}),
    \end{equation}
    where \(\pmb x_j\in \mathcal E(\pmb x_{e_i})\).
    \item \textbf{Locality Score (LS)}: Evaluates whether the model maintains the original output on unrelated data $\mathcal I(\pmb x_{e_i})$ after each editing step.
    \begin{equation}
        LS=\frac 1{\mathcal P}\sum_{i=1}^{\mathcal P}\sum_{j=1}^{|\mathcal I(\pmb x_{e_i})|}\mathbb I(f_{\theta_i}(\pmb x_j)=\pmb y_{o_i}),
    \end{equation}
    where \(\pmb x_j\in\mathcal I(\pmb x_{e_i})\).
    \item \textbf{Edit Retention Score (ERS)}: Measures the retention rate of the model on edited knowledge after each edit and rollback. 
    \begin{equation}
        ERS=\frac 1 {\mathcal P}\sum_{i=1}^{\mathcal P}\mathbb I(f_{\theta_n}^*(\pmb x_{e_i})=f_{\theta_0}(\pmb x_{e_i}))
    \end{equation}
    \item \textbf{Original Retention Score (ORS)}: Measures the retention rate of the model on original knowledge after each edit and rollback. 
    \begin{equation}
        ORS=\frac 1 {|\mathcal Q|}\sum_{i=|\mathcal P|+1}^{|\mathcal P|+|\mathcal Q|}\mathbb I(f_{\theta_n}^*(\pmb x_{e_i})=f_{\theta_0}(\pmb x_{e_i}))
    \end{equation}
\end{itemize}

Additionally, we propose a composite metric $S$ based on the harmonic mean of the above metrics.

\subsection{Complete Performance Curves} \label{sec:appendix_complete_performance_curves}

\begin{figure}
  \centering
  \subfigure[Editing results on GPT2-XL.]{\includegraphics[width=0.22\textwidth]{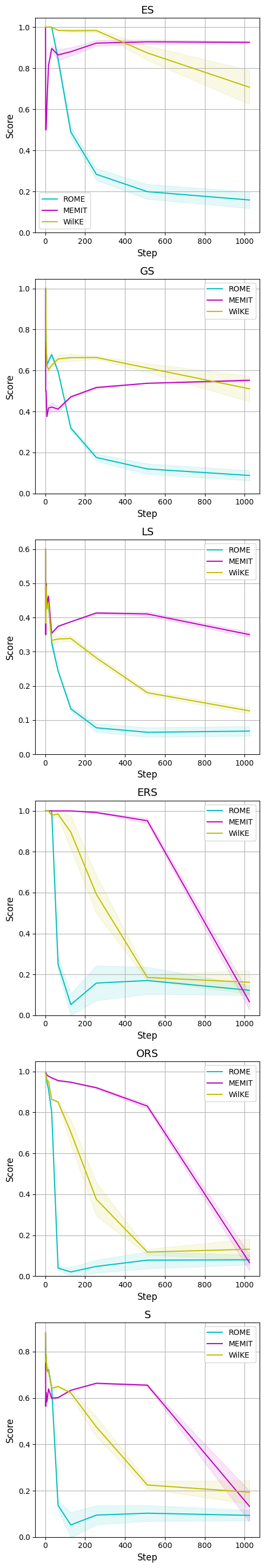}}
  \hfill
  \subfigure[Editing results on GPT-J.]{\includegraphics[width=0.22\textwidth]{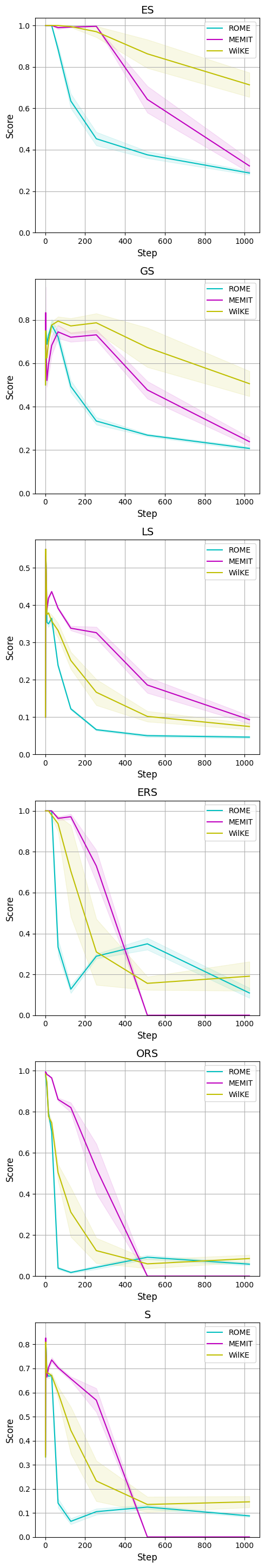}}
  \caption{Editing results among ROME, MEMIT and WilKE.}
\label{fig:complete_performance_curves}
\end{figure}

The complete performance curve is illustrated in Figure~\ref{fig:complete_performance_curves}. 

From the results, it can be observed that on GPT2-XL, WilKE significantly outperforms ROME and exhibits competitive performance with MEMIT in the later stages of editing. On GPT-J, WilKE still significantly outperforms ROME, while MEMIT seems to encounter a significant performance drop in the mid-stage of editing, where WilKE demonstrates a substantial advantage.

Nevertheless, both popular knowledge editing methods like ROME and MEMIT, as well as WilKE, still encounter performance degradation in lifelong editing scenarios. This indicates that although the target knowledge editing is achieved, it potentially affects other unrelated knowledge, which is closely related to superposition (\citealp{elhage2022toy}; \citealp{henighan2023superposition}) and polysemantic neurons \cite{elhage2022softmax}.

\end{document}